\documentclass{article}

\usepackage{PRIMEarxiv}

\usepackage[utf8]{inputenc} % allow utf-8 input
\usepackage[T1]{fontenc}    % use 8-bit T1 fonts
\usepackage{url}            % simple URL typesetting
\usepackage{booktabs}       % professional-quality tables
\usepackage{amsfonts}       % blackboard math symbols
\usepackage{nicefrac}       % compact symbols for 1/2, etc.
\usepackage{microtype}      % microtypography

\usepackage{graphicx}
\graphicspath{{media/}}     % organize your images and other figures under media/ folder

\usepackage{xcolor}
\usepackage{tikz}
\usepackage{mathtools}
\usepackage{amsmath,amssymb,amsfonts, amsthm}
\usepackage{enumitem}
\usepackage{subfig}
\usepackage{physics}
\usepackage[ruled,vlined]{algorithm2e}

\usepackage{bbm} % for indicator function 

\usepackage{makecell}

\DeclarePairedDelimiterX{\Norm}[1]{\lVert}{\rVert}{#1}

\newcommand{\E}{\mathbb{E}}
\newcommand{\pr}{\mathbb{P}}

\newcommand{\cE}{\mathcal{E}}
\newcommand{\cG}{\mathcal{G}}
\newcommand{\cN}{\mathcal{N}}
\newcommand{\cX}{\mathcal{X}}
\newcommand{\cY}{\mathcal{Y}}
\newcommand{\cZ}{\mathcal{Z}}

\newcommand{\bR}{\mathbb{R}}
\newcommand{\cT}{\mathcal{T}}
\newcommand{\bX}{\mathbf{X}}
\newcommand{\bY}{\mathbf{Y}}

\newcommand{\comp}{\mathsf{c}} 
\newcommand{\defeq}{\vcentcolon=}
\newcommand{\Sone}{S_r(x)}
\newcommand{\Stwo}{S_r}
\newcommand{\phiOne}{\Phi_{r}(x)}
\newcommand{\phiTwo}{\Phi_r}

\usepackage{algorithmic}

\usepackage{dsfont}
\DeclareSymbolFont{bbold}{U}{bbold}{m}{n}
\DeclareSymbolFontAlphabet{\mathbbold}{bbold}
\newcommand{\ind}{{\mathbbold 1 }}

\newtheorem{example}{Example} 
\newtheorem{theorem}{\textbf{Theorem}}
\newtheorem{lemma}[theorem]{Lemma} 
\newtheorem{proposition}[theorem]{Proposition} 
\newtheorem{remark}{Remark}
\newtheorem{corollary}[theorem]{Corollary}
\newtheorem{definition}{Definition}

\newtheorem{notation}[theorem]{Notation}

\begin{document}
%\title{Efficient Adaptation in Covariate-Shift Using a Classification Tree}
\title{Classification Tree Pruning under Covariate Shift}

%\author{Nicholas~R.~Galbraith,
%        Samory~Kpotufe% <-this % stops a space
%\thanks{N. Galbraith and S. Kpotufe are with the Department
%of Statistics, Columbia University, New York,
%NY, 10027 USA e-mail: nrg2136@columbia.edu}}% <-this % stops a space

\author{
  Nicholas Galbraith \\
  Department of Statistics \\
  Columbia University \\
  \texttt{nicholas.galbraith@columbia.edu} \\
\And 
  Samory Kpotufe \\
  Department of Statistics \\
  Columbia University \\
  \texttt{samory@columbia.edu}
  % examples of more authors
}

%\author{Nicholas~R.~Galbraith,
%        Samory~Kpotufe% <-this % stops a space
%\thanks{N. Galbraith and S. Kpotufe are with the Department
%of Statistics, Columbia University, New York,
%NY, 10027 USA e-mail: nrg2136@columbia.edu}% <-this % stops a space

\maketitle

\begin{abstract}
We consider the problem of \emph{pruning} a classification tree, that is, selecting a suitable subtree that balances bias and variance, in common situations with inhomogeneous training data. Namely, assuming access to mostly data from a distribution $P_{X, Y}$, but little data from a desired distribution $Q_{X, Y}$ with different $X$-marginals, we present the first efficient procedure for optimal pruning in such situations, when cross-validation and other penalized variants are grossly inadequate.  
Optimality is derived with respect to a notion of \emph{average discrepancy} $P_{X} \to Q_{X}$ (averaged over $X$ space) which significantly relaxes a recent notion---termed \emph{transfer-exponent}---shown to tightly capture the limits of classification under such a distribution shift. Our relaxed notion 
can be viewed as a measure of \emph{relative dimension} between distributions, as it relates to existing notions of information such as the Minkowski and Renyi dimensions.

\end{abstract}

%\begin{IEEEkeywords}
%Classification Tree, Covariate Shift, Model Selection and Pruning, Nonparametrics, Minkowski and Renyi dimensions. 
%\end{IEEEkeywords}

% \fi
%\IEEEpeerreviewmaketitle

\section{Introduction}

Decision trees are one of the most enduring tools for classification, given their ability to adapt to complex patterns in data. Such adaptability requires proper \emph{pruning}, our main concern in this work, that is, the selection of a most appropriate subtree, or tree level, that best fits unknown patterns. 
In particular, much of the existing theory, and resulting pruning procedures, concerns the ideal case of i.i.d. data, although in fact, it is now common in practice to combine training data from multiple inhomogeneous sources. For instance, in modern application domains such as medicine, computer speech and vision, ideal i.i.d. target data is hard to come by, so one relies on related but different data sources. 

A common assumption in these settings with inhomogeneous data---which we adopt in this work---is that of so-called \emph{covariate-shift}, whereby only \emph{marginal distributions} are shifted from source to target data distributions. For intuition, consider for instance a drug study targeting a particular country $Q$ with a given social makeup, while much data from another population $P$ might be available. In other words, most samples $(\bX,\bY)_{P}$ would be drawn from a distribution $P_{X, Y}$, in addition to some samples $(\bX,\bY)_{Q}$ from an ideal distribution $Q_{X, Y}$; the two are different but related, in that, $Q_X\neq P_X$, i.e., signifying different social makeup, while $Q_{Y|X}=P_{Y|X}$, i.e., drug responses conditioned on social variables $X$ remain the same.  
%where $Q_X\neq P_X$, e.g., signifying different populations, but where . 
We are interested in reducing error under target $Q$, beyond what is possible using $(\bX,\bY)_{Q}$ alone, by harnessing information from $(\bX,\bY)_{P}$; the extent to which this is possible depends not only on good {model selection}, but also 
on \emph{how far} $P_X$ is from $Q_X$, which we also aim to appropriately capture.  

Unfortunately, existing model selection approaches, specifically for tree pruning, do not readily extend to this setting with mixed data. For instance, cross-validation approaches, often penalizing by subtree-size, might aim to use \emph{the limited samples from $Q$}, since a priori this is most faithfully indicative of future performance under $Q$; however, failing to integrate the potentially much larger $P$ data can induce too large a variance to guarantee an optimal choice. Our main contribution is to derive a pruning approach that integrates both $P$ and $Q$ data to guarantee performance commensurate with the aggregate amount of data and the \emph{distance} between $P_X$ and $Q_X$, appropriately captured. 

As a first study of tree pruning under covariate-shift, we focus on the case of \emph{dyadic trees}, which are known to be amenable to analysis \cite{Scott2002DyadicCT, bounds_alg_DDT}, and derive a practical pruning procedure, not only guaranteed to adapt automatically to the amount of shift from $P_X$ to $Q_X$, but yielding significant practical improvements over baselines of available cross-validation approaches. Our empirical studies not only confirm our theoretical results, but demonstrate wider applicability of our procedure beyond the case of dyadic trees.   

Our approach for tree-pruning instantiates a so-called \emph{Intersecting Confidence Intervals} (ICI) approach, a generic strategy for adaptation to unknown distributional parameters \cite{ICIGoldNem, Lepski1997}. Here in particular, in the covariate-shift setting, we parametrize the unknown level of \emph{shift} $P_X \to Q_X$, via a relaxation of a recently proposed notion of \emph{transfer-exponent} \cite{KM} which captures (a worst-case notion of) the mass $P_X$ assigns to regions of large $Q_X$-mass. Our relaxation, which we term \emph{aggregate-transfer-exponent}, instead captures \emph{an average sense} of the mass $P_X$ assigns to regions of large $Q_X$-mass \emph{near the decision boundary}, and can be more easily interpretable, as it readily ties into traditional notions of information such as the Minkowski or Renyi dimensions. % (\skr{see Section \ref{}}). 

Our main technical result is a guarantee that ICI, as instantiated for tree-pruning, achieves a rate adaptive to the unknown level of shift, as expressed in terms of aggregate-transfer-exponent $P_X\to Q_X$ (Theorems \ref{adaptiveTheorem}, \ref{localized_adaptive_theorem}). As such, transfer rates are more tightly controlled by this milder measure of shift, yielding at times faster rates than expressible via the transfer-exponent of \cite{KM} (see Remark  \ref{sharper_bounds_than_KM} in Section \ref{sec_relation_to_transfer_exponent}). A main difficulty in establishing this result lies in the fact that, unlike in usual analyses of tree-based classification, our covariate-shift setting precludes the usual simplifying assumption in tree-based prediction that the marginal measure is uniform on its support $\cal X$. Without the uniform-measure assumption, which allows control of the mass of points in regions of space, and hence of prediction variance, we instead have to proceed via careful integration of the risk at appropriate scales over regions of space, resulting in the first proof that (dyadic) tree-based classification attains the minimax lower-bound given by \cite{Audibert2007} for general distributions; in particular, this reveals a choice of optimal tree level, qualitatively different than prescribed by the usual theory under a uniform-measure assumption (see Remark \ref{no_strong_density_assumption}). As such, our results are in fact of independent interest even for the usual i.i.d. setting where $P_X = Q_X$.

\subsection{Background and Related Work}
Model selection is arguably the most important challenge in the context of decision trees, and thus naturally the focus of much of the literature on the subject. In its most general form, model selection encompasses all decisions on partitioning the $X$ space, giving rise to both \emph{greedy} methods such as CART \cite{CART} which iteratively refine a space partitioning guided by labeled data $(\bX, \bY)$, and \emph{pruning} methods which select a subtree (using $\bY$ labels) of an initial hierarchical partition tree built primarily using $\bX$ data alone. While both approaches work well in practice, the greedy approaches admit fewer theoretical guarantees in the literature. 
Our focus in this work is on pruning methods as discussed above, although, interestingly, our approach may serve to refine even greedily obtained trees such as CART (see Appendix \ref{appendix_experiments}).  

Various pruning approaches have been proposed in the i.i.d. settings $P_X = Q_X$, the simplest ones consisting of selecting a \emph{tree level} by cross-validation, where a level corresponds to a partition into cells of a given maximum diameter. These methods are efficient in that there are typically $O(\log n)$ levels to explore for a sample of size $n$, as opposed to visiting all possible subtrees, of which there are $O(2^{n})$ \cite{subtrees_of_trees}. Quite importantly, such simple cross-validation admits strong guarantees in the i.i.d. setting, by extending from traditional guarantees on the performance at each level---typically for dyadic trees \cite{gyorfiNonparametrics}. 
However, in our context where $P_X \neq Q_X$, such guarantees no longer apply, while furthermore, cross-validation is limited by the amount $n_Q \ll n_P$ of $Q$ data, making it unlikely to obtain guarantees that integrate both data sizes as obtained here. 

Methods more elaborate than cross-validation, generally referred to as \emph{cost-complexity pruning}, have been proposed in the i.i.d. setting, which in particular can return any possible subtree by relying on clever dynamic programs to efficiently explore the very large $O(2^{n})$ search space. In particular, these dynamic programs work by splitting the (estimated) classification error over cells of a partition (leaves of a subtree) to minimize this error globally. Sharp guarantees are achievable in the i.i.d. setting (see e.g. \cite{DDT})  by relating local error estimates in cells---which integrate in notions of local tree complexity to account for variance---to expected \emph{target} error. Unfortunately, it is a priori unclear how to extend such estimates of target error, or involved complexity and or local variance measures, to the covariate-shift setting where relatively little target data is available locally in each cell. Initial experiments with such approaches indicate their inadequacy in the covariate-shift setting, suggesting the need for rigorous, independent investigation as to whether they could be extended to non-i.i.d. settings as considered here (see Appendix \ref{appendix_experiments}).

Covariate-shift \cite{KM, reeve2021adaptive, IWCV, Gretton2009CovariateSB, pathak2022new}, and more general changes in distribution \cite{Cai2021, DA_bounds_and_algorithms_Mohri, impossibility_theorems_domain_adaptation, Scott_DomainAdaptation, Jordan_MetaLearning2021, Lee_MetaLearning2020}, have been receiving much renewed attention as it has become clear that the i.i.d. setting fails to capture many contemporary applications. Many interesting measures of shift have thus been proposed, the closest to the present work being the \emph{transfer-exponent} of \cite{KM} and its recent relaxations in \cite{pathak2022new}, \cite{reeve2021adaptive} which similarly aim to capture an average sense of local deviations $P_X\to Q_X$. Our initial notion of \emph{aggregate-transfer-exponent} appears most appropriate for partition-trees as it is defined over \emph{$r$-grids} of the $X$ space, and appears more readily interpretable as it relates to existing notions of information. Our main adaptivity guarantee in Theorem \ref{localized_adaptive_theorem} further localizes this aggregate measure to the unknown decision boundary so as to more tightly capture the relevant shift from $P_X$ to $Q_X$. 

While all the above works propose model-selection approaches under covariate-shift, they are invariably concerned with classifiers other than decision trees; in particular, the works of \cite{KM, reeve2021adaptive} implement an ICI approach, e.g., for nearest neighbor classifiers, which not only do not extend to our case, but result in computationally impractical methods due to the nature of the classifiers considered (see Section \ref{computational_considerations}). An important exception is the approach of \cite{IWCV}, which implements importance-weighted risk estimates based on estimates of density ratios $d_{Q_X}/d_{P_X}$. Such density ratio estimates are generally attractive, and may be applied in our context of decision trees, but not only can result in expensive pre-processing steps, but may also fail to yield accurate $Q$-risk estimates when $d_{Q_X}/d_{P_X}$ does not exist (i.e. $Q_X$ is singular w.r.t. $P_X$) which is a practical scenario allowed in this work (and also in \cite{KM, pathak2022new, reeve2021adaptive}).

\subsection{Paper Outline}

The paper is organized as follows. In Section \ref{sec_setup}, we formally state the problem, and introduce the assumptions under which our analysis proceeds, including our condition relating the source distribution $P$ to the target $Q$. In Section \ref{section_overview} we define the decision tree models that we study and present a sequence of results that build up to our main adaptive result, Theorem \ref{localized_adaptive_theorem}. Intuition about our algorithm and practical considerations are discussed in this section as well. Section \ref{secOnDim} relates the aggregate transfer exponent to various notions of information, such as the Minkowski and Renyi dimensions; proofs of results from this Section are in Appendix \ref{appendix_properties}. Our analysis is presented in Section \ref{sec_analysis_details}, with proofs of auxiliary results given in the Appendix \ref{appendixProofs}. Our results are supported experimentally via implementations on real data in Section \ref{secExperiments}; results of some supplemental experiments given in Appendix \ref{appendix_experiments}. In Appendix \ref{appendix_multiSource}, we discuss extensions to multiple source problems.

\section{Problem Setup and Notation.} \label{sec_setup}

%Whenever possible, we will follow notation used by \cite{DDT} and \cite{Audibert2007}. 
%In this section we give a set-up of the problem, and describe the %conditions under which our results hold by defining the classes of %distributions over which we can show minimax optimality. \par 
We consider the binary classification problem, with data $(X,Y)$ lying in a set $\cX \times \cY$ with $\cY = \{0,1\}$, and we assume that $\cX = [0,1]^D$. Unless otherwise specified, we write $\Norm{\cdot}, \Norm{\cdot}_\infty$ for the Euclidean and $\ell_\infty$ norms respectively, and let $B(x,r)$ denote the open $\ell_\infty$ ball of radius $r$ about $x$; that is $B(x,r) = \{y \in [0,1]^D: \Norm{y - x}_\infty < r\}$. \par In transfer learning, we simultaneously consider a \emph{source} distribution $P$ and a \emph{target} distribution $Q$, both on $\cX \times \cY$. We let the marginal feature distributions be denoted by $P_X$ and $Q_X$ respectively, and suppose that these are supported on $\cX_P, \cX_Q \subset \cX$. We assume that a sample $(X_i,Y_i)_{i=1}^n$ is observed, where $n = n_P + n_Q$ and we write \begin{align*} & (\bX,\bY)_{P} = (X_1,Y_1), \dots, (X_{n_P},Y_{n_P}) \sim P \quad \mathrm{and} \\
&(\bX,\bY)_{Q} = (X_{n_P + 1},Y_{n_P + 1}), \dots, (X_{n_P + n_Q},Y_{n_P + n_Q}) \sim Q,
\end{align*} with all observations being independent, and we let $(\bX,\bY) = (\bX,\bY)_P \cup (\bX,\bY)_Q$ denote the full sample. %We let $\Pi^n \defeq P^{ n_P} \times Q^{n_Q}$ denote the joint distribution of the source and target samples. 

The goal is to produce classifiers $f: \cX \mapsto \cY$ with good performance under the target distribution, as captured by its \emph{risk}, defined as follows.  
\begin{definition} For a function $f: \cX \to \cY$, the \textbf{risk} of $f$ under the target $Q$, denoted $R(f)$, is defined as $$R(f) \defeq \E_Q \ind\{f(X) \neq Y\},$$ which simply gives the probability that $f$ incorrectly classifies a point drawn from $Q$. Let $R^* = \inf_f R(f)$, and define the \textbf{excess risk} of $f$ as 
$$\cE(f) \defeq R(f) - R^*.$$ 

\end{definition} 

Note that in the case of binary classification ($\cY = \{0,1\}$), we have $R(f^*) = R^*$ for $f^*(x) = \ind\{\eta_Q(x) \geq 1/2\}$, where $\eta_Q(x) \defeq Q(Y = 1 \mid X = x)$ is the \emph{regression function}. We study the particular case of \emph{covariate shift}, which supposes that the class probabilities given the features are common to the source and target, so that $P(Y = y \mid X) = Q(Y = y \mid X)$; we may therefore without ambiguity write $\eta$ for the regression function common to $P$ and $Q$.

%; 
%  
%For a sample $(X_i,Y_i)_{i=1}^n$ from a %distribution $P$, we let $\hat{P}_n$ %denote the empirical distribution, that %is, for a measurable $B \subset \cX \times %\cY$ we have $\hat{P}_n(B) = \tfrac{1}{n} %\sum_{i=1}^n \ind\{(X_i,Y_i) \in B\}$, %where $\ind(\cdot)$ denotes the indicator %function; when $n$ is understood we write %simply $\hat{P}(B)$. \par 

Let $\hat f_n$ denote a generic classifier learned from the sample $(\bX, \bY)$. This work is concerned with understanding rates of convergence of $\E_{(\bX,\bY)}[\cE(\hat{f}_n)] \to 0$ in terms of source and target samples in various regimes, and developing efficient algorithms which adaptively (i.e. without recourse to problem-specific knowledge) attain these rates. Here and elsewhere $\E_{(\bX,\bY)}$ denotes expectation over the sample, while we use $\E$ for expectation over the sample and an independent target point (that is, over $P^{n_P} \times Q^{n_Q} \times Q$). Further notation will be introduced as necessary.

\subsection{Assumptions on the Regression Function.}\label{sec:conditions}
 %%% re-word?
%Recall that we consider $P,Q$ with marginals $P_X,Q_X$ supported on $\cX = [0,1]^D$ and with common regression function $\eta \defeq \eta_P = \eta_Q$. \par 
We restrict the complexity of the class of regression functions under consideration by making the the standard assumption that this regression function belongs to a class of H\"older-continuous functions. 
\begin{definition}
For $\alpha \in (0,1)$, $L > 0$, the regression function $\eta:[0,1]^D \to \bR$ is said to be $(L,\alpha)$-\textbf{H\"older continuous} if 
\begin{equation}\label{holderSmooth}
\abs{\eta(x) - \eta(x')} \leq L \norm{x - x'}_{\infty}^\alpha \quad \textrm{for all }x,x' \in \cX. \end{equation} 
\end{definition}

\begin{remark}\label{infinity_balls}
Stating the H\"older condition using $\ell_\infty$ rather than $\ell_2$ balls is done purely as a matter of convenience, since the cells of the trees naturally correspond to $\ell_\infty$ balls. This makes no difference to the analysis, since using any equivalent norms in $\mathbb{R}^d$ would simply change the constant $L$, and hence the leading constants in our rates.  
\end{remark}

We adopt the following classical {\it noise condition}, which captures the difficulty of classification under $Q$, by parametrizing the likelihood of queries $x$ with low margin between class probabilities (see \cite{Audibert2007} for a detailed discussion).
%Many authors introduce further conditions which capture additional aspects of the data. In particular, we consider the following \emph{noise condition}; 
 
\begin{definition}
%Let $Q$ be a distribution over $(X,Y)$ with marginal $X \sim Q_X$ and conditional $Q(Y = 1 \mid X) = \eta(X)$. 
We say that $Q$ satisfies \textbf{Tsybakov's noise condition} with parameters $C_\beta,\beta \geq 0$, if 
\begin{equation}\label{tsybNoise}
    Q_X(\abs{\eta(X) - 1/2} \leq t) \leq C_\beta t^\beta.
\end{equation} 
\end{definition}

\subsection{Relating the Source to the Target. } 
We require a condition on the support of $Q$; let us quickly recall a standard notion.

\begin{definition}
 A set $\cZ \subset \cX_Q$ is called an \textbf{$r$-cover} of $\cX_Q$ if $\cX_Q \subset \cup_{z \in \cZ} B(z,r)$. The size of the smallest $r$-cover is called the \textbf{$r$-covering number}, denoted $\cN(S,r)$. 
\end{definition}

\begin{definition}\label{polyCovering} We say that $\cX_Q$ is of \textbf{metric dimension} (no greater than) $d$ if there are constants $C_d,d >0$ such that for all $r > 0$ we have \begin{equation}\label{QmetricEntropy}
    \cN(\cX_Q,r) \leq C_d \, r^{-d}. 
\end{equation} 
\end{definition}
Since $\cX_Q \subset [0,1]^D$, the above is always satisfied with $d = D$.
\par In a previous work \cite{KM} on transfer learning under covariate shift, the authors introduce the \it transfer exponent \rm between the source and target measures, which we now recall: 
\begin{definition}
We call $\rho > 0$ a \textbf{transfer exponent} from $P_X$ to $Q_X$ if there exists $C_\rho > 0$ such that 
\begin{equation*}\label{transferExp0}
    \forall \, r \leq \mathrm{diam}(\cX_Q) \text{ and } \forall \, x \in \cX_Q, \quad \frac{Q_X(B(x,r))}{P_X(B(x,r))} \leq C_\rho r^{-\rho}.
    \end{equation*}
\end{definition}

\begin{figure}[t]
    \centering
    \includegraphics[height = 7.3cm, width = 13.7cm]{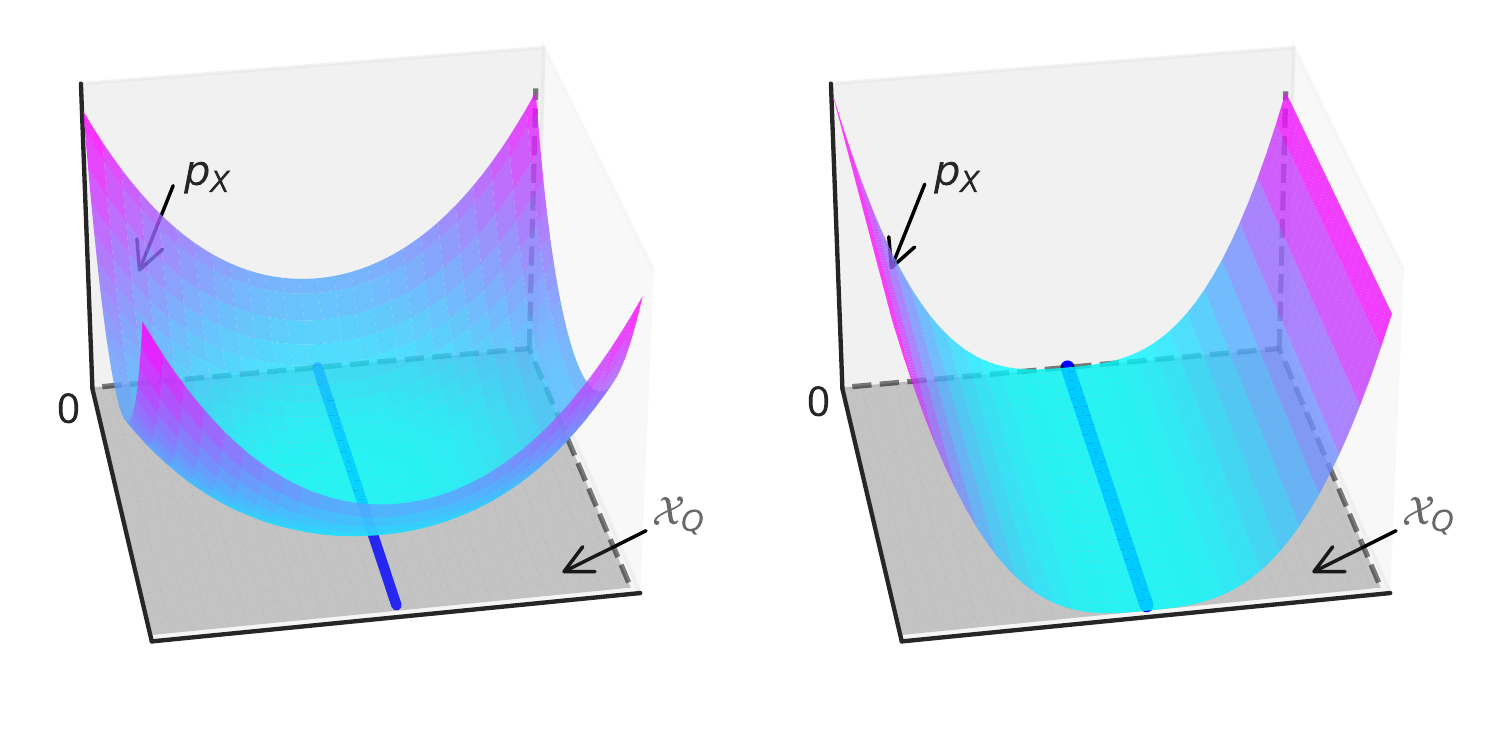}
    % \caption{Two different $P_X$ densities are shown, having the same \emph{transfer-exponent} w.r.t. the same target $Q = \mathrm{U}([0, 1]^2$, although the left density is easier to transfer from: it assigns more mass on average to regions of high $Q$-mass.}
    
    \caption{Two source densities $P_X$ are shown, which yield equal transfer exponents w.r.t $Q_X = \mathrm{U}([0,1]^2)$, despite the fact that the right-hand source density vanishes over a larger region of high target mass. This is captured by the aggregate transfer exponent: while both densities have transfer exponent $\rho = 3$ with respect to $Q_X$, the left-hand density has an aggregate exponent of $\gamma = \rho$, while the right-hand density has an aggregate exponent of $\gamma = \rho + 1$.  The effects of this distinction on transfer will be felt most strongly when, as is the case here, the density ratios vanish along the decision boundary (shown in dark blue). This will be elucidated in Section \ref{Sec_localizingToBoundary} when we introduce a notion of the aggregate exponent localized to the decision boundary.}

    %\caption{Two source densities $P_X$ are shown, having the same transfer exponent $\rho$ with respect to a uniform target measure $Q_X = \mathrm{U}([0,1]^2)$. but with different aggregate exponents $\gamma$. On the left we have the density $p_X(x) \propto \norm{x}^\nu$, giving $\rho = \nu$, $\gamma = \nu$, while on the right we have $p_X(x) \propto x_1^\nu$, leading to $\rho = \nu$ and $\gamma = \nu + 1$. Shown are plots for $\nu = 3$. The higher aggregate exponent for the source measure on the right confirms the intuition that transfer is more difficult when the source has poor coverage of high-mass target regions, since the density vanishes on $\{0\} \times [0,1]$ for the right distribution, while the left distribution has density vanishing only at the origin. 
    
    \label{fig:toy_ex_illustration}
\end{figure}

We will consider pairs of joint distributions over $\cX \times \cY$ with marginals $(P_X,Q_X)$ satisfying the following condition, which is a simple relaxation of the above in which we control the aggregate scaling of the ball-mass ratios rather than the worst-case scaling over the support of the target $\cX_Q$, which will allow us to derive tighter rates in a many scenarios; for an example, see Figure \ref{fig:toy_ex_illustration}. There are a number of ways to define this relaxation, which we will consider in Section \ref{secOnDim}. For now we will us stick to one particular version, for which we first require a definition. 

\begin{definition}\label{gridDef}
 A collection $\Xi$ of subsets of $\cX$ is an \textbf{$r$-grid} of $\cX_Q$ if $\{\cX_Q \cap E: E \in \Xi\}$ is a partition of $\cX_Q$, and for each $E \in \Xi$ such that $E \cap \cX_Q \neq \emptyset$ there is an $x_E \in E \cap \cX_Q$ such that $B(x_E,r) \subset E \subset B(x_E,2r)$.
\end{definition}

We now relate $P$ to $Q$ as follows. 

\begin{definition}\label{aggExpDef}
     %For $Q_X$ satisfying \eqref{QmetricEntropy} for $d \geq 0$, 
     We say that $\gamma \in [0,\infty]$ is an {\bf aggregate transfer exponent} from $P_X$ to $Q_X$ with constant $C_\gamma > 0$ if for any $0 < r \leq 1$, we have that for all $r$-grids $\Xi$ of $\cX_Q$ in $\cX$, 
\begin{equation}\label{globalTransferExp}
    \sum_{E \in \Xi} \frac{Q_X(E)}{P_X(E)} \leq C_\gamma \cdot r^{-\gamma}.
\end{equation}   
\end{definition}
Note that the above holds vacuously for $\gamma = \infty$, so the definition always applies. In Section \ref{secOnDim}, we will show that \eqref{globalTransferExp} is, under regularity conditions, equivalent to an integrated version of the transfer exponent (this quantity was in fact independently introduced recently by \cite{pathak2022new}). Remark that one could consider an alternative measure where the sum in \eqref{globalTransferExp} is instead over elements of the dyadic partition of the ambient space $\cX = [0,1]^D$; this in fact leads to a more pessimistic measure: as argued in Appendix \ref{notAmbientDyadic}, such a  measure is not in general equivalent to the aggregate exponent nor the integrated exponent, and could yield worse rates unless stronger assumptions are imposed on $P_X$ and $Q_X$.  \par 

We will see later how to refine the relation of $P$ to $Q$ in such a way as to account for the conditional distribution of the labels $\bY$ given the features, although for simplicity we will first develop results using the aggregate transfer exponent as we have just defined it. These results will allow us to build towards the main result in Section \ref{Sec_localizingToBoundary}, in which we provide stronger guarantees by localizing the aggregate transfer exponent to the decision boundary.  

\par In Section \ref{secOnDim}, we will consider some properties of the aggregate transfer exponent. In particular, we explicitly relate the aggregate exponent to the transfer exponent of \cite{KM} and demonstrate that considering \eqref{globalTransferExp} refines the rates that can be derived for the convergence of the excess expected risk to zero. In Figure \ref{fig:E1_experiment}, we provide a simulation that demonstrates that the aggregate transfer exponent can more effectively capture the difficulty of transfer between $P$ and $Q$; a simple example of a situation in which this is the case is given in Figure \ref{fig:toy_ex_illustration}. We also show that the aggregate transfer exponent can be related to various notions of \emph{dimension} for measures, for instance, the \emph{Renyi dimension}; in particular we will argue that the aggregate transfer exponent can be interpreted as a \emph{relative dimension} of $P_X$ with respect to $Q_X$.

%\begin{equation}\label{gammaInf}
    %\gamma(P \mid Q) \defeq \inf\left \{ \gamma \geq 0: \, \exists C >0 \quad \mathrm{s.t.} \, \int_{\cX} \frac{1}{P(B(x,r))} \, dQ(x) \leq C r^{-\gamma} \right\}
%\end{equation}

Formally, we consider the following classes of pairs of distributions. 

\begin{definition} \label{classesDef}
Let $\cT = \cT(C_\alpha, \alpha,C_\beta, \beta,C_d, d, C_\gamma, \gamma)$ denote the \textbf{transfer class} of distribution pairs $(P,Q)$ such that  
\begin{itemize}
\item [i)] The conditional distributions $P_{Y\mid X}$ and $Q_{Y \mid X}$ are equal. 
    \item[ii)] The regression function $\eta$ is $(C_\alpha,\alpha)$-H\"older continuous. 
    \item[iii)] $Q_X$ satisfies a $(C_\beta, \beta)$-Tsybakov noise condition. 
    \item[iv)] $Q_X$ has metric dimension $d$ with constant $C_d$. 
    \item[v)] $P_X$ has aggregate transfer exponent $(C_\gamma, \gamma)$,  w.r.t. $Q_X$. %, with constant $C_\gamma$. \eqref{globalTransferExp}. 
\end{itemize}
\end{definition}
Note that under weak conditions on $Q_X$ of metric dimension $d$ we have $\gamma \geq d$, as shown in Proposition \ref{noSuperTransfer}. As we shall see, this implies that in most scenarios of interest it is impossible to improve on the learning rate attainable using target samples if only source samples are available.

%% FOR THE CLASS T', WHICH HAS BEEN REMOVED FROM MAIN BODY

%We consider the following class which assumes added regularity on $Q_X$ that is studied in much of the related literature on classification.

%%%% make this clearer. 
%\begin{definition}
%The transfer class $\cT^\prime$ is defined by saying $(P,Q) \in \cT^\prime$ if $(P,Q) \in \cT$, and if $Q_X$ has a density $q$ with respect to $d$-dimensional Hausdorff measure (denoted $\mathcal{H}^d$) that is uniformly bounded above and below; that is, if there exist $0 < q_0 < q_1$ such that $q_0 \leq q(x) \leq q_1$ for all $x \in [0,1]^d$. Further, we suppose that there exist $c_1,c_2 > 0$ such that $c_1 r^d \leq \mathcal{H}^d(\cX_Q \cap B(x,r)) \leq c_2 r^d$ for all $x \in \cX_Q$, and $0 < r < r_0$ for some $r_0$ (i.e., $Q_X$ is \emph{regular of dimension $d$}).   
%\end{definition}

%\begin{definition}[Constrained Target]
%Let $\cT^\prime$ denote the class of pairs of distributions $(P,Q)$ satisfying the conditions of Definition \ref{classesDef}, and with $Q_X$ satisfying the following additional regularity condition: for all $x \in \cX_Q$ and $r > 0$, we have $Q_X(B(x,r)) \geq c_d r^d$, for $d$ as in \eqref{QmetricEntropy}. 
%\end{definition}

%%% give this condition purely in terms of how Q behaves on balls... 

%need to define Hausdorff measure or provide a reference. -- 
% don't need to cite, and do away with 'regular of dim d'. 
We are now ready to state our results.

\section{Overview of Results.}\label{section_overview}

In this section, we present our results. These will build towards our main theorem to be presented in Section \ref{Sec_localizingToBoundary}: an excess risk bound for a fully adaptive procedure which features a refined notion of the aggregate exponent that is localized to the decision boundary. Our first result towards this end is an oracle rate for a regular dyadic tree with respect to the (non-localized) aggregate transfer exponent. We then present an efficient scheme for adaptively selecting a tree level, yielding a procedure that attains the oracle rate up to log factors, and then finally we show that this adaptive procedure in fact adapts to the localized aggregate exponent.

%In this section we state our main results, beginning with 
%oracle rates for a regular dyadic tree. We then present an %efficient scheme for adaptively selecting a tree level which %attains the oracle rate up to a log factor, and then finally %provide information-theoretic lower bounds for the optimal %transfer learning rates for the excess expected risk over the %class $\cT$, demonstrating that our oracle rates are tight. %The proofs are outlined in the next section, with supporting %Lemmas proved in the Appendices. \par 

We begin by introducing the procedure from which the oracle rate is derived. 

\subsection{Tree-Based Classification}

We briefly introduce some further notation in the following definitions:

 \begin{definition} \label{dyadicTree} For $r\in \{ 2^{-i} \}_{i \in \mathbb{N}}$, let $\Pi_r$ denote a regular partition of $[0, 1]^D$ into hypercubes of side length $r$. A \textbf{dyadic partition tree} refers to a collection of nested partitions $\{\Pi_r\}_{r\in \cal R}$, where we say that $\Pi_r$ is nested in $\Pi_{r'}$, $r< r'$, if every cell (partition element) of $\Pi_r$ is properly contained in a cell of $\Pi_{r'}$. %These partitions, being nested, form a \emph{parent-child} relation where any cell in $\Pi_r$ for some $r = 2^{-i} < 1$ is a subset of a \emph{parent} cell in $\Pi_{r'}, r'= 2^{-i +1}$. 
 We denote cells in any given $\Pi_r$ by $A$, and in particular we let $A_r(x)$ denote the cell of $\Pi_r$ containing $x$. % where any fixed method might be employed to ensure such a cell is unique. 
 \end{definition}

\begin{definition}
For a set $A \in \Pi_r$, the \textbf{$r$-envelope} of the cell $A$, denoted $\tilde A$ is defined as $$ \tilde{A} \defeq \bigcup_{x^\prime \in A} B(x^\prime,r),$$ %Since we will consider the $r$-envelopes of cells at level $r$, we will write simply $\tilde{A}_r$ in this case.  
\end{definition}  so that $\tilde A_r(x)$ is the $r$-envelope of $A_r(x)$. 

Recall that we have observed $(\bX,\bY)= ((X,Y)^{n_P},(X,Y)^{n_Q})$, where $(X,Y)^{n_P} \sim P^{n_P}$ and $(X,Y)^{n_Q} \sim Q^{n_Q}$. We consider a regular classification tree defined as follows:

\begin{definition} \label{treeClassifier}
We consider \textbf{tree-based classifiers} $\hat f_r \doteq \ind\{\hat \eta_r(x) \geq 1/2\}$, for regression estimates $\hat \eta_r$ defined over levels $r$ of $T$ as follows: let \begin{align*}  \abs{\tilde A_r(x) \cap \bX} \defeq \abs{\tilde A_r(x) \cap \{X_1, \dots, X_{n_P + n_Q}\}}
\end{align*} denote the number of sample points falling in $\tilde{A}_r(x)$.  We set $\hat\eta_r(x) = 0$ if  $|\tilde{A}_r(x) \cap \bX| = 0$, otherwise we set 
\begin{equation}\label{etaHat}
\hat{\eta}_r(x) = \frac{1}{|\tilde{A}_r(x) \cap \bX|}\sum_{i=1}^n Y_i \ind\{X_i \in \tilde{A}_r(x)\}.
\end{equation}
\end{definition}

\begin{remark}
We point out that the regression estimates \eqref{treeClassifier} being defined over envelopes of the cells and the aggregate exponent \eqref{aggExpDef} being taken over grids is done so that our results will hold in the fullest generality on the marginal feature distributions $P_X, Q_X$, and obviates the need to add extra assumptions on $\cT$ or data distributions to rule out corner cases. In practice, one may opt to simply use a version of \eqref{etaHat} using the cell $A_r(x)$ rather than the envelope $\tilde A_r(x)$, as our risk guarantees will hold in this case for all but very particular cases; see Appendix \ref{notAmbientDyadic} for such an example. 
\end{remark}

%%% might be more proper to do this with cases. 

\subsection{Oracle Rates.}

Our first result is an oracle bound on the risk of such an estimator.

\begin{theorem}[Oracle Rates]\label{oracleRate}
 Let $\hat{f}_r$ be the plug-in classifier based on the dyadic tree regression \eqref{etaHat} at level $r$. For $r = \min\left\{ n_P^{-1/(2 \alpha + \alpha \beta + \gamma)}, n_Q^{-1/(2\alpha + \alpha \beta + d)}\right\}$ there is a constant $C = C (\cT)$ such that
\begin{align}\label{oracleUpperBound} 
    \sup_{(P,Q) \in \cT} \E[\cE(\hat{f}_r) ] \leq C \min\left\{ n_P^{-\tfrac{\alpha(\beta +1)}{2 \alpha + \alpha \beta + \gamma}}, n_Q^{-\tfrac{\alpha(\beta+1)}{2\alpha + \alpha \beta + d}} \right\}. 
\end{align}
%Similarly, for $r = \min\left\{ n_P^{-1/(2 \alpha + \alpha \beta + \gamma)}, n_Q^{-1/(2\alpha + d)}\right\}$ there is a constant $C^\prime = C^\prime(\cT^\prime)$ such that
%\begin{equation}\label{oracleUpperBoundQSD}
%\sup_{(P,Q) \in \cT^\prime } \E_{(\bX,\bY)}[\cE(\hat{f}_r) ] \leq C^\prime  \min\left\{ n_P^{-\tfrac{\alpha(\beta +1)}{2 \alpha + \alpha \beta + \gamma}}, n_Q^{-\tfrac{\alpha(\beta+1)}{2\alpha + d}} \right\}. 
%\end{equation}
\end{theorem}

\begin{remark} \label{no_strong_density_assumption}
To the best of our knowledge, this is the first result that gives an upper bound for the excess risk of dyadic trees, under the smoothness and noise assumptions that we have considered, that does not impose a uniform-measure assumption on the marginal distribution $Q_X$, and is therefore of interest even in the case $P_X = Q_X$. Under the standard theory with a uniform-measure assumption (that is, that $Q_X(B(x,r)) \geq C_0 r^{d}$ for all $r > 0$, $x \in \cX_Q$), we would find the optimal level to be $r^* = n_Q^{-1/(2 \alpha + d)}$. We remark that \cite{reeve2021adaptive} work under an interesting assumption that allows them to consider both uniform-type measures and measures which may not be absolutely continuous with respect to the ambient Lebesgue measure, and the rates that they give are optimal in both settings. 
\end{remark}

\begin{figure}[t]
    \centering
    \subfloat{{ \includegraphics[height = 6cm, width = 8.5cm]{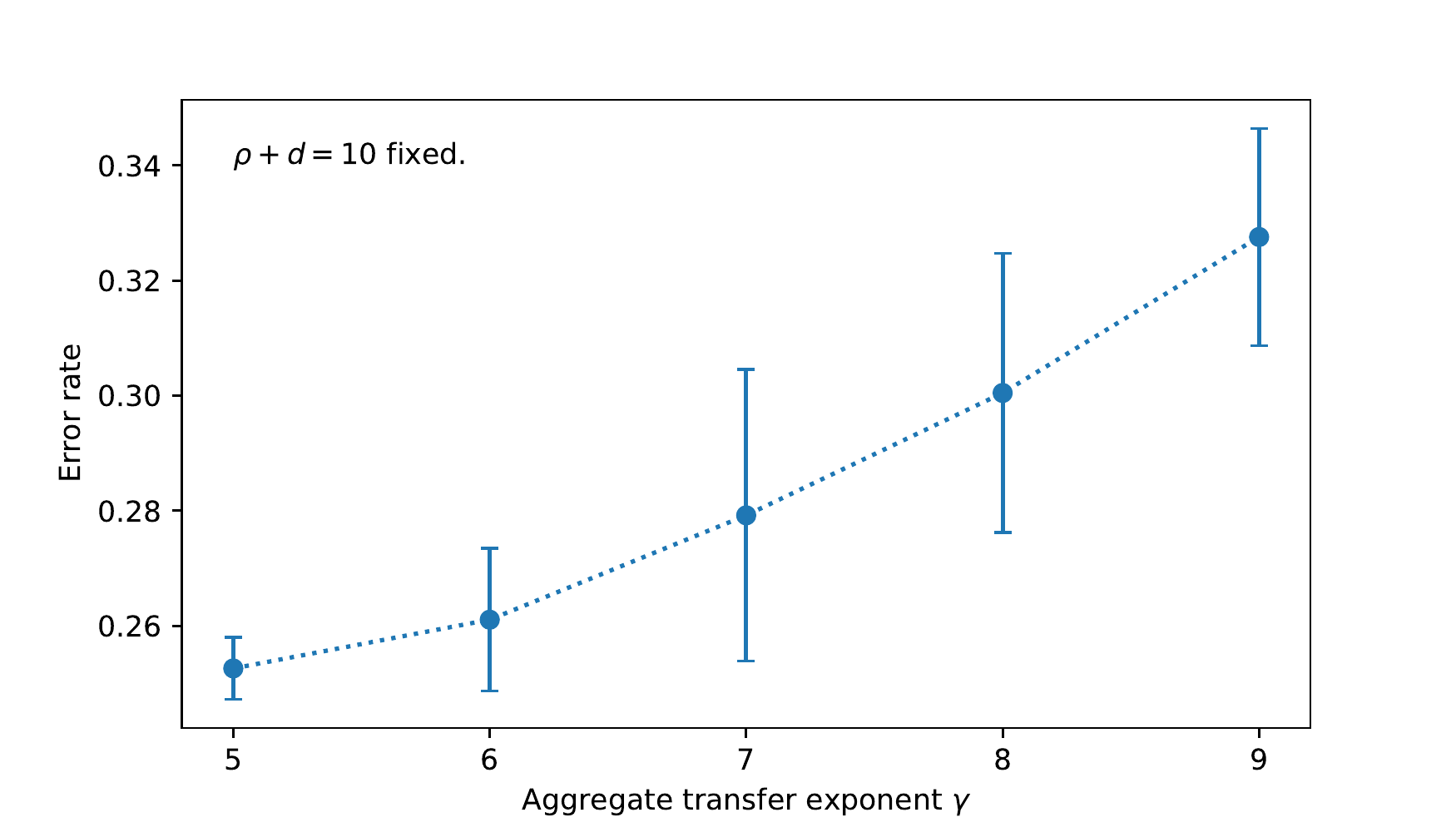} }}
    %\caption{Risk estimates for a regular dyadic tree, with depth selected by 2-fold cross-validation, for the five datasets outlined above. Error bars give the standard deviation over 10 runs. The Bayes risk is $0.18$.}
    %\label{fig:E1_experiment}
%\end{figure}
%\begin{figure}[t]
    \centering
    \subfloat{{\includegraphics[height = 6cm, width = 8.5cm]{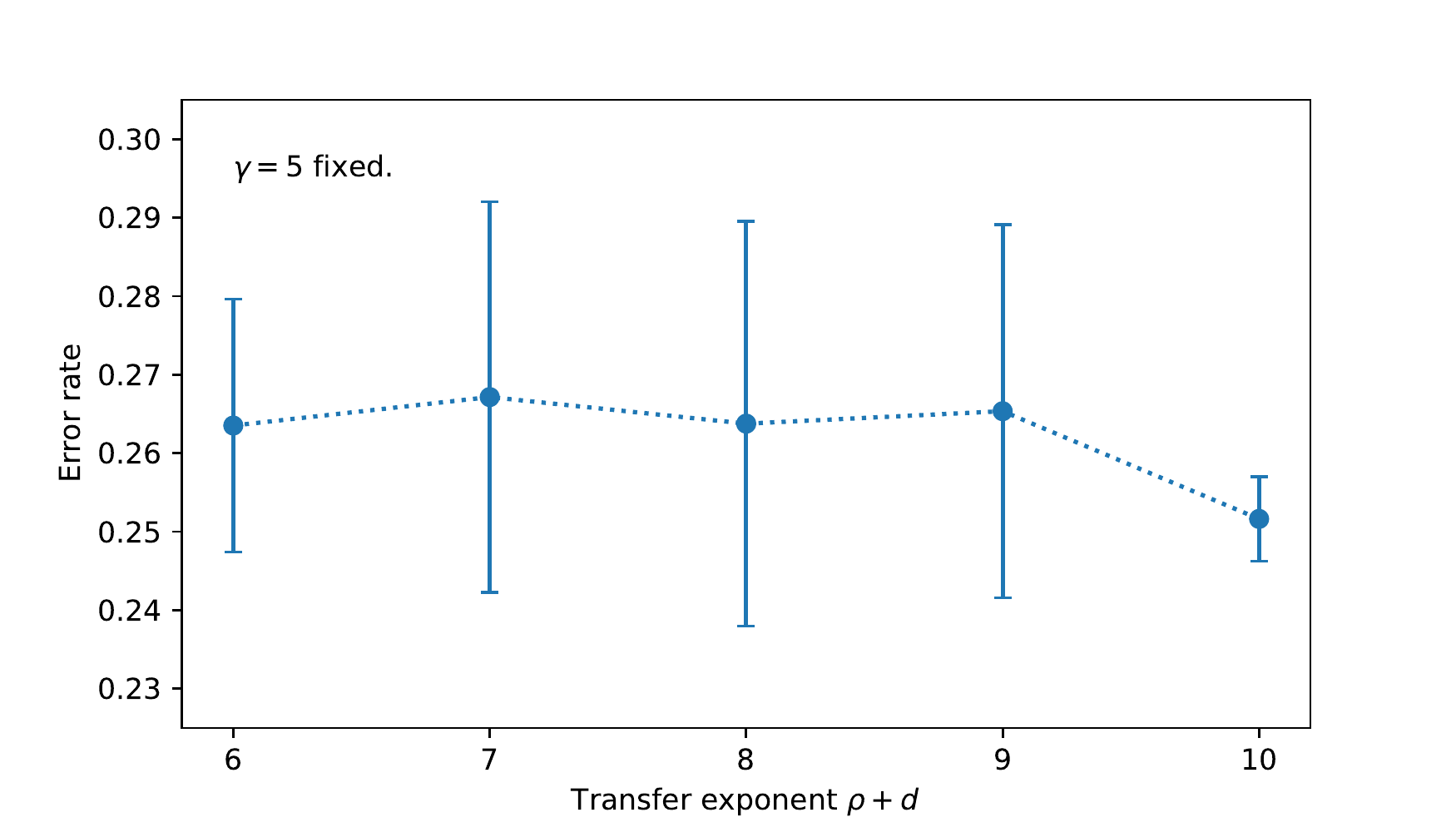} }}
    \caption{Risk estimates for a regular dyadic tree, with depth selected by 2-fold cross-validation over $100$ target samples ($n_Q = 100$), for datasets generated according to the distributions outlined in Appendix \ref{appendix_figure2_details}; we use $1000$ source samples ($n_P = 1000$). Error bars give the standard deviation over 10 runs. The Bayes risk is $0.18$. \it Top: \rm empirical error curves for five distributions with varying aggregate transfer exponents but identical exponents; the construction of the marginal distributions mirrors the construction from Figure \ref{fig:toy_ex_illustration}. \it Bottom: \rm empirical error curves for five distributions with varying transfer exponents, but with identical aggregate exponents. Again the construction from Figure \ref{fig:toy_ex_illustration} is used, except in this case the singularity strengths $\nu$ decrease as they occur over larger regions, so as to keep the aggregate transfer exponents identical across the five distributions. Taken together, these two simulations support our claim that the aggregate exponent more adequately captures the difficulty of transferring between distributions. 
    }
    \label{fig:E1_experiment}
\end{figure}

Note the role played by the quantity $\gamma$ in \eqref{oracleUpperBound}. We see that the form of the rate of convergence to zero of the excess expected risk in terms of the number of source samples is the same as the rate in terms of target samples, with the exception of the substitution of the aggregate exponent $\gamma$ for the metric dimension $d$ - this supports our interpretation of $\gamma$ as a notion of \it relative dimension, \rm an interpretation which is further justified in Section \ref{secOnDim}. Note that previously, \cite{KM} obtained analogous results for a nearest-neighbour procedure, with the same bound on the excess expected risk as in \eqref{oracleUpperBound}, but with $\rho + d$ in place of $\gamma$, that is, they found that  $\E_{(\bX,\bY)}[\cE(\hat f_r)] \leq C n_P^{- \alpha(\beta + 1)/(2\alpha + \alpha \beta + \rho + d)}$ for pairs $(P,Q)$ with a transfer exponent of $\rho$. In Section \ref{secOnDim} we demonstrate that the rate in Theorem \ref{oracleRate} is tighter by showing that $\gamma \leq \rho + d$ under some regularity conditions, and we argue that strict inequality holds here in all but edge cases. Figure \ref{fig:E1_experiment} shows the results of a simple simulation which shows that the aggregate exponent can capture the difficulty of transferring from $P$ to $Q$ while the transfer exponent fails to do so. Using the same idea as in Figure \ref{fig:toy_ex_illustration}, we construct two sequences of marginal distributions $P_X$: the top plot of Figure \ref{fig:E1_experiment} shows distributions for which the aggregate exponent increases while the transfer exponent stays fixed, while the bottom plot shows the reverse - the aggregate exponent is fixed while the transfer exponent increases. In this instance, the transfer exponent between source and target does not give a proper indication of the capacity for transfer, while we see that the aggregate exponent distinguishes a spectrum of easy-to-difficult transfer. Full details of the construction of the distributions used here are provided in Appendix \ref{appendix_figure2_details}. 

\par 

The above upper-bound is indeed tight, which follows as a Corollary to a lower-bound of \cite{KM}, stated below for completion. This follows from our Proposition \ref{sharperBounds} below, via which we can show that the class $\cT$ contains a class of distribution pairs with bounded transfer exponent, over which we invoke the lower bound of \cite{KM}. The result is as follows (the formal argument is in Appendix \ref{appendixProofs}):

\begin{proposition}\label{lowerBound}
 Let $\cT$ be fixed such that $d < \gamma$, $\alpha \beta \leq d$, and that $\gamma < \infty$. Then there is an absolute constant $C = C(\cT)$ such that for any classifier $\hat{h}$ learned on a sample from $P^{n_P} \times Q^{n_Q}$, we have
 \begin{equation*} \sup_{(P,Q) \in \cT} \E_{(\bX,\bY)} \left[ \mathcal{E}(\hat{h}) \right] \geq  C^\prime \min\left\{ n_P^{-\tfrac{\alpha(\beta +1)}{2 \alpha + \alpha \beta + \gamma}}, n_Q^{-\tfrac{\alpha(\beta+1)}{2\alpha + \alpha \beta + d}} \right\}. \end{equation*}
\end{proposition}

We note that, as pointed out by \cite{Audibert2007}, the assumption that $\alpha \beta \leq d$ is not overly restrictive, as the complementary case $\alpha \beta > d$ contains only examples in which the decision boundary does not intersect the support of the target. The minimax rate for this problem in the non-transfer setting is well-known and goes back to \cite{Audibert2007}, and their result is implied by Proposition \ref{lowerBound} by setting $n_P = 0$ (though our assumptions on the measure $Q_X$ are more general as we do not require a density). 

%\begin{figure}[t]
%    \centering
%    \includegraphics[height = 6cm, width = 9cm]{E1_fig_final.pdf}
    %\caption{Risk estimates for a regular dyadic tree, with depth selected by 2-fold cross-validation, for the five datasets outlined above. Error bars give the standard deviation over 10 runs. The Bayes risk is $0.18$.}
    %\label{fig:E1_experiment}
%\end{figure}

%\begin{figure}[t]
%    \centering
%    \includegraphics[height = 6cm, width = 9cm]{E12_fig_final.pdf}
%    \caption{Risk estimates for a regular dyadic tree, with depth selected by 2-fold cross-validation over $100$ target sample, for the five datasets outlined above. Error bars give the standard deviation over 10 runs. The Bayes risk is $0.18$. \it Top: \rm empirical error curves for five distributions with varying aggregate transfer exponents but identical exponents. \it Bottom: \rm empirical error curves for five distributions with varying transfer exponents, but with identical aggregate exponents. Taken together, these two simulations support our claim that the aggregate exponent more adequately captures the difficulty of transferring between distributions. }
%    \label{fig:E1_experiment}
%\end{figure}

%UNCOMMENT FOR ONE-COLUMN FORMAT:

\subsection{Adaptive Rates.}

%add remark about why s > gamma rather than simply gamma. 
%actually don't have a theorem for the Tprime case, so that needs to be typed up

Our next result concerns an adaptive procedure -  that is, a procedure that can be implemented without prior knowledge of the structural parameters $(\alpha,\beta,\gamma,d)$ - based on the dyadic tree from which we derived the oracle rate. We demonstrate that this method can attain the optimal rate of Theorem \ref{oracleRate} up to a logarithmic factor. This procedure is based on the Intersecting Confidence Intervals method of \cite{ICIGoldNem}, which features in the work of Lepski (e.g. \cite{Lepski1997}). The key idea is given a concise exposition in \cite{WassNonparametrics}. 

The method proceeds by locally constructing confidence bounds for $\eta(x) = P(Y = 1 \mid X = x)$ based on the data-dependent quantity  
\begin{equation}\label{sigHat1}
     \hat{\sigma}_r(x) \defeq \frac{C_{n_P, n_Q}}{\sqrt{|\tilde A_r(x) \cap \bX |}}, \quad \mathrm{where } \quad C_{n_P, n_Q} = \frac{1}{2}\left(1 + 2\sqrt{\log (n_P + n_Q)}\right), 
\end{equation} an estimate of the local variance in the cell $A_r(x)$ and serves to upper bound the error $\abs{\hat \eta_r(x) - \tilde \eta_r(x)}$, where $\tilde \eta_r(x) = \E \left[ \hat \eta_r(x) \mid \bX \right]$; note that the quantity $\hat{\sigma}_r(x)$ requires no problem-specific information to be computable. The procedure begins at a leaf node, and moves up the branch containing $x$ until a level is reached that achieves an approximate balance of the local variance and bias.
Details are given in Algorithm \ref{lepskiAlg} below, and the result is as follows: 

%% eta hat and sigma hat are only defined in the proofs section, so this has to be introduced. 

\begin{theorem}[Adaptive Rates]\label{adaptiveTheorem}
Let $\hat{f}$ denote the classifier determined by Algorithm 1. Then there is a universal $C = C(\cT)$ not depending on $n_P,n_Q$ such that

%UNCOMMENT FOR TWO-COLUMN FORMAT:
%\begin{align*} & \sup_{(P,Q) \in \cT}  \E_{(\bX,\bY)}[\cE(\hat{f})] \leq  \\ & \quad C  \min\left\{ \left(\frac{\log n_P }{n_P}\right)^{\tfrac{\alpha(\beta +1)}{2 \alpha + \alpha \beta + \gamma}}, \left(\frac{\log  n_Q }{n_Q}\right)^{\tfrac{\alpha(\beta+1)}{2\alpha + \alpha \beta + d}} \right\}. \end{align*} 

%UNCOMMENT FOR ONE-COLUMN FORMAT
\begin{align*} \sup_{(P,Q) \in \cT}  \E_{(\bX,\bY)}[\cE(\hat{f})] \leq   C  \min\left\{ \left(\frac{\log n_P }{n_P}\right)^{\tfrac{\alpha(\beta +1)}{2 \alpha + \alpha \beta + \gamma}}, \left(\frac{\log  n_Q }{n_Q}\right)^{\tfrac{\alpha(\beta+1)}{2\alpha + \alpha \beta + d}} \right\}. \end{align*}

%and a $C^\prime = C^\prime(\cT^\prime)$ such that 
%\begin{align*} \sup_{(P,Q) \in \cT} & \E_{(\bX,\bY)}[ \cE(\hat{f})] \leq \\ & C  \min\left\{ \left(\frac{\log n_P }{n_P}\right)^{\tfrac{\alpha(\beta +1)}{2 \alpha + \alpha \beta + \gamma}}, \left(\frac{\log  n_Q }{n_Q}\right)^{\tfrac{\alpha(\beta+1)}{2\alpha + d}} \right\}. \end{align*} 
\end{theorem}

As far as we are aware, Theorem \ref{adaptiveTheorem} is the first adaptivity result for decision tree classifiers in the covariate-shift setting. Comparable rates have been derived for K-nearest neighbour procedures, as well as kernel methods, in \cite{KM} and \cite{pathak2022new}. In the one-sample setting, \cite{DDT} have derived adaptive rates under conditions related to ours, but due to differing assumptions are not directly comparable. Their method, based on structural risk minimization, does not seem to extend naturally to the problem of transfer learning (as we argue in Appendix \ref{appendix_experiments}).

\begin{algorithm}[ht]
\caption{Local Depth Selection.} 
\label{lepskiAlg} % and a label for \ref{} commands later in the document
%\centering 
\begin{algorithmic} % enter the algorithmic environment
    %(with $\hat \sigma_r$ as in \eqref{sigHat1})
    \REQUIRE A point $x \in \cX_Q$, and a depth $r_0$.
    \ENSURE A classification estimate $\hat{f}(x) = \ind\{ \hat{\eta}(x) \geq 1/2\}$.
    \BlankLine
    \STATE Set $r = r_0$. Set $\hat{\eta}^+ = \hat{\eta}_r(x) + 2 \hat{\sigma}_r(x)$ and $\hat{\eta}^- = \hat{\eta}_r(x) - 2 \hat{\sigma}_r(x)$; set $\hat{\eta} = \hat{\eta}_r(x)$.  
    \WHILE{$\hat{\eta}^- \leq 1/2$,$\hat{\eta}^+ \geq 1/2$, $r < 1/2$}
        \STATE $r \leftarrow 2r$, 
        \STATE $\hat{\eta}^- \leftarrow \max\{ \hat{\eta}^-, \hat{\eta}_r(x) - 2 \hat{\sigma}_r(x)\},$ 
        \STATE $\hat{\eta}^+ \leftarrow \min\{ \hat{\eta}^+, \hat{\eta}_r(x) + 2 \hat{\sigma}_r(x)\}$, 
        \IF{ $\hat{\eta}^{+} \leq \hat{\eta}^-$, }
            \STATE \textbf{stop}
            \STATE $\hat{\eta} \leftarrow (\hat{\eta}^+ + \hat{\eta}^-)/2$
        \ENDIF
        \STATE $\hat{\eta} \leftarrow \hat{\eta}_{r}$
    \ENDWHILE
    \RETURN $\hat{f}(x) = \ind\{ \hat{\eta}(x) \geq 1/2\}$. 
\end{algorithmic}
\end{algorithm}

\subsection{Remarks on Algorithm \ref{lepskiAlg}.} \label{computational_considerations}

\paragraph{Algorithmic Intuition: ICI}

Algorithm \ref{lepskiAlg} is an instance of the so-called Intersecting Confidence Interval methods, which operate on the basis of the following observation: for small $r$ (that is, at nodes that are deep in the tree), the regression error at a point $x$ will roughly satisfy $|\hat \eta_r(x) - \eta(x)| \approx \hat \sigma_r(x) + r \leq 2 \hat \sigma_r(x)$. Therefore, the interval $\hat \eta_r \pm 2 \hat \sigma_r$ contains the truth, $\eta$, for all levels $r$ up to $r^*$, where $r^*$ denotes the level at which an approximate balance is achieved between bias and variance. It remains only to argue that the regression error at $\hat r$, the level selected by the algorithm, cannot be far from the error at level $r^*$, and therefore that the classification loss suffered by the classifier output by Algorithm 1 cannot be much different from that suffered by the oracle dyadic tree; see Lemma \ref{lepskiKeyLemma} for the formal statement.

\paragraph{Practical Implementation}

Two brief remarks on the implementation of Algorithm \ref{lepskiAlg} are in order. Firstly, we note that the value given in \eqref{sigHat1} for $C_{n_P, n_Q}$, which comes from taking a union bound over events in each cell of the tree, is in practice much too loose. In practice it was found that setting $C = 1/4$, so that $\hat \sigma$ remains an upper bound for the variance within a cell, yielded stable performance (see Section \ref{secExperiments} for more comments). Secondly, while Theorems \ref{adaptiveTheorem} and \ref{localized_adaptive_theorem} apply to the regular dyadic trees of Definition \ref{dyadicTree}, the analysis is easily extended (see Corollary \ref{cyclicalTreeRates}) \emph{cyclical} dyadic trees \cite{DDT}, in which each node of the tree is successively split into two nodes along each of the $d$ coordinate axes in turn (so that a regular cube of side length $2^{-m}$ is split into the $2^d$ cubes of side length $2^{-(m+1)}$ after $d$ levels). All of the experiments in Section \ref{secExperiments} are carried out on such trees.   

\paragraph{Prediction Time} Although many minimax-optimal methods for classification and regression using the Intersecting Confidence Interval method \cite{ICIGoldNem} have been proposed, we note that the underlying dyadic-tree structure makes Algorithm \ref{lepskiAlg} particularly appealing due to its low computational cost at test-time. Once the tree is initialized, for a given input point $x$, it takes at most $- \log_2 r_0 = O(\log (n))$ steps to compute $\hat f (x)$, where $n$ is the combined sample size. The initialization of the tree can be achieved in $O(n \log n)$ time as well (see e.g. \cite{ODDT, hush2010algorithms}), and for a further cost of $O(n \log n )$ operations, the output $\hat f(x)$ can be computed for all further $x$ at once, reducing the burden to $O(1)$ operations at test time. On the other hand, using the ICI method with a nearest-neighbour procedure (as in \cite{KM, reeve2021adaptive}) implies locally finding $k(x)$ neighbours for each query point $x$ at testing time, which requires at least $O(n \log n )$ operations since $k(x)$ will be at least a root of $n$. We note also that once the base tree is initialized, it can be updated in online fashion at cost only $O(\log n)$ for each new data point, since to update the tree to include a new point $(y, x)$ requires only one pass along the branch that ends in the leaf containing $x$.

\subsection{Main Result: Localizing to Decision Boundary}\label{Sec_localizingToBoundary}

Below we consider how to refine the rates presented in the previous section. Throughout, we suppose that we have measures $(P,Q) \in \cT$, where $\cT = \cT(\alpha, C_\alpha, \beta, C_\beta, \gamma^*, C_{\gamma^*}, d, C_d)$. Note that in this section we use $\gamma^*$ to denote the aggregate transfer exponent of $P_X$ relative to $Q_X$ in order to distinguish it from local exponents defined below. \par To begin, recall that we suppose $\cX_P, \cX_Q \subset \cX$, and for $\epsilon > 0$ let us define the decision boundary margin region $\cG_\epsilon^- \defeq \{x \in \cX: \, \abs{\eta(x) - \tfrac{1}{2}} \leq \epsilon\}$ and $\cG_\epsilon^+ \defeq \{x \in \cX: \, \abs{\eta(x) - \tfrac{1}{2}} > \epsilon\}$. The first step is to define an aggregate transfer exponent restricted to $\cG_\epsilon^-$, which contains the decision boundary.  

\begin{definition}
Let $(P,Q) \in \cT$. The aggregate transfer exponent \textbf{localized} to the region $\cG_\epsilon^-$, denoted $\gamma(\epsilon)$, is defined to be the least positive number such that for all $r$-grids $\xi(\epsilon)$ of $\cX_Q \cap \cG_\epsilon^-$ we have 
$$ \sum_{E \in \xi(\epsilon)} \frac{Q_X(E)}{P_X(E)} \leq C_{\gamma^*} r^{-\gamma(\epsilon)}$$ for all $0 < r < 1$, where $C_{\gamma^*}$ is the parameter from $\cT$. 
\end{definition}

Our first result is an excess risk bound for an oracle classifier based on an arbitrary and fixed margin value.  
\begin{theorem} \label{oracleEpsilonMarginBound}
There is a universal constant $C = C(\cT)$ such that, for any $\epsilon > 0$, there is a classifier $\hat f_\epsilon$ such that

%UNCOMMENT FOR ONE-COLUMN FORMAT:
\begin{align}\label{epsilon_arbitrary_bound}
    \E \, \cE( \hat f_\epsilon)  \leq C \left[\min\left( n_P^{-\tfrac{\alpha(\beta +1)}{2 \alpha + \alpha \beta + \gamma(\epsilon)}}, n_Q^{-\tfrac{\alpha(\beta+1)}{2\alpha + \alpha \beta + d}} \right) \right. \left. + \min\left(  \frac{1}{n_P} \epsilon^{-(1 + \gamma^*/\alpha)}, \frac{1}{n_Q} \epsilon^{-(1 + d/\alpha)} \right)  \right]. 
\end{align}

%UNCOMMENT FOR TWO-COLUMN FORMAT:
%\begin{align}\label{epsilon_arbitrary_bound}
%    \E \, \cE( \hat f_\epsilon) & \leq C \left[\min\left( n_P^{-\tfrac{\alpha(\beta +1)}{2 \alpha + \alpha \beta + \gamma(\epsilon)}}, n_Q^{-\tfrac{\alpha(\beta+1)}{2\alpha + \alpha \beta + d}} \right) \right. \nonumber \\& \quad \left. + \min\left(  \frac{1}{n_P} \epsilon^{-(1 + \gamma^*/\alpha)}, \frac{1}{n_Q} \epsilon^{-(1 + d/\alpha)} \right)  \right]. 
%\end{align}

\end{theorem}

Our main result demonstrates that, up to log terms, the above oracle bound is attained by the adaptive classifier of Algorithm \ref{lepskiAlg}, \emph{for an optimal value of the margin parameter $\epsilon$}. Choosing $\epsilon$ to approximately balance the terms in \eqref{epsilon_arbitrary_bound} (adjusted to include log-terms) gives the finite-sample rate that is achievable at sample size $n  = (n_P, n_Q)$, determined by the behaviour of the aggregate exponent in the vicinity of the decision boundary. 

%%%%%\epsilon_n^* = \inf \left\{\epsilon > 0: \epsilon \geq  \left(\frac{\log n_P}{n_P}\right)^{\tfrac{\alpha (\alpha + \gamma(\epsilon))}{(\alpha + \gamma^*)(2\alpha + \alpha \beta + \gamma(\epsilon))}}\right\},

\begin{theorem}[Adaptive Localized Rate] \label{localized_adaptive_theorem}
Consider the following critical level \begin{equation*}
    \epsilon_n = \inf \left\{\epsilon > 0: \frac{1}{n_P} \epsilon^{-(1 + \gamma^*/\alpha)} \leq \left(\frac{\log n_P}{n_P}\right)^{\tfrac{-\alpha(\beta+1)}{2\alpha + \alpha \beta + \gamma(\epsilon)}}\right\},
\end{equation*} and let $\gamma_n = \gamma(\epsilon_n)$. Let $\hat f$ denote the adaptive classifier of Algorithm \ref{lepskiAlg}, and suppose that $\alpha + d > 1$. Then there are constants $C_1,C_2$, depending only on $\cT$, such that the following holds. If $\max(n_P,n_Q) \geq C_1$, then for any $\epsilon  > 0$ we have 

%UNCOMMENT FOR ONE-COLUMN FORMAT:
\begin{align*}  \E_{(\bX,\bY)}[\cE(\hat{f})] \leq 2 C  \cdot \min\left\{ \left(\frac{\log n_P }{n_P}\right)^{\tfrac{\alpha(\beta +1)}{2 \alpha + \alpha \beta + \gamma_n}}, \left(\frac{\log  n_Q }{n_Q}\right)^{\tfrac{\alpha(\beta+1)}{2\alpha + \alpha \beta + d}} \right\},
\end{align*}

%UNCOMMENT FOR TWO-COLUMN FORMAT:
%\begin{align*} & \E_{(\bX,\bY)}[\cE(\hat{f})] \\& \quad \leq 2 C  \cdot \min\left\{ \left(\frac{\log n_P }{n_P}\right)^{\tfrac{\alpha(\beta +1)}{2 \alpha + \alpha \beta + \gamma_n}}, \left(\frac{\log  n_Q }{n_Q}\right)^{\tfrac{\alpha(\beta+1)}{2\alpha + \alpha \beta + d}} \right\},
%\end{align*} 

with $C$ as in Theorem \ref{oracleEpsilonMarginBound}. 
\end{theorem}

\begin{remark}
Note that the critical margin level $\epsilon_n$ does not depend on $n_Q$. This is because the level is chosen to minimize the source-data contribution to the intermediate bound \eqref{epsilon_arbitrary_bound}.
%(that is, the first arguments to the two minimums in \eqref{highMargin}) - the rate in terms of target samples is not affected by the value of $\epsilon_n$.  
\end{remark}

From the above, we see that if the relative dimension of $P$ to $Q$ is small on a low-margin region (that is, close to the decision boundary), then for sufficiently large sample size, the expected excess risk will converge to zero at a rate determined by this localized value of the relative dimension.

\begin{remark} Suppose that for some $\gamma > 0$ there is a corresponding $\epsilon_0 > 0$ such that $\gamma(\epsilon_0) = \gamma$. Then, it is immediate by rearranging that for 
\begin{equation}\label{requiredSamples1}
    \frac{n_P}{\log n_P} \geq \epsilon_0^{-\left( \tfrac{2\alpha + \alpha \beta + \gamma}{\alpha(\alpha + \gamma)}\right)(\alpha + \gamma^*)},
\end{equation} we have $$ \frac{1}{n_P}\epsilon_0^{-(1 + \gamma^*/\alpha)} \leq \left( \frac{\log n_P}{n_P} \right)^{\tfrac{\alpha(\beta +1)}{2 \alpha + \alpha \beta + \gamma(\epsilon_0)}},$$ which by construction implies that $\epsilon_n \leq \epsilon_0$ from which we get $\gamma_n = \gamma(\epsilon_n) \leq \gamma(\epsilon_0) = \gamma$, since $\gamma(\cdot)$ is non-decreasing. It then follows from Theorem \ref{localized_adaptive_theorem} that we have 

%UNCOMMENT FOR ONE-COLUMN FORMAT
\begin{align} \E_{(\bX,\bY)}[\cE(\hat{f})] \label{gammarate} \leq C  \min\left\{ \left(\frac{\log n_P }{n_P}\right)^{\tfrac{\alpha(\beta +1)}{2 \alpha + \alpha \beta + \gamma}}, \left(\frac{\log  n_Q }{n_Q}\right)^{\tfrac{\alpha(\beta+1)}{2\alpha + \alpha \beta + d}} \right\},
\end{align}

%UNCOMMENT FOR TWO-COLUMN FORMAT
%\begin{align} & \E_{(\bX,\bY)}[\cE(\hat{f})] \nonumber  \\& \label{gammarate} \leq C  \min\left\{ \left(\frac{\log n_P }{n_P}\right)^{\tfrac{\alpha(\beta +1)}{2 \alpha + \alpha \beta + \gamma}}, \left(\frac{\log  n_Q }{n_Q}\right)^{\tfrac{\alpha(\beta+1)}{2\alpha + \alpha \beta + d}} \right\},
%\end{align} 

where $C$ is as in Theorem \ref{localized_adaptive_theorem}. This illustrates the following behaviour: supposing that for some margin value $\epsilon_0$ we have a low value $\gamma(\epsilon_0) = \gamma < \gamma^*$, then when enough source samples have been collected, the expected excess risk will decrease to zero at the faster rate \eqref{gammarate}. Until this threshold has been achieved however, the convergence will be at the slower rate, determined by $\gamma^*$, and unfortunately, the requisite sample size is exponential in the global aggregate transfer exponent $\gamma^*$. However, we emphasize that with stronger assumptions on the behaviour of $P_X, Q_X$ away from the decision boundary this would no longer be the case - see for example Proposition 3.7 of \cite{Audibert2007}, where it is shown that exponential convergence rates can be derived for error away from the boundary under the strong density condition. 
\end{remark}

\begin{example}
The following construction in one dimension is an example of a case in which the aggregate transfer exponent becomes smaller when localized to the decision boundary, giving a faster rate of convergence. Suppose that $d = 1$, $\cX_Q = \cX_P = [0,1]$, and take $Q_X \sim U([0,1])$. Let $P_X$ have density $p(x) \propto x^{\gamma}$ for some 
$\gamma > 1$, and let $\eta(x) = x$. One can show that that $(P,Q)$ are in the transfer class with parameters $\alpha = 1$, $\beta = 1$, and $\gamma^* = \gamma$, and so Theorem \ref{adaptiveTheorem} tells us that 

\begin{align} \label{non_local_rate}
    \E[ \cE(\hat f)] &\leq C \left( \frac{\log n_P}{n_P} \right)^{\tfrac{2}{3 + \gamma}}.
\end{align} On the other hand, in this example the singularity in the density ratio occurs away from the decision boundary, and one readily finds for any $\epsilon < 1/2$, we have $\gamma(\epsilon) = 1$, and so by the argument given above we find that for $n_P/\log n_P \geq 4^\gamma$ we have 
\begin{align*}
    \E[ \cE(\hat f)] \leq C \left( \frac{\log n_P}{n_P} \right)^{\tfrac{1}{2}},
\end{align*} a faster rate than \eqref{non_local_rate}. 
\end{example}

%% LOWER BOUNDS MATERIAL REMOVED FROM MAIN BODY:
%\subsection{Lower Bounds.}

%Finally, we show that the oracle rates of Theorem \ref{oracleRate} cannot be improved in general as they admit matching lower-bounds that hold for any classifier, over nontrivial ranges of distributional parameters. 

%\begin{theorem}\label{lowerBoundTheorem}
% Let $\cT, \cT^\prime$ be fixed such that $d < \gamma$, $\alpha \beta \leq d$, and that $\gamma < \infty$. Then there are absolute constants $C = C(\cT)$ and $C^\prime = C^\prime(\cT^\prime)$ such that for any classifier $\hat{h}$ learned on a sample from $P^{n_P} \times Q^{n_Q}$, we have
% \begin{equation*} \sup_{(P,Q) \in \cT} \E_{(\bX,\bY)} \left[ \mathcal{E}(\hat{h}) \right] \geq  C^\prime \min\left\{ n_P^{-\tfrac{\alpha(\beta +1)}{2 \alpha + \alpha \beta + \gamma}}, n_Q^{-\tfrac{\alpha(\beta+1)}{2\alpha + \alpha \beta + d}} \right\}, \end{equation*} and 
 
% \begin{equation*} \sup_{(P,Q) \in \cT^\prime}   \E_{(\bX,\bY)} \left[ \mathcal{E}(\hat{h}) \right] \geq C^\prime \min\left\{ n_P^{-\tfrac{\alpha(\beta +1)}{2 \alpha + \alpha \beta + \gamma}}, n_Q^{-\tfrac{\alpha(\beta+1)}{2\alpha + d}} \right\}. \end{equation*}
% \end{theorem}

%We note that, as pointed out by \cite{Audibert2007}, the assumption that $\alpha \beta \leq d$ is not overly restrictive, as the complementary case $\alpha \beta > d$ contains only examples in which the decision boundary does not intersect the support of the target.  

%not nec. a trivial example, so reword. 

\section{Relation to Notions of Dimension.}\label{secOnDim}

In this section we will consider how the aggregate transfer exponent defined in  \eqref{globalTransferExp} relates to traditional notions of dimension of a measure. In particular, it is immediate (as argued below) that, in the limits, $\gamma_* \doteq \inf \{\gamma\}$ (see Definition \ref{relDim}) can be viewed as a relative version of the Minkowski dimension. Similarly, we can argue that, under regularity conditions, $\gamma_*$ is equivalent to an \emph{average} version of 
the \emph{transfer exponent} of \cite{KM}. This average version in turn defines a relative version  
of the Renyi dimension. 
\par Note that the ensuing discussion applies to general measures supported on a common metric space $(\cX,d)$, with open balls of radius $r$ denoted $B(\cdot,r)$, although the rest of our work considers chiefly the case $\cX = [0,1]^D$ with the $\ell_\infty$ metric. We use $P_X,Q_X$ throughout this section to denote generic measures supported on $\cX_P,\cX_Q \subset \cX$ respectively.

%though it should be remembered that the quantity that determines the excess risk rate is determined by the relative dimension of the feature marginals $P_X,Q_X$. \par 

\begin{definition}\label{relDim}  For measures $P_X,Q_X$, take $\Xi_r$ to be the set of all $r$-grids of $\cX_Q$ in $\cX$, and define the \textbf{minimal aggregate transfer exponent} $\gamma_*= \gamma_*(P_X, Q_X)$ as follows: \begin{align}\label{gammaStar}
\gamma_*  \doteq \inf_{\gamma \geq 0} \left\{\, \exists C_\gamma \hspace{.1cm} \text{\rm s.t.} \, \sum_{E \in \xi} \frac{Q_X(E)}{P_X(E)} \leq C_\gamma r^{-\gamma} \: \mathrm{for}\, \mathrm{all} \, \xi \in \Xi_r\right\}. 
\end{align}
%We call $\gamma_*(P_X,Q_X)$ the \emph{relative dimension} (or, relative \emph{Minkowski} dimension) between a source $P_X$ and a target $Q_X$.
\end{definition}

Observe that one can express all of the main results in terms of $\gamma_*(P_X, Q_X)$, so that, for instance, Theorem \ref{oracleRate} implies that for any $s > \gamma_*(P_X,Q_X)$, the excess risk as a function of the source samples converges to zero at a rate no worse than $n_P^{-\alpha(1 + \beta)/(2 \alpha + \alpha \beta + s)}$. 
 
As it turns out, the minimal $\gamma_*$ can be viewed as a relative version of the Minkowski dimension of a measure, which is defined as follows (see \cite{Pesin1997}). % we use the definition from \cite{weed2019}).

\begin{definition} 
 Let $S \subset \cX_Q$. The \textbf{Minkowski dimension} of $S$, denoted $\dim_M(S)$, is defined as 
 \begin{equation}\label{MinkowskiDimDef} 
 \dim_M(S) = \limsup_{r \to 0} \frac{\log \mathcal{N} (S, r)}{- \log r},
 \end{equation} and the Minkowski dimension of the measure $Q_X$ is defined via 
 \begin{align*}%\label{minkowskiDimDef}
 \dim_M(Q_X) = \inf\{ \dim_M(S): Q_X(S) = 1\}. 
\end{align*}  
\end{definition} 

Note that (see \cite{Falconer2014}, Section 3.1) one obtains an equivalent definition of the Minkowski dimension if the covering number $\cN(S,r)$ in \eqref{MinkowskiDimDef} is replaced by the maximal packing number, which in turn is easily seen to equal the maximal size of an $r$-grid. The following proposition immediately follows from this observation.  

\begin{proposition}[$\gamma_*$ is relative Minkowski] 
We have $$\gamma_*(Q_X, Q_X) = \dim_M(Q_X).$$ 
\end{proposition}
\begin{proof}
Let $\mathcal{M}(\cX_Q, r)$ denote the $r$-packing number of $\cX_Q$, and let $\Xi$ denote the set of $r$-grids of $\cX_Q$. Since an $r$-grid gives an $r$-packing, and an $r$-packing can be transformed into a grid by using a Voronoi tessellation, we have 
\begin{align*}
    \sup_{\xi \in \Xi} \sum_{E \in \xi} \frac{Q_X(E)}{Q_X(E)} & = \sup_{\xi \in \Xi} \vert \xi \vert 
    = \mathcal{M}(\cX_Q, r),
\end{align*} and the result follows since \eqref{MinkowskiDimDef} yields the same value for $\dim_M$ if the packing number $\mathcal{M}$ is used in place of the cover number $\mathcal{N}$.  
\end{proof}

\subsection{Relation to transfer exponent.} \label{sec_relation_to_transfer_exponent}
 
 Recall the \emph{transfer exponent} of \cite{KM}, introduced in Definition \ref{transferExp0}. We now consider an alternative way to obtain an average version of this notion. As we show thereafter, this alternative is, under some conditions, equivalent to the aggregate transfer exponent introduced in Definition \ref{aggExpDef}. 

\begin{definition}\label{defTransferIntegral}
We call $\bar \rho > 0$, an {\bf integrated transfer exponent} from $P_X$ to $Q_X$, if there exists $C_{\bar \rho} > 0$ such that 
\begin{equation}\label{transferIntegral}
 \forall r > 0, \quad \int_{\cX} \frac{1}{P_X(B(x,r))} \, Q_X(dx) \leq C_{\bar \rho} \cdot r^{-\bar \rho}.
\end{equation}

We will be interested in the \textbf{minimal integrated transfer exponent} $\bar \rho$: 
\begin{equation*}\label{rhoBarStar}
\bar \rho_* = \inf\{ \bar \rho>0: \exists \,  C_{\bar \rho} \, \text{ such that } \, \eqref{transferIntegral} \text{ holds.}\}.\end{equation*}
\end{definition}

Note that the integrated transfer exponent \eqref{transferIntegral} was recently introduced and investigated by \cite{pathak2022new}, who use it to derive excess risk bounds similar to our Theorem \ref{oracleRate} in the context of nonparametric regression under covariate shift (they do not consider classification). The following proposition gives a simple relation between $\bar \rho_*$ and the minimal transfer exponent $\rho_*$. The implication of this result is discussed at the end of the section. 

\begin{proposition}\label{sharperBounds}
Let $\rho_*$ be the minimal transfer exponent from $P_X$ to $Q_X$, and let $\bar \rho_*$ be the minimal integrated transfer exponent from $P_X$ to $Q_X$. Suppose $Q_X$ is of metric dimension $d$. We then have that $\bar \rho_* \leq \rho_* + d$.
\end{proposition}

Note that Proposition \ref{sharperBounds} is a straightforward extension of Lemma 1 from \cite{pathak2022new}. Example \ref{E1} below shows that we can have the strict inequality $\bar \rho_* < \rho + d$. As it turns out, $\bar \rho_*$ can be interpreted as a relative form of the \emph{Renyi dimension}, similar to the interpratation of $\gamma_*$ in terms of Minkowski dimension. This is immediate from the following definition (see \cite{Olsen2005}): 
\begin{definition}
The \textbf{Renyi dimension} (of order zero) of a measure $Q_X$, $\dim_R(Q_X)$, is given by \begin{equation*}
    \dim_R(Q_X) = \limsup_{r \to 0} \frac{ \log \int Q_X(B(x,r))^{ - 1} \, Q_X(dx)}{- \log r}.
\end{equation*}
\end{definition}

One immediately sees that the minimal integrated transfer exponent $\bar \rho_*(Q_X,Q_X)$ corresponds to the Renyi dimension of order zero, $\dim_R(Q_X)$. It is known that, for any measure $Q_X$, $\dim_R(Q_X)$ is equal to the Minkowski dimension $\dim_M(Q_X)$ (see \cite{Pesin1997}); stated differently, we have 
$\bar \rho_* (Q_X, Q_X) = \gamma_*(Q_X, Q_X)$. In what follows, we will show that in the case of two measures $P_X,Q_X$, under some regularity condition we also have that $\bar \rho_* (P_X, Q_X) = \gamma_*(P_X, Q_X)$ (Proposition \ref{relDimEquivalence}). 
We start with the following condition (\cite{Pesin1997}, \cite{CoincidenceOfDims}) on the measures which in particular allow us to relate conditions on \emph{grids} in the definition of $\gamma_*$ to conditions on balls in the definition of $\bar \rho_*$.

\begin{definition}
A measure $\mu$ is said to be \textbf{doubling} with constant $C$ if for all $r > 0$ and all $x \in \mathrm{supp}(\mu)$, we have 
\begin{equation}\label{doubling}
    \mu(B(x,r)) \geq C \mu(B(x,2r)).
\end{equation}
\end{definition}

The simplest examples of doubling measures are measures supported on well-behaved compact subsets $\bR^d$ with densities that are bounded above and away from zero with respect to the Lebesgue measure - a regularity condition that is frequently imposed in the nonparametrics literature, for instance in the seminal works \cite{stone1982} and \cite{Audibert2007} (in the latter this is the \emph{strong-density} assumption). The strong-density condition is not necessary however; for example, 
%the Gibbs measure on $\mathbb{R}$ given by $P(dx) = 
%\tfrac{1}{Z_f} \exp\{- \beta f(x)\} \, dx $ is a doubling %measure whenever $f$ is H\"older continuous, as is 
the measure on $[0,1]$ with density given by $p(x) \propto x^\nu \, dx$ for any $\nu > 0$ is doubling. The doubling condition is also commonly seen in the literature on dimensions of measures, where a measure satisfying \eqref{doubling} is sometimes referred to as a \emph{Federer} measure, or \emph{diametrically regular} \cite{Pesin1997}; this work also lists larger classes of doubling measures, and further examples are provided by \cite{SteinBook}. The survey of nearest-neighbour methods \cite{clarkson2006nearest} presents connections between neighbour searching and notions of metric space dimension that includes a discussion of doubling measures, and may therefore be of interest. 

Under a doubling assumption, we can establish the following result, complementing Proposition \ref{sharperBounds}:

\begin{proposition}\label{rhoStarLowerBound}
 Suppose that the minimal transfer exponent satisfies $\rho^*(P_X,Q_X) \geq \rho$. If $Q_X$ is a doubling measure, then any integrated transfer exponent $\bar \rho$ satisfies $\bar \rho \geq \rho$.
 \end{proposition}
 
Note that the doubling assumption allows us to relate the measures of sets that are of comparable size; this assumption could be dropped if, say, we were to define the transfer exponent condition as needing to hold for all sets of aspect ratio bounded by some constant $\phi > 1$ rather than all balls (that is to say, if we required that $P_X(A) \geq C \, Q_X(A) r^{\rho}$ whenever there exists an $x \in \cX_Q$ such that $B(x,r) \subset A \subset B(x,\phi r)$).

%underlines some of the connections between the field of geometric measure theory and discusses the particular role of doubling measures, and is therefore of some interest.  
%%% cite some examples. Whole classes provided in Pesin '97. 
%alter this a bit. 

We have the following relation.

\begin{proposition}\label{relDimEquivalence}
Let $\gamma_*, \bar \rho_*$ be the minimal aggregate and integrated transfer exponents from $P_X$ to $Q_X$, respectively. If $P_X,Q_X$ are doubling measures, then $\gamma_* = \bar \rho_*$. \end{proposition}

The condition that $P_X,Q_X$ be doubling measures is sufficient but not necessary; in fact the equivalence can be shown to hold in greater generality, though we do not pursue this here. We note however that a simpler definition of the aggregate transfer exponent, in which the sum in \eqref{globalTransferExp} is taken over a regular dyadic partition of $[0,1]^D$ rather than over grids, can give a different value for the relative dimension. In Appendix \ref{notAmbientDyadic}, we give an example of non-doubling $P_X,Q_X$ for which $\gamma_* = \rho_*$, and yet these differ from the value one would obtain if a dyadic partition of the ambient space had been used to define the relative dimension. \par 
\begin{remark} \label{sharper_bounds_than_KM} Together, Propositions \ref{sharperBounds} and \ref{relDimEquivalence} show that upper bounds for the excess risk that we derive are at least as sharp as those of \cite{KM}, and sharper in the case $\gamma_* < \rho_* + d$. Whether or not this occurs depends on the geometry of the sub-region of $\cX_Q$ where the worst case local scaling of ratios $Q_X(B(x,r))/P_X(B(x,r))$ occurs; when this region has the same dimension as $\cX_Q$, one obtains $\gamma_* = \rho_* + d$. This occurs, for example, in the case in which both $P_X$, $Q_X$ are regular of dimension $d_P, d$ respectively, with $d_P > d$. In this case we have $\gamma_*(P_X,Q_X) = d_P$ and $\rho = d_P - d_Q$ (see \cite{KM}) so clearly $\gamma_* = \rho + d_Q$. On the other hand, when this worst-case scaling is restricted to a lower-dimensional subset of $\cX_Q$, we expect the strict inequality $\gamma_* < \rho_* + d$, and in this case the upper bound of \cite{KM} is not optimal. 
\end{remark}
The following example illustrates this. 

\begin{example} \label{E1}
Let $Q_X$ be uniformly distributed on $[0,1]^d$, and for $x \in [0,1]^d$ let $P_X$ have density $p(x) \propto \| x \|^\nu$ for some $\nu > 0$. In Appendix \ref{example1CalcAppendix}, we show that there is a constant $C$ for which $\gamma = \max(\nu, d)$ is an aggregate transfer exponent of $P_X$ relative to $Q_X$. In this example, it is shown in \cite{KM} that the transfer exponent is simply $\rho(P_X,Q_X) = \nu$, and so we have $\gamma_*(P_X,Q_X) = \max(\nu,d) < \rho(P_X,Q_X) + d$ when $\nu > 0$. This example can be extended in a simple way so as to yield aggregate exponents between $\max(\nu, d)$ and $\nu + d$, and it is precisely this construction that is used in Figure \ref{fig:E1_experiment}. Fix an integer $k \in \{1, \dots, d-1\}$, and let \begin{align*}
A_k = \{x \in [0,1]^d \mid (x_{k+1}, \dots, x_d) = 0\};  
\end{align*} note that $A_k$ is a subset of $[0,1]^d$ of Minkowski dimension $k$. Letting $d(x,A) = \inf_{y \in A} \norm{x - y}$ denote the standard distance function, we can define distributions $P_X^k$ on $[0,1]^d$ via density functions $p^k(x) \propto d(x,A_k)^\nu$; again, we take $Q_X \sim \mathrm{U}([0,1]^d)$. With this notation, the example given above corresponds to the choice $k = 0$. Now, note that for $x \in A_k$, we have that $Q_X(B(x,r)) / P_X(B(x,r)) \sim O(r^{-\nu})$. Since the Minkowski dimension of $A_k$ is $k$, there will be $O(r^{-k})$ elements of any grid that intersect $A_k$, yielding a contribution to the aggregate transfer exponent that is $O(r^{-(k + \nu)})$. In the rest of the space, we have $Q_X(B(x,r))/P_X(B(x,r)) \sim O(1)$, giving a contribution to the aggregate exponent of order $O(r^{-d})$. As $r \to 0$ the sum of these contributions grows as $O(r^{-\max(k + \nu, d)})$, giving an aggregate transfer exponent of $\max(k + \nu, d)$; this argument is made precise in the case $k = 0$ in Appendix B; for higher $k$ the calculations are entirely analogous and are therefore omitted. In each of these cases, we have $\rho(P_X^k, Q_X) = \nu$, so we see that the aggregate exponent accounts for both the strength of any marginal singularities and also the `size' of the region over which they occur. An analysis involving only the transfer exponent $\rho$ essentially assumes the worst case: that the singularity strength is of uniform order over the entire support $\cX_Q$; see the discussion of Figure \ref{fig:E1_experiment}. 
\end{example}

%%% move this to earlier and make it clearer. 
\par We close this section with the following Proposition, which demonstrates that, under a regularity condition on the target, the minimal aggregate transfer exponent is bounded below by the Minskowski dimension of the target. This implies that the upper bound on the excess risk from Theorem \ref{oracleRate} decreases no faster in $n_P$ than in $n_Q$, and precludes the possibility of \emph{super-transfer} (that is, transfer with faster rates using source samples alone; see \cite{HK_val_targ_data}). 

\begin{proposition}\label{noSuperTransfer}
Let $P_X,Q_X$ be probability measures on $[0,1]^D$. If there is an $S \subset \cX_Q$ with $\dim_M(S) = d$ and constants $C, r_0 > 0$ such that for all $x \in S$ and all $0 < r < r_0$ we have $Q_X(B(x,r)) \geq C r^d$, then $\gamma_*(P_X,Q_X) \geq d$. 

%Let $P,Q$ be probability measures on $[0,1]^D$, such that $\cX_Q$ is of %Minkowski dimension $d$. If there is a $C > 0$ such that for all $x \in %\cX_Q$ and all $0 < r < r_0$, we have $Q(B(x,r)) \geq C r^d$,   
\end{proposition}

\section{Analysis Details.} \label{sec_analysis_details}

This section contains an overview of the proofs of our main results; any details not fully provided here are to be found in the Appendices.

%Results concerning the relative dimension, including supporting lemmas required for the main results, are in the next section. 
\subsection{Main Tools.}

%\marginpar{\sk{Mult. CB are well known, but you can put a version in the appendix ...}}
We begin by recalling some well-known results that will give a decomposition of the excess risk of $\hat f_r$ under the target distribution $Q$. Let $\hat{f}_r$ denote the plug-in estimator   $\hat{f}_r \defeq \ind\{\hat{\eta}_r \geq 1/2\}$ based on an estimate $\hat \eta_r$ of $\eta$; let $f^*$ denote the Bayes classifier. Then it is well-known (see \cite{Devroye1996}, Theorem 2.2) that the  excess risk $\mathcal{E}_Q (\hat{f}_r )$ can be expressed as 
\begin{align} \mathcal{E}_Q(\hat{f}_r) & = R_Q(\hat{f}_r) - R^* \nonumber \\& = 2\int \abs{\eta(x) - \tfrac{1}{2}} \cdot \ind\{\hat f_r (x) \neq f^*(x)\}\, Q_X(dx). \label{devroyeLemma2}
\end{align}

 It will be convenient to also consider the  expectation of $\hat{\eta}$ conditional on the features $\bX$, which we denote $\tilde{\eta}_r \doteq \E_{\bY \mid \bX} \left[ \hat{\eta}_r\right] $. Note that by \eqref{etaHat} we have $\tilde{\eta}_r(x) =  0$ if $ |\tilde A_r(x) \cap \bX| = 0$ and
\begin{equation} \label{etaTilde} 
 \tilde \eta_r(x)= \frac{1}{|\tilde A_r(x) \cap \bX |}\sum_{i=1}^n \eta(X_i) \ind\{X_i \in \tilde{A}_r(x)\}
\end{equation} otherwise. 
%\sk{We also use the following shorthands: let
%\begin{align}\label{hatShortHands}
%    \hat B_r(x) = \abs{\tilde \eta_r (x) - \eta(x)}, \quad \hat \sigma_r(x) = \abs{\hat \eta_r(x) - \tilde \eta_r(x)}, \quad \hat V_r(x) = \frac{1}{n_P \hat P_X(\tilde A_r(x)) + n_Q \hat Q_X(\tilde A_r(x))},
%    \end{align} and let 
%\begin{align}\label{proxShortHands}
%B_r(x) = C_\alpha r^\alpha, \quad V_r(x) = \frac{1}{n_P P_X(\tilde A_r(x)) + n_Q  Q_X(\tilde A_r(x))}.        
%\end{align}
%}
Observe now that  on $\{\hat{f}_r(x) \neq f^*(x)\}$, we have $\abs{\eta(x) - 1/2} \leq \abs{\hat{\eta}_r(x) - \eta(x)}.$ A standard bias-variance decomposition of the error $\abs{\hat \eta_r - \eta}$ then yields 

\begin{equation*}%\label{tieToAdaptive} 
\abs{\hat \eta_r(x) - \eta(x) } \leq \abs{\hat \eta_r(x) - \tilde \eta_r(x)} + \abs{\tilde \eta_r(x) - \eta(x)}. \end{equation*}

%\begin{lemma}\label{tieToAdaptive} The plug-in estimator $\hat f_r$ satisfies
%\begin{equation}
%\ind\{ \hat f_r(x) \neq f^*(x)\} \leq \ind\{ \abs{\eta(x) - \tfrac{1}{2}} \leq \sk{\hat B_r(x) + \hat \sigma_r(x)} \}.
%\end{equation}
%\end{lemma}
%\abs{\hat \eta_r(x) - \tilde \eta_r(x)} + \abs{\tilde \eta_r(x) - \eta(x)}
 The following is a simple consequence of equation \ref{devroyeLemma2} and the inequality $\ind\{A \leq B + C\} \leq \ind\{A \leq 2B\} + \ind\{A \leq 2C\}$:

%below:
%\abs{\hat{\eta}_r(X) - \tilde{\eta}_r(X)}
%\abs{\tilde{\eta}_r(X) - \eta(X)}
%_{\Omega_{n,r}}
\begin{lemma}\label{nextlemma2} 
Let $\hat{f}_r$ be as in equation \ref{devroyeLemma2}, and write $M(x) \defeq \abs{\eta(x) - \tfrac{1}{2}}$ for the margin function. Then  
\begin{align} 
 \mathcal{E}_Q(\hat{f}_r) & \leq  \, 2 
 \underbrace{\int M(x) \cdot \ind\{M(x) \leq 2 \abs{\hat \eta_r(x) - \tilde \eta_r(x)}\} \, Q_X(dx)}_{\text{ $ = A_1$}}  \label{estimError} 
 \\& \quad + 2 \underbrace{\int M(x) \cdot \ind\{M(x) \leq 2 \abs{\tilde \eta_r(x) - \eta(x)}\} \, Q_X(dx)}_{\text{$ = A_2$}} \label{approxError2}.
%\leq & \, 2 \E_{Q} \left[\abs{\eta(X) - \tfrac{1}{2}} \ind\{\abs{\eta - \tfrac{1}{2}} \leq 2\abs{\hat{\eta}_r - \tilde{\eta}_r}\}\right] \label{estimError}\\& + 2\E_Q \left[\abs{\eta(X) - \tfrac{1}{2}} \ind\{\abs{\eta - \tfrac{1}{2}} \leq 2 \abs{\tilde{\eta}_r - \eta}\}\right]. \label{approxError}  
\end{align}
\end{lemma}

\subsection{Overview of Proofs.}
\subsubsection{Upper Bounds for Dyadic Trees}

In this section we outline the proof of our oracle upper bound. The tools that we employ are standard, although our approach makes it possible to derive the minimax optimal rate, unspoiled by a log-term, with minimal assumptions on the distributions $P_X,Q_X$.

As a first step towards bounding the terms in Lemma \ref{nextlemma2} in expectation over the sample, we derive some simple bounds on the error terms $\abs{\hat \eta_r(x) - \tilde \eta_r(x)}$ and $\abs{\tilde \eta_r(x) - \eta(x)}$.
Let us consider the following event: 

\begin{notation}\label{Sdef} Let 
\begin{align*} \Sone \defeq \{|\tilde A_r(x) \cap \bX | > 0\}\end{align*} 
denote the event that the envelope of the cell containing $x$ is non-empty, and likewise let $\Stwo(X) \defeq \{|\tilde A_r(X) \cap \bX| > 0\}.$

\end{notation} The following bounds are immediate:

\begin{lemma}[Bias]\label{firstBoundBias} Let $\tilde \eta, \Sone$ be as above. The bias term can be bounded as follows for all $x \in \cX_Q$:
\begin{equation*}
\abs{\tilde \eta_r(x) - \eta(x)} \ind\{\Sone\} \leq C_\alpha r^\alpha.
\end{equation*}
\end{lemma}

\begin{lemma}[Variance]\label{firstBoundVar}
Let $\hat \eta, \tilde \eta, \Sone$ be as above. For all $x \in \cX_Q$, we have:

%UNCOMMENT FOR TWO-COLUMN FORMAT:
%\begin{align}\label{VhatDef}
%&  \E_{\bY \mid \bX} \left[\abs{\hat \eta_r(x) - \tilde \eta_r(x)}^2  \mid \bX \right] \ind\{\Sone\} \nonumber \\ & \leq \frac{1}{4 \, |\tilde A_r(x) \cap \bX|} \defeq  \frac{1}{4} \hat V_r(x). 
%\end{align}

%UNCOMMENT FOR ONE-COLUMN FORMAT:
\begin{align}\label{VhatDef}
 \E_{\bY \mid \bX} \left[\abs{\hat \eta_r(x) - \tilde \eta_r(x)}^2  \mid \bX \right] \ind\{\Sone\} \leq \frac{1}{4 \, |\tilde A_r(x) \cap \bX|} \defeq  \frac{1}{4} \hat V_r(x). 
\end{align}

\end{lemma}

\par 

Using Lemma \ref{firstBoundBias}, we obtain a bound on the expectation over $(\bX,\bY)$ of \eqref{approxError2} when we restrict to $\Stwo(X)$ (that is, when $X$ is in the region corresponding to non-empty cells in $\Pi_r$): 

\begin{proposition}[Bounding $A_2$ in expectation.] \label{approxErrorProp}
Let $(P,Q) \in \cT$. Let $\tilde{\eta}_r$ be as in \eqref{etaTilde}, let $\Stwo(X)$ be as in Notation \ref{Sdef}, and let $M(x) = \abs{\eta(x) - \tfrac{1}{2}}$. Then for some $C > 0$ free of $n_P,n_Q$ we have

%UNCOMMENT FOR ONE-COLUMN FORMAT:
\begin{align*} \E \left[M(X) \cdot \ind\left\{M(X) \leq 2 \, \abs{\tilde \eta_r(X) - \eta(X)} \right\} \cdot  \ind\{\Stwo(X)\}\right] \leq C \, r^{\alpha (\beta + 1)},\end{align*} 

%UNCOMMENT FOR TWO-COLUMN FORMAT:
%\begin{align*} \E \left[M(X) \cdot \ind\left\{M(X) \leq 2 \, \abs{\tilde \eta_r(X) - \eta(X)} \right\} \cdot  \ind\{\Stwo(X)\}\right] \nonumber \\ \leq C \, r^{\alpha (\beta + 1)},\end{align*} 

where the expectation is over the full sample $(\bX,\bY)$ and the target $X \sim Q_X$.
\end{proposition}

\begin{proof}
 By Lemma \ref{firstBoundBias}, we have

%UNCOMMENT FOR ONE-COLUMN FORMAT: 
\begin{align*}
    \E \left[ M(X) \cdot \ind\left\{M(X) \leq 2 \abs{\tilde \eta_r(X) - \eta(X)} \right\} \cdot  \ind\{\Stwo(X)\} \right] &\leq \E_{Q_X} \left[M(X) \cdot \ind\left\{M(X) \leq 2  C_\alpha r^\alpha \right\} \right]
    \\& \leq 2C_\alpha r^\alpha \cdot Q_X( 0 < \abs{\eta(X) - \tfrac{1}{2}} < 2C_\alpha r^\alpha) 
 \\& \leq 2 C_\alpha r^\alpha \cdot C_\beta (2 C_\alpha r^\alpha)^\beta \\& = C r^{\alpha(\beta + 1)},
\end{align*}

%UNCOMMENT FOR TWO-COLUMN FORMAT: 
%\begin{align*}
%    \E M(X) & \cdot \ind\left\{M(X) \leq 2 \abs{\tilde \eta_r(X) - \eta(X)} \right\} \cdot  \ind\{\Stwo(X)\} \\ \quad & \leq \E_{Q_X} \left[M(X) \cdot \ind\left\{M(X) \leq 2  C_\alpha r^\alpha \right\} \right]
%    \\& \leq 2C_\alpha r^\alpha \cdot Q_X( 0 < \abs{\eta(X) - \tfrac{1}{2}} < 2C_\alpha r^\alpha) 
%    \\& \leq 2 C_\alpha r^\alpha \cdot C_\beta (2 C_\alpha r^\alpha)^\beta = C \cdot r^{\alpha(\beta + 1)},
%\end{align*}

where the final inequality applies the noise condition \eqref{tsybNoise}.
\end{proof}
To obtain a corresponding bound in expectation for the first term of Lemma \ref{nextlemma2}, we consider the event in which the quantity $\hat V_r(x)$ in \eqref{VhatDef} is close to its population counterpart $V_r(x)$, defined below. 

% was previously given with [Concentration Event] in parentheses

\begin{notation}\label{phiDef}
Let $x \in \cX_Q$, and let $$V_r(x)\defeq \frac{1}{n_P  P_X (\tilde A_r(x)) + n_Q Q_X(\tilde A_r(x))}.$$ Consider the event 

%UNCOMMENT FOR ONE-COLUMN FORMAT: 
\begin{align*}
    \phiOne \defeq \{\hat V_r(x) < 2 V_r(x)\} \nonumber = \left\{ \vert \tilde A_r(x) \cap \bX \vert > \tfrac{1}{2} \E \ | \tilde A_r(x) \cap \bX| \right\}, %\label{eq:phir}
\end{align*}

%UNCOMMENT FOR TWO-COLUMN FORMAT: 
%\begin{align*}
%    \phiOne &\defeq \{\hat V_r(x) < 2 V_r(x)\} \nonumber \\& = \left\{ \vert \tilde A_r(x) \cap \bX \vert > \tfrac{1}{2} \E \ | \tilde A_r(x) \cap \bX| \right\}, %\label{eq:phir}
%\end{align*}

that is, the event that the number of samples in $\tilde A_r(x)$ concentrates (multiplicatively) to its expectation. We further set $\phiTwo(X) \defeq \{ \hat V_r(X) < 2 V_r(X)\}$
% $\phiTwo \defeq \{\hat V_r(X) \leq 2 V_r(X)\}$.
\end{notation}

Now, since $\phiOne \subset \Sone$, restricting attention to $\phiOne$ for $x \in \cX_Q$ gives (via Lemma \ref{firstBoundVar}) the bound $$\E_{(\bX,\bY)} \abs{\hat \eta(x) - \tilde \eta(x)}^2 \cdot\ind\{\phiOne\} \leq  V_r(x)/2,$$ and we use this to bound the first term in Lemma \ref{nextlemma2} in terms of $\E_{Q_X} [V_r(X)]$. We obtain the following:

\begin{lemma}[Bounding $A_1$ in expectation]\label{estErrorLemma2} Let $(P,Q) \in \cT$, let $\phiTwo(X)$ be as in Notation \ref{phiDef} and let $M(x) = \abs{\eta(x) - \tfrac{1}{2}}$. Then there is a constant $C > 0$, independent of $n_P,n_Q$, such that for all $0 < t < 1$ we have 

%UNCOMMENT FOR ONE-COLUMN FORMAT: 
\begin{align*}
\E \left[M(X) \cdot \ind\{M(X) \leq 2 \abs{\hat \eta_r(X) - \tilde \eta_r(X)} \} \cdot \ind\{\phiTwo(X)\}\right] \leq C \left(t^{\beta + 1} + \frac{\E_{Q_X} \left[ V_r(X) \right]}{t} \right),
\end{align*}

%UNCOMMENT FOR TWO-COLUMN FORMAT: 
%\begin{align*}
%\E \left[M(X) \cdot \ind\{M(X) \leq 2 \abs{\hat \eta_r(X) - \tilde \eta_r(X)} \} \cdot \ind\{\phiTwo(X)\}\right] \\ \leq C \left(t^{\beta + 1} + \frac{\E_{Q_X} \left[ V_r(X) \right]}{t} \right),
%\end{align*} 

where the expectation is over the full sample $(\bX,\bY)$ and the target $X \sim Q_X$. 
\end{lemma}

%\skr{{\bf Simplify the line of equations below ...} (not just here but wherever the same type f situation arises): 
%Inline in the proof, simply define the common part of what's you're bounding first as in (to make it all more readable, as one should avoid carrying the same long lines of equations over multiple lines): 

%\begin{quote} 
%Define $\Gamma(X) \doteq M(X) \ind\{\ldots\} ...$. We then have
%$$\E \ \Gamma(X) = \E \ \Gamma(X)\cdot (\ind\{\hat \sigma_r(X) \leq t\} + \ind\{\hat \sigma_r(X) > t\},$$
%and we next proceed to bounding the two terms separately...
%\end{quote} 

%\\Consider the following two terms, which we shall bound in turn:\\ \begin{equation}\label{term1}  \E \left[M(X) \ind\{M(X) \leq 2 \hat \sigma_r(X)\} \ind(\phiTwo(X)) \ind\{\hat \sigma_r(X) \leq t\}\right]\end{equation} and \begin{equation}\E \left[M(X) \ind\{M(X) \leq 2 \hat \sigma_r(X)\} \ind(\phiTwo(X)) \ind\{\hat \sigma_r(X) > t\}\right].\end{equation}
\begin{proof}
Let $\Phi_{r}$ be as above, let $\hat M_r(x) \defeq \abs{\hat \eta_r(x) - \tilde \eta_r(x)}$, and let \begin{align*} \Gamma(X) \defeq M(X) \cdot \ind\{ M(X) \leq 2 \hat M_r(X)\}. \end{align*} Fixing a $t \in (0,1)$, we then have 

%UNCOMMENT FOR ONE-COLUMN FORMAT:
\begin{align*}  \E \ \Gamma(X) \cdot \ind\{\phiTwo(X)\}  = \E \left[ \Gamma(X) \cdot \ind\{\phiTwo(X)\} \cdot \ind\{ \hat M_r(X) \leq t\} \right]  + \E \left[ \Gamma(X)\cdot \ind\{\phiTwo(X)\}\cdot \ind\{ \hat M_r(X) > t\}\right],
\end{align*}

%UNCOMMENT FOR TWO-COLUMN FORMAT:
%\begin{align*}  \E \ \Gamma(X) \ind\{\phiTwo(X)\} & = \E \ \Gamma(X) \ind\{\phiTwo(X)\} \ind\{ \hat M_r(x) \leq t\} \\& \quad + \E \ \Gamma(X) \ind\{\phiTwo(X)\} \ind\{ \hat M_r(x) > t\},
%\end{align*}

and we proceed to bound these two terms separately. 
 For the first term, note that we have \begin{align*} \E \left[ \Gamma(X) \cdot\ind(\phiTwo(X)) \cdot\ind\{\hat M_r(X) \leq t\}\right] \leq 2t \, Q_X(|\eta(X) - \tfrac{1}{2}| \leq t),\end{align*} and by the noise condition we have $$Q_X(|\eta(X) - \tfrac{1}{2}| \leq t) \leq C_\beta t^{\beta},$$  therefore we find 
\begin{align*}
  \E \left[ \Gamma(X)\cdot \ind(\phiTwo(X))\cdot \ind\{\hat M_r(X) \leq t\}\right] \leq 2 C_\beta t^{(\beta + 1)}.
\end{align*} For the second term, we have 

%UNCOMMENT FOR ONE-COLUMN FORMAT:
\begin{align*}
     \E \left[ \Gamma(X)\cdot \ind(\phiTwo(X))\cdot \ind\{\hat M_r(X) > t\}  \right] \leq 2 \, \E \left[ \hat M_r(X) \cdot\ind\{ \hat M_r(X) > t\} \cdot \ind(\phiTwo(X)) \right],
\end{align*} 

%UNCOMMENT FOR TWO-COLUMN FORMAT:
%\begin{align*}
%     \E \ \Gamma(X) \ind(\phiTwo(X)) \ind\{\hat M_r(X) > t\} \\ \quad \leq 2\E \ \hat M_r(X) \ind\{ \hat M_r(X) > t\} \ind(\phiTwo(X)),
%\end{align*} 

and we further let $$\hat M_r(X,t) \defeq \hat M_r(X)\cdot \ind\{\hat M_r(X) > t\} \cdot\ind\{\phiTwo(X)\}.$$ Then, writing $\pr$ for the joint distribution over the full sample and the target marginal $Q_X$, we have

%UNCOMMENT FOR ONE-COLUMN FORMAT:
\begin{align*}
    \E \ \hat M_r(X,t)  =  \int_0^1 \pr(\hat M_r(X,t) > s) \, ds = \int_0^t \pr(\hat M_r(X,t) > s) \, ds + \int_t^1 \pr(\hat M_r(X,t) > s) \, ds . 
\end{align*} 

%UNCOMMENT FOR TWO-COLUMN FORMAT:
%\begin{align*}
%    \E \ \hat M_r(X,t)  &=  \int_0^1 \pr(\hat M_r(X,t) > s) \, ds \nonumber \\   & = \int_0^t \pr(\hat M_r(X,t) > s) \, ds \\ & \quad  + \int_t^1 \pr(\hat M_r(X,t) > s) \, ds . 
%\end{align*} 

Now, since $\hat M_r(X,t)$ takes values on $\{0\} \cup (t,1]$, for $s \in [0,t]$ we have \begin{align*} & \pr(\hat M_r(X,t)  > s) = \pr(\hat M_r(X,t) > t), \end{align*} whence 
\begin{align*}
     \int_0^t \pr(\hat M_r(X,t) > s) \, ds  = t \, \pr(\hat M_r(X,t) > t)  \leq \frac{1}{t} \, \E \left[\hat M_r(X)^2 \ind\{\phiTwo(X)\} \right],
\end{align*}
 by Chebyshev's inequality. By Lemma \ref{firstBoundVar}, we have (note that $\phiTwo(X) \subset \Stwo(X)$)
\begin{align*} \E \left[\hat M_r(X)^2 \ind\{\phiTwo(X)\} \right] \leq  \E \left[ \hat V_r(X) \ind\{\phiTwo(X)\} \right]  \leq 2 \E_{Q_X}[V_r(X)].\end{align*} 
Another application of Chebyshev's inequality gives 

%UNCOMMENT FOR ONE-COLUMN FORMAT:
\begin{align*}
 \int_t^1 \pr(\hat M_r(X,t)  > s) \, ds   &  \leq \E[\hat M_r(X)^2 \ind\{\phiTwo(X)\} ] \int_t^1 s^{-2} \, ds \\& = \E\left[\hat M_r(X)^2 \ind\{\phiTwo(X)\} \right] (t^{-1} - 1) \leq 2  \frac{1}{t} \, \E_{Q_X}[V_r(X)],
\end{align*} 

%UNCOMMENT FOR TWO-COLUMN FORMAT:
%\begin{align*}
%& \int_t^1 \pr(\hat M_r(X,t)  > s) \, ds  \\  & \quad  \leq \E[\hat M_r(X)^2 \ind\{\phiTwo(X)\} ] \int_t^1 s^{-2} \, ds \\ & \quad = \E\left[\hat M_r(X)^2 \ind\{\phiTwo(X)\} \right] (t^{-1} - 1) \\& \quad  \leq 2  \frac{1}{t} \, \E_{Q_X}[V_r(X)],
%\end{align*} 

where the final inequality again uses Lemma \ref{firstBoundVar} and the definition of $\phiTwo(X)$; this completes the proof. 
\end{proof}

\begin{remark}
The above Lemma constitutes the critical step to proving Theorem \ref{oracleRate}. Interestingly, the proof given above essentially follows a strategy outlined by Audibert and Tsybakov in \cite{Audibert2007} (see Lemma 5.2 and the ensuing discussion) but rejected by these authors as being inadequate to establish upper bounds of this type under the noise condition \eqref{tsybNoise} with $\beta > 0$. The issue is resolved by splitting the excess risk into terms controlled respectively by $\vert \tilde \eta - \eta \vert$ and  $\vert \hat \eta - \tilde \eta \vert$, and applying the technique they outline only to the term featuring $\vert \hat \eta - \eta \vert$, as we do in Lemma \ref{estErrorLemma2} above. In the absence of this splitting, we would end up with an upper bound for the excess risk in terms of $\E (\hat \eta(X) - \tilde \eta(X))^2$; this would be insufficient to derive the bound we seek, since under the noise condition with $\beta > 0$, the optimal plug-in model for classification differs from the plug-in model that uses an estimate of the regression function that minimizes the $L^2$ regression error. 
\end{remark}

Choosing $t$ appropriately to balance the terms above immediately gives:
\begin{corollary}\label{corollary1}
Under the conditions of Lemma \ref{estErrorLemma2}, choosing $t = \E_{Q_X}[V_r(X)]^{\frac{1}{\beta + 2}}$, we have 
\begin{align*}
\E \left[M(X) \cdot \ind\{M(X) \leq 2 \abs{\hat \eta_r(X) - \tilde \eta_r(X)} \} \cdot \ind\{\phiTwo(X)\}\right] \leq C \cdot  \E_{Q_X} \left[ V_r(X) \right]^{\frac{\beta + 1}{\beta +2}}.
\end{align*}
\end{corollary}

In order to bound $\E_{Q_X}[V_r(X)]$, we will make use of the aggregate transfer exponent (see Definition \ref{aggExpDef}) to account for the contribution of the source feature distribution $P_X$. This is done via the following technical Lemma, whose proof may be found in Appendix \ref{appendixProofs}: 

 \begin{lemma}\label{gammaSumLemma}
Let $(P,Q) \in \cT$, and let $\Pi_r(\cX_Q)\defeq \{A \in \Pi_r: Q_X(A) > 0\}$. There exists $C > 0$ independent of $r$ such that \begin{equation*}\label{envSumBound}
    \sum_{A \in \Pi_r(\cX_Q)} \frac{Q_X(\tilde A)}{P_X(\tilde A)} \leq C r^{- \gamma}. \end{equation*}
\end{lemma}
Note that in order to obtain Lemma \ref{gammaSumLemma}, it is essential that the sum on the left-hand side should take the mass-ratios of the \it envelopes \rm of the cells $A \in \Pi_r(\cX_Q)$; indeed, it is for this reason that we define the algorithm in this way. Equipped with this Lemma, we obtain the following bound on the expectation of $V_r$:

\begin{lemma}\label{VexpBound}
Let $(P,Q) \in \cT$, and let $V_r$ be as above. Then for some $C > 0$ free of $r$, we have
\begin{equation*}
    \E_{Q_X}[V_r(X)] \leq C \min \left( \frac{r^{-\gamma}}{n_P}, \frac{r^{-d}}{n_Q} \right).
\end{equation*}
\end{lemma}

\begin{proof}
 Since $V_r$ is constant over the cells $A \in \Pi_r$, we have  

%UNCOMMENT FOR TWO-COLUMNT FORMAT: 
\begin{align*}
     \E_{Q_X}[V_r(X)] & = \sum_{A \in \Pi_r(\cX_Q)} \left\{ \frac{1}{n_P P_X(\tilde A) + n_Q Q_X(\tilde A)} \right\} \cdot Q_X(A)  \leq  \sum_{A \in \Pi_r(\cX_Q) } \frac{Q_X(\tilde A)} {n_P P_X(\tilde A) + n_Q Q_X(\tilde A)}
     \\&  \leq \frac{1}{n_Q} \left\vert \Pi_r(\cX_Q)  \right\vert \leq C_d \frac{r^{-d}}{n_Q}, 
\end{align*} 

%UNCOMMENT FOR TWO-COLUMNT FORMAT: 
%\begin{align*}
%    & \E_{Q_X}[V_r(X)] \\ & \quad = \sum_{A \in \Pi_r(\cX_Q)} \left\{ \frac{1}{n_P P_X(\tilde A) + n_Q Q_X(\tilde A)} \right\} \cdot Q_X(A) \\ & \quad \leq  \sum_{A \in \Pi_r(\cX_Q) } \frac{Q_X(\tilde A)} {n_P P_X(\tilde A) + n_Q Q_X(\tilde A)}
%     \\& \quad  \leq \frac{1}{n_Q} \left\vert \Pi_r(\cX_Q)  \right\vert \leq C_d \frac{r^{-d}}{n_Q}, 
%\end{align*} 

where the final step uses $|\Pi_r(\cX_Q)| \leq C_d r^{-d}$, which follows from assumption \eqref{QmetricEntropy}. Similarly, we have 

%UNCOMMENT FOR ONE-COLUMNT FORMAT: 
\begin{align*}
       \E_{Q_X}[V_r(X)]  \leq \sum_{A \in \Pi_r(\cX_Q) } \frac{Q_X(\tilde A) }{n_P P_X(\tilde A) + n_Q Q_X(\tilde A)}  \leq \frac{1}{n_P} \sum_{A \in \Pi_r(\cX_Q)} \frac{Q_X( \tilde A)}{P_X(\tilde A)} \leq C \frac{r^{-\gamma}}{n_P},
\end{align*} 

%UNCOMMENT FOR TWO-COLUMNT FORMAT: 
%\begin{align*}
%       \E_{Q_X}[V_r(X)] & \leq \sum_{A \in \Pi_r(\cX_Q) } \frac{Q_X(\tilde A) }{n_P P_X(\tilde A) + n_Q Q_X(\tilde A)} \\ & \leq \frac{1}{n_P} \sum_{A \in \Pi_r(\cX_Q)} \frac{Q_X( \tilde A)}{P_X(\tilde A)} \leq C \frac{r^{-\gamma}}{n_P},
%\end{align*} 
where the final inequality is an application of Lemma \ref{gammaSumLemma}, and the result follows immediately. \end{proof}

Combining Lemma \ref{firstBoundVar}, Corollary \ref{corollary1}, and Lemmas \ref{gammaSumLemma} and \ref{VexpBound} gives:

\begin{proposition}%[Bounding \eqref{estimError}]
\label{varTermProp}
 Let $(P,Q) \in \cT$, let $\phiTwo(X)$ be as in Notation \ref{phiDef}. Then there is a constant $C > 0$, independent of $n_P,n_Q$, such that 
\begin{align*}
 \E \left[M(X) \cdot \ind\left\{M(X) \leq 2 \abs{\hat \eta_r(X) - \tilde \eta_r(X)} \right\} \cdot \ind\{\phiTwo(X)\}\right] \leq C \left( \min\left( \frac{r^{-\gamma}}{n_P}, \frac{r^{-d}}{n_Q} \right)\right)^{\frac{\beta +1}{\beta + 2}},
\end{align*} where the expectation is over the full sample $(\bX,\bY)$ and the target $X \sim Q_X$. 
\end{proposition}

Up until now, we have established bounds on the excess expected risk of $\hat f_r$ conditioned on the events $\Stwo(X)$ and $\phiTwo(X)$, and it therefore remains to show that these events occur with high enough probability. Since clearly $ \phiTwo(X) \subset \Stwo(X)$, we must show that $\phiTwo(X)^\comp$ is sufficiently rare. In order to see this, we employ a multiplicative Chernoff bound (Lemma \ref{relChernoff} in Appendix \ref{appendixProofs}), followed by an application of Lemma \ref{gammaSumLemma}. This gives:

\begin{lemma}\label{compPhiRare}
Let $\mathbb{P}$ denote the joint probability over the sample and the target, and let $\phiTwo(X)$ be as in Notation \ref{phiDef}. 
%Under the conditions of Lemma \ref{estErrorLemma2}, 
There is a constant $C > 0$ free of $r, n_P,n_Q$ such that we have 
\begin{equation*}
    \mathbb{P}(\phiTwo(X)^\comp) \leq  C \min\left( \frac{r^{-\gamma}}{n_P}, \frac{r^{-d}}{n_Q} \right) 
\end{equation*}
\end{lemma}

\begin{proof}
Recall that $\phiTwo(X) = \{\hat V_r(X) < 2 V_r(X)\}$, and note that $\hat V_r^{-1}(X) = | \tilde A_r(x) \cap \bX |$ is a sum of $n_P + n_Q$ independent Bernoulli variables, and has mean $V_r(X)^{-1}$. Fixing $x \in \cX_Q$ and applying the multiplicative Chernoff bound (Lemma \ref{relChernoff}) gives 
\begin{align*}%\label{phiLoc}
    \pr(\phiOne^\comp) = \pr(\hat V^{-1}_r(x) \leq \tfrac{1}{2} V_r^{-1}(x))  \leq \exp\{-\tfrac{1}{8} V^{-1}_r(x)\}.
\end{align*} It follows then that

%UNCOMMENT FOR ONE-COLUMNT FORMAT:
\begin{align*}
    \pr(\phiTwo(X)^\comp)  &= \sum_{A \in \Pi_r} \pr(\phiTwo(X)^\comp \cap \{X \in A\}) = \sum_{A \in \Pi_r} \pr(\phiTwo(X)^\comp \mid X \in A)Q_X(A) \\&  \leq \sum_{A \in \Pi_r} \exp\left\{-\tfrac{1}{8} V_r(A)^{-1} \right\} \cdot Q_X(A) 
      =  \sum_{A \in \Pi_r} V_r(A)^{-1} \exp\left\{-\tfrac{1}{8} V_r(A)^{-1} \right\} V_r(A) \cdot Q_X(A) \\& \leq \frac{8}{e} \sum_{A \in \Pi_r} V_r(A) Q_X(A) \leq 3 \E_{Q_X}[V_r(X)],
\end{align*}

%UNCOMMENT FOR TWO-COLUMNT FORMAT:
%\begin{align*}
%    &\pr(\phiTwo(X)^\comp)  \\ & \quad = \sum_{A \in \Pi_r} \pr(\phiTwo(X)^\comp \cap \{X \in A\}) \\ 
%    & \quad= \sum_{A \in \Pi_r} \pr(\phiTwo(X)^\comp \mid X \in A)Q_X(A) \\
%    & \quad \leq \sum_{A \in \Pi_r} \exp\left\{-\tfrac{1}{8} V_r(A)^{-1} \right\} \cdot Q_X(A) \\
%    & \quad =  \sum_{A \in \Pi_r} V_r(A)^{-1} \exp\left\{-\tfrac{1}{8} V_r(A)^{-1} \right\} V_r(A) \cdot Q_X(A) \\
%    &\quad \leq \frac{8}{e} \sum_{A \in \Pi_r} V_r(A) Q_X(A) \leq 3 \E_{Q_X}[V_r(X)],
%\end{align*}

where the second inequality uses $xe^{-ax} \leq 1/ae$, valid for all real $x$. The proof is completed by applying Lemma \ref{VexpBound}. 
\end{proof}

Having now fully established a bound on the expected excess risk of $\hat f_r$, the proof of Theorem \ref{oracleRate} is completed upon collecting and balancing terms. 
\begin{proof}{(\textit{Theorem \ref{oracleRate}})}
By Lemma \ref{nextlemma2}, Propositions  \ref{approxErrorProp}, \ref{varTermProp} and Lemma \ref{compPhiRare}, we can choose $C > 0$ such that \begin{align*}
\E[\cE(\hat f_r)] \ \leq C \left( r^{\alpha(\beta + 1)} + \min\left( \frac{r^{-\gamma}}{n_P}, \frac{r^{-d}}{n_Q}\right)^{\frac{\beta + 1}{\beta + 2}} \right) + C \min \left( \frac{r^{-\gamma}}{n_P}, \frac{r^{-d}}{n_Q}\right).
\end{align*} Some algebra shows that choosing $$r_n^* \defeq \min\left( n_P^{-\frac{1}{2\alpha + \alpha \beta + \gamma}}, n_Q^{-\frac{1}{2\alpha + \alpha \beta + d}} \right)$$ balances the first two terms, and the third term is clearly strictly smaller than the second. This gives a final bound of 
$$\E\left[\cE(\hat f_{r_n^*})\right] \leq C \min \left( n_P^{-\frac{\alpha(\beta + 1)}{2\alpha + \alpha \beta + \gamma}}, n_Q^{-\frac{\alpha(\beta + 1)}{2\alpha + \alpha \beta + d}} \right),$$ 
\end{proof} completing the proof.

\subsubsection{Adaptive Rates.}

In this section we outline the proof of Theorem \ref{adaptiveTheorem}, establishing an upper bound on the excess risk of the classifier output by Algorithm \ref{lepskiAlg} that locally determines a depth at which to estimate $\eta$ at a point $x$. We demonstrate that it achieves the minimax rate plus a log-term. Note that `log-spoiled' adaptive rates are usually unavoidable (see the discussion in \cite{Lepski1997} for more detail on this point)
%unsure if this is the correct Lepski paper to cite
without any knowledge of the unknown parameters. The method proceeds by making an initial choice of tree-depth, and then for an input point $x$, proceeds up the branch containing $x$ until a level is found at which the local error due to bias and variance are approximately balanced (see Algorithm \ref{lepskiAlg}). The level of the tree which achieves this will depend on the problem parameters and is therefore unknown, and the crux of the argument lies in demonstrating that this can nonetheless be achieved with only an estimate of the local variance.  \par  
We introduce the following notation: fix $\epsilon = (n_P + n_Q)^{-1/2}$, and let the set of admissible levels be given by \begin{align} \hat{R}_n \defeq \{2^{-i}\}_{i=0}^{ \lceil \log_2(1/\epsilon) \rceil}. \label{levels_def} 
\end{align} 
%Again let $\Pi_r$ denote the regular partition of $[0,1]^D$ into hypercubes of length $r$, which in this case correspond to nodes of the dyadic tree. 
%As usual, for $x \in [0,1]^D$, let . For a fixed $r$, let $\Pi_r(\cX_Q)$ denote those cells $A_r \in \Pi_r$ such that $Q_X(A_r) > 0$. Let $\Pi(\cX_Q)$ denote the collection of all cells that are in $\Pi_r(\cX_Q)$ for some $r \in \hat{R}_n$; that is, \begin{align*} %\label{allCells} 
%\Pi(\cX_Q) = \cup_{r \in \hat R_n} \Pi_r(\cX_Q).\end{align*}
%By assumption, we have $|\Pi_r(\cX_Q)| \leq C^\prime_d \cdot r^{-d}$,
%% add a comment in minkowski dim section that this is valid? i.e. that can count covers, packings, grid covers, etc. and all yield the same 'd'
%and it follows then that there is some constant $C$ such that $|\Pi(\cX_Q)| \leq C \epsilon^{-d}$.  
%\skr{relation between $C_d$, $C^\prime_d$?} %For a set $B \subset [0,1]^D$, let $n = n_P + n_Q$ and 
%\begin{equation*}
%    W_n(B) \defeq \frac{n_P \hat{P}_{X,n}(B) + n_Q \hat{Q}_{X,n}(B)}{n_Q + n_Q}.
%\end{equation*}

The key to this approach is to establish control of the error $\abs{\hat \eta_r(x) - \tilde \eta_r(x)}$ uniformly for $x \in \cX_Q$, and at all levels $r \in \hat R_n$, simultaneously. Using McDiarmid's bounded differences inequality \cite{mcdiarmid_1989}, we can derive the following bound in high probability. Recall that $A_r(x)$ denotes the cell containing $x$ at level $r$, $\tilde{A}_r$ is its $r$-envelope, $|\tilde A_r(X) \cap \bX|$ denotes the number of sample points falling in $\tilde A_r(X)$, and we let $\hat V_r(X)  = |\tilde A_r(X) \cap \bX|^{-1}$. 

\begin{lemma}\label{lepskiControl}
Let $P,Q \in \cT$. Fix $\delta > 0$, and let $\hat{\eta}_r,\tilde{\eta}_r$ be as in \eqref{etaHat} and \eqref{etaTilde}, respectively. Then with probability at least $1 - \delta$ over the joint sample $(\bX,\bY)$, we have, for all $x \in \cX_Q$ and each $r \in \hat{R}_n$, that 
\begin{align} 
     \abs{\hat{\eta}_{r}(x) - \tilde{\eta}_r(x)} 
    \leq \frac{1}{2} \sqrt{\hat V_r(x)} \left( 1 + 2\, \sqrt{\log (n_P + n_Q)/\delta) }\right). \label{PhiEvent}
\end{align}
\end{lemma}

Using this, the triangle inequality, and the fact that the regression function is $\alpha$-H\"older continuous immediately gives 

\begin{lemma}\label{regBoundEstApprox}
Let $P,Q \in \cT$. Fix $\delta > 0$, and let $\hat{\eta}_r,\tilde{\eta}_r$ be as in \eqref{etaHat} and \eqref{etaTilde}, respectively. Then with probability at least $1 - \delta$ over the joint sample $(\bX,\bY)$, we have, for all $x \in \cX_Q$ and each $r \in \hat{R}_n$, that 
\begin{align*}
 \abs{\hat{\eta}_r(x) - \eta(x) }  \leq \frac{1}{2} \sqrt{\hat V_r(x)}\left( 1 + 2\, \sqrt{\log (n_P + n_Q)/\delta) }\right) + C_\alpha r^\alpha. 
\end{align*}
\end{lemma}

Note that above we need not worry about controlling for empty cells in applying the H\"older condition above, as the bound is trivial in this case (as $|\tilde{A}_r(x) \cap \bX | = 0$). Now, let $C_{n,\delta}
= \left( 1 + 2\, \sqrt{\log (n_P + n_Q)/\delta) }\right)$ and set $\hat{\sigma}_r(x) = C_{n,\delta} \, \hat V_r(x)^{1/2}$, and set $r_0 = 2^{- \lceil \log(n_P + n_Q)/2 \rceil}$ (the smallest value in $\hat{R}_n$), and choose $\hat{r}(x)$ by implementing Algorithm \ref{lepskiAlg} with these specifications. To show that this procedure yields the correct rates we appeal to the following Lemma, after which the proof follows the same argument as for Theorem \ref{oracleRate}.

Let \begin{align}\label{oracleDepthLog}
%& r_n \nonumber \\& \quad 
r_n \defeq \min \left\{ \left(\frac{\log n_P}{n_P}\right)^{\tfrac{1}{2 \alpha + \alpha \beta + \gamma}}, \left(\frac{\log n_Q}{n_Q}\right)^{\tfrac{1}{2 \alpha + \alpha \beta + d}}\right\};
\end{align} the optimal depth choice implied by Theorem \ref{oracleRate} with an adjustment due to the logarithmic dependence of $C_{n,\delta}$ on $n = n_P + n_Q$. The critical result, which relies on an idea developed by Lepski \cite{Lepski1997} and formalized as the `Intersecting Confidence Interval' method \cite{ICIGoldNem}, implicitly relates the error of the adaptive classifier $\hat f$ to the error of the oracle classifier $\hat f_{r_n}$ (note that, on the event of Lemma \ref{regBoundEstApprox}, we have $|\hat \eta_{r_n}(x) - \eta(x)| \leq C_{n,\delta} \hat V_{r_n}^{1/2} + C_\alpha r_n^\alpha$, so the right-hand side of equation \eqref{lepskiRateBound} below is an upper bound on the loss suffered by the oracle) is as follows:

% as in \eqref{oracleDepthLog}
%\skr{What is $C$ in the next Lemma? }

\begin{lemma}\label{lepskiKeyLemma}
There is a universal constant $C$ such that the following holds. Let $r_n$ be as in \eqref{oracleDepthLog}. For a fixed $x$, let $\hat{f}(x)$ denote the output of Algorithm \ref{lepskiAlg} when run from level $r_0 = (n_P + n_Q)^{-1/2}$, with $\hat{\sigma}_r(x)$ as above and $C_{n,\delta} = \tfrac{1}{2}\left( 1 + 2\, \sqrt{\log (n_P + n_Q)/\delta) }\right)$. Then with probability at least $1 - \delta$ over the full sample, simultaneously at all $x \in \cX_Q$ , we have:  

\begin{align}\label{lepskiRateBound}
      \ind\{ f^*(x) \neq \hat{f}(x)\} \leq \ind\left\{ \abs{\eta(x) - \tfrac{1}{2}} \leq C \left( C_{n,\delta}\hat V_{r_n}(x)^{1/2} + C_\alpha r_n^\alpha \right)\right\}. 
\end{align}
\end{lemma}

This allows us to write the following bound, in analogy with Lemma \ref{nextlemma2} from the previous subsection:

\begin{lemma}\label{lemmaAdaptiveDecomp}
Let $C_{n,\delta}$ and $\hat f$ be as above, and take $r_n$ as in \eqref{oracleDepthLog}; set $M(x) = \abs{\eta(x) - \tfrac{1}{2}}$. With probability at least $1 - \delta$ over the full sample $(\bX,\bY)$, we have 
\begin{align*} 
 \mathcal{E}_Q(\hat{f}) \leq 2 \int M(x) \cdot \ind\left\{M(x) \leq 2 C_{n,\delta}\hat V_{r_n}(x)^{1/2} \right\} \, Q_X(dx) + 2\int M(x) \cdot \ind\{M(x) \leq 2 C_\alpha r_n^\alpha\} \, Q_X(dx).
\end{align*}
\end{lemma}
The rest of the proof of Theorem \ref{adaptiveTheorem} can be completed in the exact same manner as Theorem \ref{oracleRate}, except that Lemmas \ref{firstBoundBias}, \ref{firstBoundVar} are superseded by Lemma \ref{regBoundEstApprox}. Note that the presence of the additional term involving $C_{n,\delta} \asymp \sqrt{\log ( n_P + n_Q)}$ is accounted for in the final balancing step by choosing $r_n$ as in \eqref{oracleDepthLog}. The full argument is given in Appendix \ref{appendixProofs}. 

\subsubsection{Localization to the Decision Boundary}

%%%%% LOCALIZATION STUFF TO BE MOVED TO PROOF SECTION / APPENDIX

In order to prove Theorem \ref{oracleEpsilonMarginBound}, we consider an oracle classification model, based on the dyadic-tree classifier introduced above that, for fixed $\epsilon$, estimates the regression function at different levels of the tree based on whether the input is in $\cG_\epsilon^-$ or $\cG_\epsilon^+$. 

\begin{definition}
Let $\hat f_r$ denote the dyadic-tree classifier of Definition \ref{treeClassifier}, built from the tree from Definition \ref{dyadicTree} at level $r$. For fixed $\epsilon$, let $r_- = r_-(\epsilon)$, $r_+ = r_+(\epsilon)$ be functions of $\epsilon$ (to be specified later), and let $\hat f_\epsilon$ be the hybrid dyadic-tree classifier that estimates at level $r_-$ when $x \in \cG_\epsilon^-$ and at level $r_+$ when $x \in \cG_\epsilon^+$; that is, 
\begin{equation*} \hat f(x) = \hat f_{r_-}(x) \ind\{x \in \cG_\epsilon^{-}\} + \hat f_{r_+}(x) \ind\{x \in \cG_\epsilon^+\}. \end{equation*} 
\end{definition}

Using the techniques used to prove Theorem \ref{oracleRate}, we can bound the contribution to the excess risk of $\hat f$ of the region $\cG_\epsilon^-$.

\begin{notation} For a set $A \subset \cX$ we let \begin{align}
    \cE(\hat f; A) \defeq 2 \int_{A} \abs{\eta(x) - \tfrac{1}{2}} \ind\{\hat f(x) \neq f^*(x)\} \, Q_X(dx),  \label{riskOverSetA}
\end{align} so that $\cE(\hat f; \cG_\epsilon^-)$ denotes the excess risk of $\hat f$ from the region $\cG_\epsilon^-$.
\end{notation}

We have the following bound.

\begin{lemma}[\textit{Risk contribution from $\cG_\epsilon^-$.}] \label{riskFromGminus} There is a universal constant $C = C(\cT)$ such that the following holds. Fix $\epsilon > 0$, and let $\hat f_\epsilon$ and $\cG_\epsilon^-$ be as above, and set $r_- = \min \left( n_P^{-1/(2\alpha + \alpha \beta + \gamma(\epsilon)}, n_Q^{-1/(2\alpha + \alpha \beta + d)}\right)$. Then we have
\begin{align*}  \E \ \cE(\hat f_\epsilon; \cG_\epsilon^-) \leq C \min\left\{ n_P^{-\tfrac{\alpha(\beta +1)}{2 \alpha + \alpha \beta + \gamma(\epsilon)}}, n_Q^{-\tfrac{\alpha(\beta+1)}{2\alpha + \alpha \beta + d}} \right\}.
\end{align*}
\end{lemma}

\begin{proof}
This follows immediately from Theorem 1, since by \eqref{riskOverSetA} we need only integrate over $\cG_\epsilon^-$, and in this region we have the local aggregate exponent $\gamma(\epsilon)$. 
\end{proof}

\begin{lemma}[\textit{Risk contribution from $\cG_\epsilon^+$.}]\label{riskFromGplus}
There is a universal constant $C = C(\cT)$ such that the following holds. Fix $\epsilon > 0$, and let $\hat f_\epsilon$ and $\cG_\epsilon^+$ be as above, and suppose that $r_+ < (\epsilon/2 C_\alpha)^{1/\alpha}$. Then we have
\begin{equation*}
    \E \ \cE(\hat f_\epsilon; \cG_\epsilon^+) \leq C \: \frac{1}{\epsilon} \min\left(  \frac{1}{n_P} r_+^{-\gamma^*}, \frac{1}{n_Q} r_+^{-d} \right). \end{equation*}
\end{lemma}

 Notice that the above expression is minimized when $r_+$ is taken as large as possible. This immediately implies: 
\begin{corollary}
Setting $r_+ \defeq (\epsilon/4 C_\alpha)^{1/\alpha}$, we have 
\begin{equation*}
% \ind\{X \in \cG_\epsilon^+\} 
    \E \ \cE(\hat f; \cG_\epsilon^+) \leq C \:  \min\left(  \frac{1}{n_P} \epsilon^{-(1 + \gamma^*/\alpha)}, \frac{1}{n_Q} \epsilon^{-(1 + d/\alpha)} \right). \end{equation*}
\end{corollary}

Let $\hat f_\epsilon$ be as above, where $$r_- = \min \left( n_P^{-1/(2\alpha + \alpha \beta + \gamma(\epsilon)}, n_Q^{-1/(2\alpha + \alpha \beta + d)}\right)$$ and $r_+ = (\epsilon/4 C_\alpha)^{1/\alpha}$. Combining the above results gives

\begin{corollary}
There is a universal constant $C = C(\cT)$ such that, for any $\epsilon > 0$, if $\hat f_\epsilon$ is set as above then we have
\begin{align*}%\label{excessRisk_f_epsilon}
    \E \ \cE( \hat f_\epsilon) & \leq C \left[\min\left( n_P^{-\tfrac{\alpha(\beta +1)}{2 \alpha + \alpha \beta + \gamma(\epsilon)}}, n_Q^{-\tfrac{\alpha(\beta+1)}{2\alpha + \alpha \beta + d}} \right)  + \min\left(  \frac{1}{n_P} \epsilon^{-(1 + \gamma^*/\alpha)}, \frac{1}{n_Q} \epsilon^{-(1 + d/\alpha)} \right)  \right]. 
\end{align*}
\end{corollary}

% CAN USE \right. and \left. to get the square brackets on two lines for the two-col. version. 

%\skr{there's a disconnect here:} Although the hybrid model $\hat f_\epsilon$ is idealized, we can in fact show that the the excess risk of the adaptive method that makes local depth choices can be bounded, up to log terms, by the same quantity - for \it any \rm  $\epsilon$. 

%%%% statement here in main body

% threshold: \epsilon \geq 4 C_\alpha (n_P + n_Q)^{-\alpha}
\begin{theorem}\label{adaptiveAttainsEpsilonBound}
Let $\hat f$ denote the adaptive classifier of Algorithm \ref{lepskiAlg}, and suppose that $\alpha + d > 1$. Then there are constants $C_1,C_2$, depending only on $\cT$, such that the following holds. If $\max(n_P,n_Q) \geq C_1$, then for any $\epsilon  > 0$ we have  
\begin{align} \E \, [\cE(\hat{f})]  \leq  C_2 \left[ \min \left(\left(\frac{\log n_P }{n_P}\right)^{\tfrac{\alpha(\beta +1)}{2 \alpha + \alpha \beta + \gamma(\epsilon)}}, \left(\frac{\log  n_Q }{n_Q}\right)^{\tfrac{\alpha(\beta+1)}{2\alpha + \alpha \beta + d}} \right) + \min\left(  \frac{1}{n_P} \epsilon^{-(1 + \gamma^*/\alpha)}, \frac{1}{n_Q} \epsilon^{-(1 + d/\alpha)} \right) \right].  \label{highMargin}
\end{align}
\end{theorem}

\begin{remark} Note that in the proof of Theorem \ref{adaptiveAttainsEpsilonBound}, we in fact consider only $\epsilon$ bounded away from zero: indeed, we assume that $\epsilon \geq 4 C_\alpha (n_P + n_Q)^\alpha$. We show however that this introduces no loss of generality, since taking $\epsilon$ below this threshold gives a vacuous excess risk bound. 
\end{remark}

The proof of Theorem \ref{adaptiveAttainsEpsilonBound} can now be completed using techniques used to prove Theorem \ref{adaptiveTheorem}; details are in Appendix B. Theorem \ref{localized_adaptive_theorem} now follows immediately.

\begin{proof}{[Theorem \ref{localized_adaptive_theorem}]}
It is easy to see that the function $\gamma(\epsilon)$ is right-continuous, and therefore by continuity we have $$ \frac{1}{n_P}(\epsilon_n)^{-(1 + \gamma^*/\alpha)} \leq \left( \frac{\log n_P}{n_P} \right)^{\tfrac{\alpha(\beta +1)}{2 \alpha + \alpha \beta + \gamma_n}}.$$ Letting $C$ be the constant appearing in Theorem \ref{adaptiveAttainsEpsilonBound} and bounding the minimums in \eqref{highMargin} by their first arguments yields $$ \E_{(\bX,\bY)}[\cE(\hat{f})] \leq 2 C  \left(\frac{\log n_P }{n_P}\right)^{\tfrac{\alpha(\beta +1)}{2 \alpha + \alpha \beta + \gamma_n}} .$$ Taking $\epsilon = 1$ in Theorem \ref{adaptiveAttainsEpsilonBound} and bounding the minimums by their second arguments gives $$ \E_{(\bX,\bY)}[\cE(\hat{f})] \leq 2 C  \left(\frac{\log n_Q }{n_Q}\right)^{\tfrac{\alpha(\beta +1)}{2 \alpha + \alpha \beta + d}}, $$ and combining these bounds gives the result. 
\end{proof}

\section{Experiments On Real Data}\label{secExperiments}

\begin{figure*}[t]
    \centering
    \includegraphics[height = 11cm]{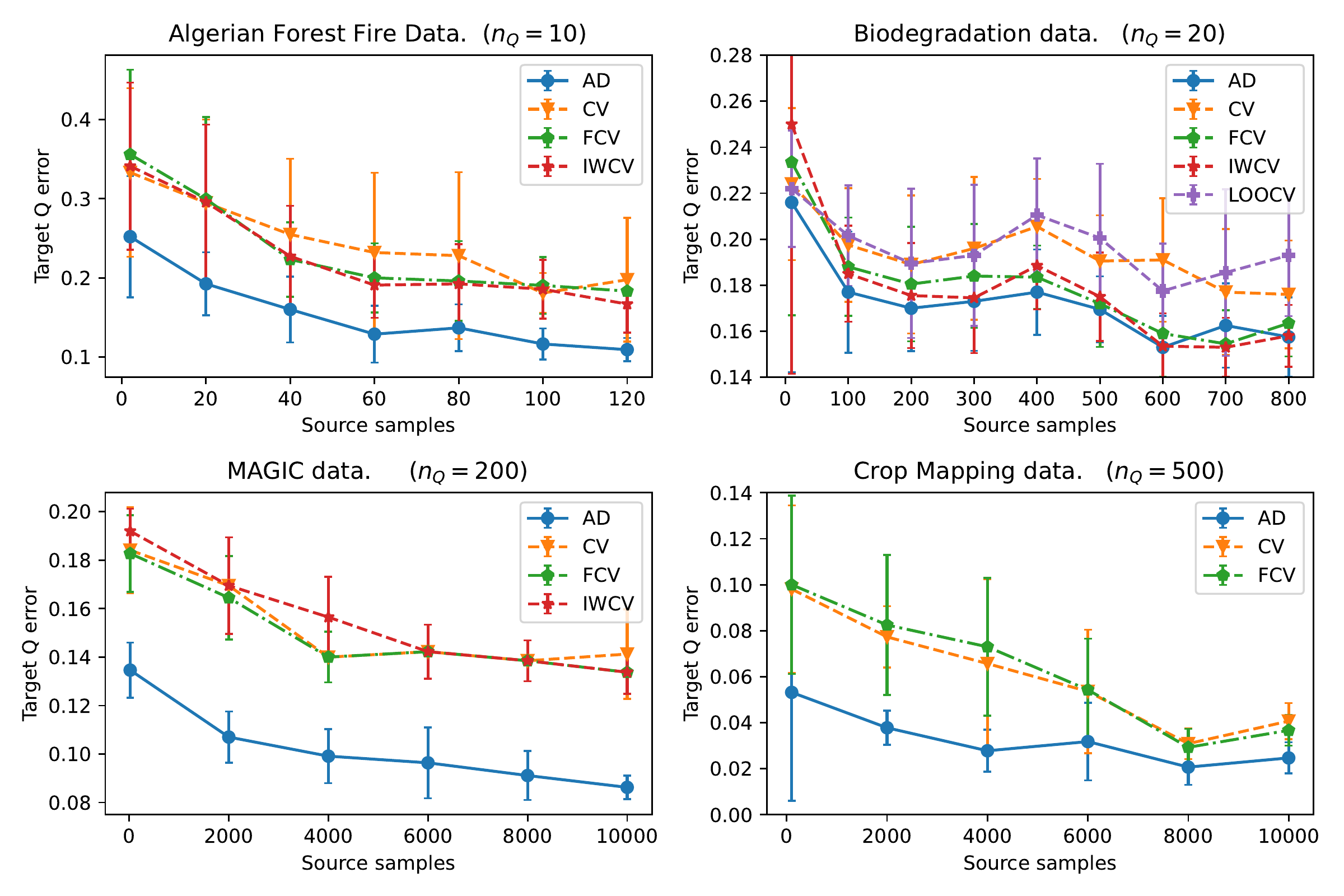}
    \caption{Target risk estimates for four tree-pruning methods used on dyadic trees applied respectively to the Algerian Forest Fire data ($d = 11$), the MAGIC data ($d = 10$), the SUSY data ($d = 18$), and the Biodegradation data ($d = 41$). Error bars show the standard errors over 10 iterations (except on the Forest Fire data, where we use 50 iterations). AD (blue) indicates the adaptive method, CV (yellow) is 2-fold cross validation performed by taking the hold-out risk on target samples only, FCV (green) is 2-fold cross validation performed by taking the hold-out risk on the combined sample, and IWCV is the importance weighted CV method, where the  hold-out risk is taken over all points, and at source points it is weighted by the density ratio of the features at those points. LOO denotes leave-one-out cross-validation.}
    \label{fig:experiments_main}
\end{figure*}

In this section we carry out a brief sequence of experiments using real-world datasets to illustrate the effectiveness of our proposed adaptive method for tree-pruning as opposed to alternative methods under covariate-shift based on empirical risk minimization. In particular, in each experiment, we grow a slight variant of the dyadic tree \eqref{treeClassifier} in which nodes are split by randomly choosing a feature along which the node in question is longest, so that the ratio of the longest to shortest edge never exceeds 2; we note that it is easy to see that the analysis that we have undertaken for the regular dyadic tree applies to these cyclical trees as well (see Corollary \ref{cyclicalTreeRates} in appendix \ref{appendixProofs}), so this procedure inherits the convergence rates from Theorem \ref{adaptiveTheorem}. \par 
%or a tree grown using the vanilla CART algorithm of \cite{CART} (that is, splits are chosen to minimize Gini impurity at each node).
We contrast the effectiveness of locally selecting the tree depth adaptively via Algorithm \ref{lepskiAlg} (AD) with three alternative strategies, described below. 

% comment somewhere on the value of C

\begin{table*}[t]
\caption{Datasets.} % title name of the table
\centering % centering table
\begin{tabular}{c p{9cm} c} % creating 10 columns
\hline\hline % inserting double-line
\\ [0.05ex]
 Dataset   & \hspace{2cm} Source/Target Split & Test Size 
\\ [0.5ex] \hline % inserts single-line
% Entering 1st row
% & &soft &1 & $-1$ & 1 & 1 & $-1$ & $-1$ & 1 \\[-1ex]
%\raisebox{1.5ex}{Police} & \raisebox{1.5ex}{5}&hard
%& 2 & $-4$ & 4 & 4 & $-2$ & $-4$ & 4 \\[1ex]
\\ [0.5ex]
\makecell{Algerian Forest Fires ($d = 11$) \\ See \cite{forestFireData}.} & Observations from Sidi bel-Abbas constitute the Source sample, while the Target data are observations from the Bejaia region. & $100$ \\[0.5ex] 
\hline
\\[0.5ex]
\makecell{MAGIC Gamma ($d = 10$) \\ See \cite{magicGammaData}.} & Three variable indices ($2,7$, and $9$) were chosen at random. Observations with normalized values of these variables each above $0.3$ go in the Target set with probability $p = 0.95$; the rest go into the Source set with probability $p$. & $2500$ \\[.5ex]
\hline
\\[0.5ex]
\makecell{SUSY ($d = 18$) \\  See \cite{SUSYData}.} & Three variable indices ($5,10$, and $15$) were chosen at random. Observations with normalized values of these variables each above $0.1$ go in the Target set with probability $p = 0.95$; the rest go into the Source set with probability $p$.  & $5000$ \\[0.5ex]
\hline
\\[0.5ex]
\makecell{Wine Quality ($d = 11$) \\ See \cite{wineQualityData}.}   & White wine data is used as Source, red wine as Target. & $1000$ \\[0.5ex]
\hline
\\[0.5ex]
\makecell{Biodegradation ($d = 41$) \\ See \cite{biodegradationData}.}  &   Observations with more than $75\%$ Carbon atoms go in the Target set with probability $p = 0.95$; those with less go in the Source set with probability $p$. & $200$ \\[0.5ex]
\hline
\\[0.5ex]
\makecell{Crop Mapping ($d = 174$) \\ See \cite{cropmapping}.} &  Task is to discriminate between Canola and Soybeans. Two variable indices ($2$ and $13$) were chosen at random. Observations with normalized values of these variables each above $0.25$ go in the target set with probability $p = 0.95$; the rest go into the source set with probability $p$. & $5000$ \\[0.5ex]
\hline % inserts single-line
\end{tabular}
\label{tab:datasets}
\end{table*}

\subsection{Baselines.}

We implement three cross-validation based pruning methods to serve as baselines against which to compare Algorithm \ref{lepskiAlg}. For all three of the cross-validation methods (CV, FCV, IWCV), a base (cyclical dyadic) tree is grown. We only cross-validate over the $O(\log n)$ subtrees defined by levels of the base tree as opposed to all subtrees (the number of such subtrees is super-exponential in the depth of the base tree, which is computationally infeasible for cross-validation since in our case this would be $O(2^n)$). For all three methods, we use 2 folds, chosen such that the proportion of source and target samples is the same in each fold. For each fold, the train portion is used to fit the classifiers, while the hold-out risk is then computed for the test portion. The tree level with minimal averaged hold-out risk is chosen, and the classifier based on this tree is re-fit on the full training sample and the error is reported on the target validation set.  The three methods implement three natural ways of computing the hold-out risk $\hat R$ in practice:  
    \begin{itemize}
        \item {\bf CV:} $\hat R$ is the empirical risk over the target samples in the fold (CV). That is, $\hat R_{CV} = \sum_{i=1}^{n_Q} \ind\{Y_i \neq f^{(i)}(X_i)\}$, where it is understood that the $f^{(i)}$ in each summand is the classifier trained on the segment of data that excludes the indexed data point $(Y_i, X_i)$.  
        \item {\bf FCV:} In this case, $\hat R$ is the empirical risk over all samples (source and target) in the fold (FCV); that is, $\hat R_{FCV} = \sum_{i=1}^{n_P + n_Q} \ind\{Y_i \neq f^{(i)}(X_i)\}$. 
        \item {\bf IWCV:} In this case, we take $\hat R$ to be the empirical risk over all samples in the fold, where the errors from the source samples are importance weighted according to a local estimate of the feature density ratio between source and target (IWCV), that is, 
        \begin{equation*}
        \hat R_{IWCV} = \frac{1}{n_P + n_Q} \left( \sum_{i = 1}^{n_P} 
        \hat \Lambda(X_i) \cdot \mathbf{1}\{Y_i \neq f^{(i)}(X_i)\} + \sum_{i = n_P + 1}^{n_P + n_Q} \mathbf{1}\{Y_i \neq f^{(i)}(X_i)\}\right),
        \end{equation*} with $\hat \Lambda(X_i)$ being an estimate of the density ratio $q_{X}(X_i)/p_{X}(X_i)$ at the point $X_i$. Since $\hat R_{IWCV}$ is unbiased for $R_Q$, it may be possible to show that $IWCV$ can achieve optimal rates. However, the method is obviously sensitive to the quality of the density ratio estimates $\hat \Lambda$, and in particular rules out the possibility of being implemented in cases with $n_Q$ being very small (or possibly zero). In our experiments, we compute $\hat \Lambda$ in advance using \emph{all available samples}, that is, all samples given in the full dataset (including validation set) even when we report performance for lower sample size settings, which implies that our comparisons are in fact overly favourable to IWCV. 
    \end{itemize}

To estimate the density ratios required by the IWCV method, we use the RuLSIF method (see \cite{densRatio}) as implemented in the associated \texttt{densratio} package. 
Note that density ratios for the IWCV method are pre-computed using \emph{both full samples} (that is, using all the data at our disposal, even when we report error rates for the methods learned from smaller subsamples) to get the best possible estimates. All the datasets are preprocessed by scaling the features so that they lie in $[0,1]^d$.

%We grow a baseline cyclical dyadic tree, and the maximal tree depth is treated as a model parameter to be selected) by a) cross validation (CV) in which the hold-out sets are made by splitting the target sample points, over which we compute the empirical risk (in one experiment we also implement the leave-one-out cross-validation (LOO), where we fit $n_Q$ times in the model-selection stage), b) cross validation performed as usual over the full (source and target) sample (FCV), and c) importance-weighted cross validation (IWCV), in which the hold-out sets use the whole sample, but the empirical risk is calculated by assigning a weight to the risk contribution of each source point according to the estimated density ratio of the source/target distributions at that point; see \cite{IWCV} for details and discussion.

\subsection{Datasets.}

Each of the datasets that we use can be found on the UCI machine learning respository \cite{UCIrepository}. See Table \ref{tab:datasets} for details and references.

\subsection{Results.}
Figure \ref{fig:experiments_main}  display the empirical target risk for the aforementioned methods at various source/target sample size configurations for the datasets in Table \ref{tab:datasets}. At each of the chosen sample size settings, we use the three cross-validation based methods to perform model selection, and then grow a single tree from which the predictions for all four models are computed for the target test set. This is averaged over 10 iterations for each setting (except for the Algerian Forest Fire data, where we use 50 iterations); for each iteration, a source/target sample of the required size is taken uniformly at random from all samples available. 

\par Immediately we observe that in all configurations, on each dataset, the adaptive method performs at least as well as the empirical-risk based methods, and in certain settings considerably better. On the Algerian Forest Fire data (Figure \ref{fig:experiments_main}, top left), the MAGIC data (Figure \ref{fig:experiments_main}, bottom left) and the Crop Mapping data (Figure \ref{fig:experiments_main}, bottom right) the gulf in performance is considerable, while for the Biodegradation data (Figure \ref{fig:experiments_main}, top right), the adaptive method is roughly matched in performance by the cross-validation approaches that use the full sample to compute the empirical risk (FCV and IWCV), although we note that even in this setting where the adaptive method does not seem to offer much improvement, it nevertheless performs no worse than the baselines. It should of course be remembered that the AD method is computationally much cheaper to implement, since it does not require multiple trees to be grown for model selection purposes, nor does it require the calculation of any empirical risks.

%In many practical problems, particularly in higher dimensions, such trees may exhibit poor performance, although the results that we find here suggest that it may be fruitful to implement the adaptive local depth selection of Algorithm \ref{lepskiAlg} on individual dyadic trees within an ensemble method, such as forests of extremely random trees (see \cite{ERT}). 

 %\skr{For example, on the SUSY data with $n_P = 10 \, 000$, $n_Q = 1000$, and $n_{test} = 5000$, to perform depth-selection, grow a tree, and compute predictions for the test set took $1.6\mathrm{s} \pm 82\mathrm{ms}$ seconds for AD, $4.06\mathrm{s} \pm 184\mathrm{ms}$ for CV, and $102\mathrm{s} \pm 1.78\mathrm{s}$ for IWCV, respectively (and this \it does not \rm include the time required to estimate the density ratios required for the IWCV method). } \sk{Should I indicate my computer specs?} 
\par All the experiments above use a value of $C_{n_P, n_Q} \equiv 1/4$ when computing $\hat \sigma$ (via \eqref{sigHat1}) for the AD method, which was found to perform reasonably well across a wide variety of settings. At the price of increasing computation time by a constant factor, this quantity could be straightforwardly selected in a data-dependent way by simply doing cross-validation over hold-out sets of the target sample, although we do not pursue this here (\cite{reeve2021adaptive} demonstrate that, for a similar nearest-neighbours based procedure, this strategy does not affect the attained rates).  Further, we note as well that the error bars tend to be considerably smaller for the AD method than for the cross-validation based methods, especially when the ratio of source/target samples is large. \par Finally, we remark that while the CV method generally exhibits worse performance than the IWCV method (as expected), there is almost no difference between the models selected with IWCV and with the full pooled sample cross-validation (FCV). We suspect that this is an artefact of the datasets we have chosen; \cite{IWCV} demonstrate examples where FCV performs considerably more poorly than IWCV. Our results nevertheless indicate that using ICI for tree pruning may be fruitful even in the single population case $P_X = Q_X$, an indication that is further reinforced by supplementary experiments in Appendix \ref{appendix_experiments}.
\par In Appendix \ref{appendix_experiments}, we also provide further empirical results indicating that even when the trees are grown using a greedy procedure, the adaptive depth selection (AD) method can still yield a uniform improvement over the cross-validation based model selection procedures, albeit by a seemingly smaller margin.

%%% have to briefly comment on the IWCV / FCV

%\par .  The performance gain is least pronounced when there is almost no source data, %suggesting that much of the observed difference in performance is due to the advantageous nature of %adaptive local methods for learning under covariate-shift, rather than being an artefact of the %limited power of the underlying balanced dyadic tree (although this is conspicuously not the case for %the MAGIC data). Another feature that we observe - most saliently in the second plot from the top of %Figure \ref{fig:wineData} - is the relatively poor performance of the importance-weighted %cross-validation (IWCV). 

%\section{Conclusion} \label{sec_conclusion}
%The conclusion goes here.

% or
% do not use \section anymore after \appendix, only \section


\bibliographystyle{plain}
\bibliography{Main_ARXIV_V2}

\begin{thebibliography}{10}

\bibitem{forestFireData}
Faroudja Abid and Nouma Izeboudjen.
\newblock {\em Predicting Forest Fire in Algeria Using Data Mining Techniques:
  Case Study of the Decision Tree Algorithm}, pages 363--370.
\newblock 02 2020.

\bibitem{Audibert2007}
Jean-Yves Audibert and Alexandre~B. Tsybakov.
\newblock Fast learning rates for plug-in classifiers.
\newblock {\em The Annals of Statistics}, 35(2):608--633, April 2007.

\bibitem{SUSYData}
P.~Baldi, P.~Sadowski, and D.~Whiteson.
\newblock Searching for exotic particles in high-energy physics with deep
  learning.
\newblock {\em Nature Communications}, 5(1), July 2014.

\bibitem{impossibility_theorems_domain_adaptation}
Shai Ben-David, Tyler Lu, Teresa Luu, and D\`avid P\`al.
\newblock Impossibility theorems for domain adaptation.
\newblock {\em Journal of Machine Learning Research - Proceedings Track},
  9:129--136, 01 2010.

\bibitem{BERTSEKAS1982}
Dimitri~P. Bertsekas.
\newblock {\em Constrained Optimization and Lagrange Multiplier Methods}.
\newblock Academic Press, 1982.

\bibitem{ODDT}
Gilles Blanchard, C.~Schäfer, Yves Rozenholc, and Klaus-Robert Müller.
\newblock Optimal dyadic decision trees.
\newblock {\em Machine Learning}, 66:209--241, 03 2007.

\bibitem{bounds_alg_DDT}
Gilles Blanchard, Christin Schäfer, and Yves Rozenholc.
\newblock Oracle bounds and exact algorithm for dyadic classification trees.
\newblock volume 3120, pages 378--392, 07 2004.

\bibitem{magicGammaData}
R.K. Bock, A.~Chilingarian, M.~Gaug, F.~Hakl, T.~Hengstebeck, M.~Jiřina,
  J.~Klaschka, E.~Kotrč, P.~Savický, S.~Towers, A.~Vaiciulis, and W.~Wittek.
\newblock Methods for multidimensional event classification: a case study using
  images from a cherenkov gamma-ray telescope.
\newblock {\em Nuclear Instruments and Methods in Physics Research Section A:
  Accelerators, Spectrometers, Detectors and Associated Equipment},
  516(2):511--528, 2004.

\bibitem{CART}
L.~Breiman, Jerome~H. Friedman, Richard~A. Olshen, and C.~J. Stone.
\newblock {\em Classification and Regression Trees}.
\newblock Wadsworth \& Brooks/Cole Advanced Books \& Software., 1983.

\bibitem{Cai2021}
T.~Tony Cai and Hongji Wei.
\newblock Transfer learning for nonparametric classification: Minimax rate and
  adaptive classifier.
\newblock {\em Annals of Statistics}, 49(1):100--128, February 2021.

\bibitem{clarkson2006nearest}
Kenneth~L Clarkson.
\newblock Nearest-neighbor searching and metric space dimensions.
\newblock {\em Nearest-neighbor methods for learning and vision: theory and
  practice}, pages 15--59, 2006.

\bibitem{wineQualityData}
Paulo Cortez, António Cerdeira, Fernando Almeida, Telmo Matos, and José Reis.
\newblock Modeling wine preferences by data mining from physicochemical
  properties.
\newblock {\em Decision Support Systems}, 47(4):547--553, 2009.
\newblock Smart Business Networks: Concepts and Empirical Evidence.

\bibitem{CoverThomas}
Thomas~M. Cover and Joy~A. Thomas.
\newblock {\em Elements of Information Theory (Wiley Series in
  Telecommunications and Signal Processing)}.
\newblock Wiley-Interscience, USA, 2006.

\bibitem{Devroye1996}
Luc Devroye, L{\'{a}}szl{\'{o}} Gy\"{o}rfi, and G{\'{a}}bor Lugosi.
\newblock {\em A Probabilistic Theory of Pattern Recognition}.
\newblock Springer New York, 1996.

\bibitem{UCIrepository}
Dheeru Dua and Casey Graff.
\newblock {UCI} machine learning repository, 2017.

\bibitem{Falconer2014}
Kenneth Falconer.
\newblock {\em Fractal Geometry: Mathematical Foundations and Applications}.
\newblock Wiley, 2014.

\bibitem{ICIGoldNem}
Alexander Goldenshluger and Arkadi Nemirovski.
\newblock On spatial adaptive estimation of nonparametric regression.
\newblock {\em Mathematical Methods of Statistics}, 6, 01 1997.

\bibitem{Gretton2009CovariateSB}
Arthur Gretton, Alex Smola, Jiayuan Huang, Marcel Schmittfull, Karsten~M.
  Borgwardt, Bernhard Sch{\"o}lkopf, Qui{\~n}onero Candela, Masashi Sugiyama,
  Anton Schwaighofer, and Neil~D. Lawrence.
\newblock Covariate shift by kernel mean matching.
\newblock In {\em NIPS 2009}, 2009.

\bibitem{CoincidenceOfDims}
Moisey Guysinsky and Serge Yaskolko.
\newblock Coincidence of various dimensions associated with metrics and
  measures on metric spaces.
\newblock {\em Discrete \& Continuous Dynamical Systems - A}, 3:591, 1997.

\bibitem{gyorfiNonparametrics}
László Györfi, Michael Kohler, Adam Krzyzak, and Harro Walk.
\newblock {\em A Distribution-Free Theory of Nonparametric Regression.}
\newblock Springer series in statistics. Springer, 2002.

\bibitem{HK_val_targ_data}
Steve Hanneke and Samory Kpotufe.
\newblock On the value of target data in transfer learning.
\newblock In {\em NeurIPS}, 2019.

\bibitem{hush2010algorithms}
Don Hush and Reid Porter.
\newblock Algorithms for optimal dyadic decision trees.
\newblock {\em Machine learning}, 80(1):85--107, 2010.

\bibitem{Lee_MetaLearning2020}
Kaiyi Ji, Jason~D Lee, Yingbin Liang, and H.~Vincent Poor.
\newblock Convergence of meta-learning with task-specific adaptation over
  partial parameters.
\newblock In H.~Larochelle, M.~Ranzato, R.~Hadsell, M.F. Balcan, and H.~Lin,
  editors, {\em Advances in Neural Information Processing Systems}, volume~33,
  pages 11490--11500. Curran Associates, Inc., 2020.

\bibitem{cropmapping}
Iman Khosravi and Seyed~Kazem Alavipanah.
\newblock A random forest-based framework for crop mapping using temporal,
  spectral, textural and polarimetric observations.
\newblock {\em International Journal of Remote Sensing}, 40(18):7221--7251,
  2019.

\bibitem{KM}
Samory Kpotufe and Guillaume Martinet.
\newblock Marginal singularity, and the benefits of labels in covariate-shift.
\newblock In {\em Annals of Statistics}, (To appear) 2021.

\bibitem{Lepski1997}
O.~V. Lepski and V.~G. Spokoiny.
\newblock Optimal pointwise adaptive methods in nonparametric estimation.
\newblock {\em The Annals of Statistics}, 25(6):2512--2546, December 1997.

\bibitem{DA_bounds_and_algorithms_Mohri}
Yishay Mansour, Mehryar Mohri, and Afshin Rostamizadeh.
\newblock Domain adaptation: Learning bounds and algorithms.
\newblock In {\em Proceedings of The 22nd Annual Conference on Learning Theory
  (COLT 2009)}, Montr\'eal, Canada, 2009.

\bibitem{biodegradationData}
Kamel Mansouri, Tine Ringsted, Davide Ballabio, Roberto Todeschini, and Viviana
  Consonni.
\newblock Quantitative structure–activity relationship models for ready
  biodegradability of chemicals.
\newblock {\em Journal of Chemical Information and Modeling}, 53(4):867--878,
  2013.
\newblock PMID: 23469921.

\bibitem{histRuleConcentration}
David McAllester and Luis Ortiz.
\newblock Concentration inequalities for the missing mass and for histogram
  rule error.
\newblock {\em Journal of Machine Learning Research}, pages 895--911, 2003.

\bibitem{mcdiarmid_1989}
Colin McDiarmid.
\newblock {\em On the method of bounded differences}, page 148–188.
\newblock London Mathematical Society Lecture Note Series. Cambridge University
  Press, 1989.

\bibitem{Olsen2005}
L.~Olsen.
\newblock Typical lq-dimensions of measures.
\newblock {\em Monatshefte f\"{u}r Mathematik}, 146(2):143--157, September
  2005.

\bibitem{pathak2022new}
Reese Pathak, Cong Ma, and Martin~J. Wainwright.
\newblock A new similarity measure for covariate shift with applications to
  nonparametric regression, 2022.

\bibitem{Pesin1997}
Yakov~B. Pesin.
\newblock {\em Dimension Theory in Dynamical Systems}.
\newblock University of Chicago Press, 1997.

\bibitem{reeve2021adaptive}
Henry W.~J. Reeve, Timothy~I. Cannings, and Richard~J. Samworth.
\newblock Adaptive transfer learning, 2021.

\bibitem{DDT}
C.~Scott and R.D. Nowak.
\newblock Minimax-optimal classification with dyadic decision trees.
\newblock {\em {IEEE} Transactions on Information Theory}, 52(4):1335--1353,
  April 2006.

\bibitem{Scott_DomainAdaptation}
Clayton Scott.
\newblock A generalized neyman-pearson criterion for optimal domain adaptation.
\newblock In Aurélien Garivier and Satyen Kale, editors, {\em Proceedings of
  the 30th International Conference on Algorithmic Learning Theory}, volume~98
  of {\em Proceedings of Machine Learning Research}, pages 738--761. PMLR,
  22--24 Mar 2019.

\bibitem{Scott2002DyadicCT}
Clayton~D. Scott and Robert~D. Nowak.
\newblock Dyadic classification trees via structural risk minimization.
\newblock In {\em NIPS}, 2002.

\bibitem{SteinBook}
Elias~M. Stein and Timothy~S. Murphy.
\newblock {\em Harmonic Analysis (PMS-43): Real-Variable Methods,
  Orthogonality, and Oscillatory Integrals. (PMS-43)}.
\newblock Princeton University Press, 1993.

\bibitem{stone1982}
Charles~J. Stone.
\newblock Optimal global rates of convergence for nonparametric regression.
\newblock {\em Ann. Statist.}, 10(4):1040--1053, 12 1982.

\bibitem{IWCV}
Masashi Sugiyama, Matthias Krauledat, and Klaus-Robert Müller.
\newblock Covariate shift adaptation by importance weighted cross validation.
\newblock {\em Journal of Machine Learning Research}, 8:985--1005, 05 2007.

\bibitem{densRatio}
Masashi Sugiyama, Shinichi Nakajima, and Hisashi Kashima.
\newblock Direct importance estimation with model selection and its application
  to covariate shift adaptation, 01 2007.

\bibitem{subtrees_of_trees}
L.A. Székely and Hua Wang.
\newblock On subtrees of trees.
\newblock {\em Advances in Applied Mathematics}, 34(1):138--155, 2005.

\bibitem{Jordan_MetaLearning2021}
Nilesh Tripuraneni, Chi Jin, and Michael~I. Jordan.
\newblock Provable meta-learning of linear representations.
\newblock In {\em ICML}, 2021.

\bibitem{WassNonparametrics}
Larry Wasserman.
\newblock {\em All of Nonparametric Statistics}.
\newblock Springer New York, 2006.

\end{thebibliography}

%\appendices
\appendix

\section{Properties of the Aggregate Transfer Exponent.} \label{appendix_properties}
In this section we prove some of the claims from Section \ref{secOnDim}, and collect some further simple results concerning the relative dimension and its relation to previously studied notions of dimension for measures.

\begin{proof}{\textit{(Proposition \ref{sharperBounds}.)}}
 Using the definition of transfer exponent \eqref{transferExp0}, we see that 

%UNCOMMENT FOR ONE-COLUMN FORMAT:
 \begin{align*}
     \int \frac{1}{P(B(x,r))} \, dQ(x) = \int \left( \frac{Q(B(x,r))}{P(B(x,r))} \right)\frac{1}{Q(B(x,r))} \, dQ(x)  \leq C_\rho r^{-\rho} \int \frac{1}{Q(B(x,r))} \, dQ(x).
\end{align*} 

%UNCOMMENT FOR TWO-COLUMN FORMAT:
% \begin{align*}
%    & \int \frac{1}{P(B(x,r))} \, dQ(x) \\ & \quad = \int \left( \frac{Q(B(x,r))}{P(B(x,r))} \right)\frac{1}{Q(B(x,r))} \, dQ(x) \\& \quad  \leq C_\rho r^{-\rho} \int \frac{1}{Q(B(x,r))} \, dQ(x).
%\end{align*} 

Let $\cZ$ be a minimal $r/2$ cover of $\cX_Q$, so that by assumption \eqref{QmetricEntropy} we have $|\cZ| \leq C_d r^{-d}$. Now, if $x \in B(z,r/2)$, then clearly $B(z,r/2) \subset B(x,r)$, so we have 
\begin{align*}
    \int \frac{1}{Q(B(x,r))} \, dQ(x)  \sum_{z \in \cZ} \int_{B(z,r/2)} \frac{1}{Q(B(x,r))} \, dQ(x)  \leq \sum_{z \in \cZ} \int_{B(z,r/2)} \frac{1}{Q(B(z,r/2)} \, dQ(x) = |\cZ| \leq C_d r^{-d},
\end{align*} so $\rho + d$ is an integrated transfer exponent from $P$ to $Q$, and the result follows. 
\end{proof}

We now provide a simple Lemma giving an equivalent definition of the relative Minkowski dimension, which will lead us to a proof of Proposition \ref{relDimEquivalence}. 

\begin{lemma}\label{gammaStarIsMinkowki}
Let $\gamma_*(P,Q)$ be as in \eqref{gammaStar}. Let $\Xi(r)$ denote the set of $r$-grids for $\cX_Q$ (see Definition \ref{gridDef}). For an $r$-grid $\xi \in \Xi(r)$, let \begin{equation*}
  \Lambda(\xi,r) = \sum_{E \in \xi} \frac{Q(E)}{P(E)},
\end{equation*} and let $\Lambda(r) = \sup_{\xi \in \Xi(r)} \Lambda(\xi,r)$. Then 
\begin{equation*}%\label{relMinkowskiLimit}
    \gamma_*(P,Q) = \limsup_{r \to 0} \frac{\log \Lambda(r)}{- \log r}.
\end{equation*} 
\end{lemma}
\begin{proof}
If $\gamma \in \{\gamma \geq 0: \, \exists C_\gamma \text{ s.t. } \eqref{globalTransferExp} \text{ holds.}\}$, then $\Lambda(r) \leq  C_\gamma r^{-\gamma}$ and clearly then $$ \limsup_{r \to 0} \frac{\log \Lambda(r)}{- \log r} \leq \gamma,$$ whence $\limsup_{r \to 0} -\log \Lambda(r) / \log (r) \leq \gamma_*$.  On the other hand, if $$\limsup_{r \to 0} \frac{\log \Lambda(r)}{- \log r}= s,$$ then if $\epsilon > 0$ we have for small enough $r$ that $-\log \Lambda(r)/\log r \leq \gamma + \epsilon$, so there is some $C$ such that $\Lambda(r) \leq C r^{-(s + \epsilon)}$, so $\gamma_* \leq s + \epsilon$, and $\epsilon$ is arbitrary so $$ \gamma_*(P,Q) \leq \limsup_{r \to 0} \frac{\log \Lambda(r)}{- \log r},$$ and this completes the proof. 
\end{proof}

\begin{proof}{\textit{(Proposition \ref{rhoStarLowerBound})}}
 Fix an arbitrary $\epsilon > 0$, so that $\rho - \epsilon$ is not a transfer exponent from $P$ to $Q$. In particular, for each $C > 0$, there is an $x \in \cX_Q$ and an $r \in (0,1)$ such that \begin{align*}
     \frac{Q(B(x,r))}{P(B(x,r))} \geq C r^{-(\rho - \epsilon)}. 
 \end{align*} Set $C = 1$, and pick a sequence $(x_n, r_n) \in \cX_Q \times (0,1)$ with $r_n \to 0$ such that 
 \begin{align*}
     \frac{Q(B(x_n,r_n))}{P(B(x_n,r_n))} \geq  r_n^{-(\rho - \epsilon)}. 
\end{align*} Note that we can certainly ensure that we can pick $r_n \to 0$, since if that were not the case it would imply that there was an $r_0 > 0$ such that for all $x \in \cX_Q$ and $r \in (0,r_0)$ we had $P(B(x, r)) \geq r^{(\rho - \epsilon)} Q(B(x,r))$. We could then let 
\begin{align*}
    C \defeq \sup_{r \in [r_0, 1), x \in \cX_Q} r^{(\rho - \epsilon)}\frac{Q(B(x,r))}{P(B(x,r))} \geq 1 
\end{align*} we would have $P(B(x,r)) \geq C^{-1} r^{-(\rho - \epsilon)} Q(B(x,r))$ for all $x \in \cX_Q$, $r \in (0,1)$ (of course, we have $C < \infty$), and this would contradict the fact that $\rho - \epsilon$ was not a transfer exponent from $P$ to $Q$. Now, let $C_{2Q}$ be the doubling constant of $Q$, so that $Q(B(x,r)) \geq C_{2Q} Q(B(x,2 r))$ for all $x \in \cX_Q$ and $r > 0$. For any $n \geq 1$, we have

%UNCOMMENT FOR ONE-COLUMN FORMAT:
 \begin{align*}
     \varphi(r_n / 2) &= \int \frac{1}{P(B(x,r_n/2))} \,  Q(dx)
     \geq \int_{B_{r_n/2}(x_n)} \frac{1}{P(B(x,r_n/2))} \, Q(dx)
      \geq \int_{B_{r_n/2}(x_0)} \frac{1}{P(B(x_n,r_n))} \, Q(dx)
     \\& = \frac{Q(B(x_n, r_n/2))}{P(B(x_0, r_n))}
   = \frac{Q(B(x_n, r_n/2))}{Q(B(x_0, r_n))} \frac{Q(B(x_n, r_n))}{P(B(x_n, r_n))}
     \geq C_{2Q}\, r_n^{-(\rho - \epsilon)}, 
 \end{align*}

%UNCOMMENT FOR TWO-COLUMN FORMAT:
% \begin{align*}
%     \varphi(r_n / 2) &= \int \frac{1}{P(B(x,r_n/2))} \,  Q(dx)
%     \\& \geq \int_{B_{r_n/2}(x_n)} \frac{1}{P(B(x,r_n/2))} \, Q(dx)
%     \\& \geq \int_{B_{r_n/2}(x_0)} \frac{1}{P(B(x_n,r_n))} \, Q(dx)
%     \\& = \frac{Q(B(x_n, r_n/2))}{P(B(x_0, r_n))}
%     \\& = \frac{Q(B(x_n, r_n/2))}{Q(B(x_0, r_n))} \frac{Q(B(x_n, r_n))}{P(B(x_n, r_n))}
%     \\& \geq C_{2Q}\, r_n^{-(\rho - \epsilon)}, 
%\end{align*} 
 
 and it follows that $\bar \rho^* \geq \rho - \epsilon$. Since $\epsilon > 0$ was arbitrary, the claim follows.  
 \end{proof}

\begin{proof}{\textit{(Proposition \ref{relDimEquivalence}.)}}
Pick an arbitrary $r$-grid $\xi$ of $\cX_Q$, and let $$\varphi(r) = \int P(B(x,r))^{-1} \, dQ(x).$$ Since for $E \in \xi$ there is an $x_E \in \cX$ such that $E \subset B(x_E,2r)$, clearly $\cX \subset \cup_{E \in \xi} B(x_E,2r)$ and so  
\begin{align*} 
\varphi(4r)  = \int \frac{1}{P(B(x,4r))} \, dQ(x) 
     \leq \sum_{E \in \xi} \int_{B(x_E,2r)}  \frac{1}{P(B(x,4r))} \, dQ(x),
\end{align*} and for $x \in B(x_E,2r)$ we have $B(x_E,2r) \subset B(x,4r)$, and therefore
\begin{align*}
    \varphi(4r) \leq \sum_{E \in \xi} \int_{B(x_E,2r)} \frac{1}{P(B(x_E,2r))} \, dQ(x)
     = \sum_{E \in \xi} \frac{Q(B(x_E,2r))}{P(B(x_E,2r))}.
    %\\&= \Lambda(\mathcal{Z},r)
\end{align*} Now, since $Q$ is a doubling measure there is a constant $C_{2Q}$ such that $Q(B(x,r)) \geq C_{2Q} Q(B(x,2r))$ for all $x \in \cX_Q$, so 
\begin{equation*}
    \frac{Q(B(x_E,2r))}{P(B(x_E,2r))} \leq \frac{C_{2Q}^{-1} Q(B(x_E,r))}{P(E)} \leq \frac{C_{2Q}^{-1} Q(E)}{P(E)},
\end{equation*} and so we have $C_{2Q} \varphi(4r) \leq \Lambda(\xi,r) \leq \Lambda(r)$, so $\bar \rho_*(P,Q) \leq \gamma_*(P,Q)$ follows after taking logarithms, dividing by $ \log(1/r)$ and letting $r \to 0$ by Lemma \ref{gammaStarIsMinkowki}.  

\par %Let $\xi$ be an $r$-grid of $\cX_Q$, and let $x_E$ be as above for $E \in \xi$. By a simple volume argument there is a constant $k = k(d)$ depending only on the Minkowski dimension $d$ of $\cX_Q$ such that no point $x \in \cX_Q$ falls into more than $k(d)$ balls in the  $2r$-cover $\{x_E: E \in \xi\}$, that is: $$ \forall x \in \cX_Q, \quad | \{E \in \xi: x \in B(x_E,2r)\}| \leq k(d).$$
Let now $\xi$ be an $r$-grid of $\cX_Q$, and let $x_E$  be chosen so that $B(x_E,r) \subset E \subset B(x_E,2r)$. Then using that $P,Q$ are doubling and that $x \in B(x_E,r)$ implies $B(x,r) \subset B(x_E,2r)$ gives

%UNCOMMENT FOR ONE-COLUMN FORMAT:
\begin{align*}
    \varphi(r) &  = \int \frac{1}{P(B(x,r))} \, dQ(x)
       \geq \sum_{E \in \xi} \int_{B(x_E,r)} \frac{1}{P(B(x,r))} \, dQ(x) 
      \geq \sum_{E \in \xi} \int_{B(x_E,r)} \frac{1}{P(B(x_E,2r)} \, dQ(x) 
      \\& = \sum_{x_E \in \xi} \frac{Q(B(x_E,r))}{P(B(x_E,2r))}
    \geq \sum_{E \in \xi} \frac{C_{2Q}}{C_{2P}}  \frac{Q(E)}{P(E)}
     = \frac{C_{2Q}}{C_{2P}} \Lambda(\xi,r),
\end{align*}

%UNCOMMENT FOR TWO-COLUMN FORMAT:
%\begin{align*}
%    \varphi(r) &  = \int \frac{1}{P(B(x,r))} \, dQ(x)
%      \\& \geq \sum_{E \in \xi} \int_{B(x_E,r)} \frac{1}{P(B(x,r))} \, dQ(x) 
%    \\&   \geq \sum_{E \in \xi} \int_{B(x_E,r))} \frac{1}{P(B(x_E,2r)} \, dQ(x) 
%      \\& = \sum_{x_E \in \xi} \frac{Q(B(x_E,r))}{P(B(x_E,2r))}
%    \\&  \geq \sum_{E \in \xi} \frac{C_{2Q}}{C_{2P}}  \frac{Q(E)}{P(E)}
%     = \frac{C_{2Q}}{C_{2P}} \Lambda(\xi,r),
%\end{align*} 
and since $\xi$ is arbitrary we have $\varphi(r) \geq C \Lambda(r)$, which gives $\gamma_*(P,Q) \leq \bar \rho_*(P,Q)$.
\end{proof}

\begin{proof}\textit{(Proposition \ref{noSuperTransfer}.)}
 For small $r$, choose an $r$ packing of $S$ of size at least $r^{-d}$, which can be done by \eqref{QmetricEntropy}; denote the centres of the chosen balls by $\mathcal{Z}$, and let $\xi$ be an $r$ grid that extends the packing, in the sense that for each $z \in \mathcal{Z}$ there is a $E_z \in \xi$ such that $B(x,r) \subset E_z$. By solving a constrained optimization using Lagrange multipliers \cite{BERTSEKAS1982}, we see that 
$$\Lambda(\xi,r) = \sum_{E \in \xi} \frac{Q(E)}{P(E)} \geq \left( \sum_{B \in \xi} Q(E)^{1/2} \right)^2,$$ and using the assumption on $S$ yields
\begin{align*}
    \left( \sum_{E \in \xi} Q(E)^{1/2} \right)^2 \geq \left( \sum_{z \in \mathcal{Z}} Q(B(z,r)^{1/2} \right)^2  \geq \left( |\mathcal{Z}| (C r^d)^{1/2} \right)^2
    \geq r^{-2d} \, C r^d = C r^{-d},
\end{align*} which implies $\Lambda(r) \geq C r^{-d}$, so taking logarithms and a limit as $r \to 0$ gives  \mbox{$\gamma_*(P,Q) \geq d$}. 
\end{proof}

%\section{}
\section{Details for Figure \ref{fig:E1_experiment}}\label{appendix_figure2_details}

Here we provide details of the constructions of the distributions used for Figure \ref{fig:E1_experiment}. Recall the setup of Example \ref{E1} from Section \ref{secOnDim} below, with $d = 5$. We choose five marginal distributions $P_X^k$, $k = 0, 1,2,3,4$, having density $p^k(x) \propto d(x,A_k)^\nu$ for some $\nu \geq 0$, where $d(x,A) = \inf_{y \in A} \norm{x - y}$ is the standard distance function and 
\begin{align*}
    A_k = \{x \in [0,1]^5 \, : \, (x_{k+1}, \dots, x_5) = 0\}
\end{align*} are subsets of $[0,1]^5$ of dimension $k$. We set $Q_X$ to be uniform on $[0,1]^5$, and it follows from the discussion in Section \ref{secOnDim} that we have $\rho(P_X^k, Q_X) = \nu$ (so $\rho + d = \nu + d)$, while $\gamma(P_X^k, Q_X) = \nu + k$.  
\par The top plot in Figure \ref{fig:E1_experiment} shows the results of applying a regular dyadic tree procedure, with depth chosen by 2-fold cross-validation, for transferring from $P^i$ to $Q$, where $P^i$ and $Q$ are determined from $P_X^i$ and $Q_X$ and the regression function $\eta(x) = \tfrac{1}{2}(1 + \sin(\pi\|x\|_1))$. As can be seen, transfer performance from $P$ to $Q$ deteriorates as $\gamma(P_X,Q_X)$ increases, as our results predict. Since each of the source measures have the same transfer exponent $\nu$, this naturally illustrates the drawback of using that quantity as an indicator of potential for transfer. The bottom plot of Figure \ref{fig:E1_experiment} shows the results of the same procedure applied to the distributions $\widetilde P^k_X$ for $k = 0,1,2,3,4$, where the density is now given by $\widetilde p^k(x) \propto d(x,A_k)^{\nu - k}$, so that now $\rho(\widetilde P_X^i, Q_X) = \nu - k$, while $\gamma(P_X^k, Q_X) = \max(\nu, d)$ is constant for each distribution, and again $\nu$ is set to $5$; the regression function is unchanged. As we can clearly see in the figure, in this case there is little to tell between the distributions in terms of how effectively we can transfer, which is in line with what we expect, since each distribution has the same aggregate transfer exponent with respect to the target measure. On the other hand, the transfer exponents do change, but in this instance the higher transfer exponents do not translate to inferior transfer.

\section{Calculations for Example \ref{E1}.}\label{example1CalcAppendix}
Recall that we take $Q_X$ to be uniformly distributed on $[0,1]^d$, and for $x \in [0,1]^d$  we let $P_X$ have density $p(x) \propto \| x \|^\nu$ for some $\nu > 0$. Note that in this case, both $P_X$ and $Q_X$ are doubling measures, so we may use the integral form \eqref{transferIntegral} to calculate $\gamma$. For $x \in B(0,2r)$, we have (up to a constant) $P_X(B(x,r)) \geq P_X(B(0,r)) = C r^{d + \nu}$ for some constant $C$. For $x \notin B(0,2r)$ and $y \in B(x,r)$, we have $\|y\| \geq (1 - \tfrac{r}{\|x\|}) \|x\| \geq \tfrac{1}{2}\|x\| $, and therefore $P_X(B(x,r)) \geq 2^{-\nu} \| x\|^\nu V_d(r)$, where $V_d(r) = C(d) r^d$ is the volume of an $r$-ball in dimension $d$. This implies that we have 
\begin{align}
    \int_{ [0,1]^d \cap B(0,2r) } P_X(B(x,r))^{-1} \, Q_X(dx) \leq \int_{B(0,2r)} C_1 r^{-(\nu + d)} \, dx 
     = C_1 C(d) 2^d r^{-\nu}, \label{aroundZero}
\end{align} while 

%UNCOMMENT FOR ONE-COLUMN FORMAT
\begin{align}
    \int_{[0,1]^d \setminus B(0,2r)} P_X(B(x,r))^{-1} \, Q_X(dx)  \leq 2^\nu C(d)^{-1} r^{-d} \int_{[0,1]^d \setminus B(0,2r)} \|x\|^{-\nu} \, dx. \label{awayFromZero1} 
\end{align}

%UNCOMMENT FOR TWO-COLUMN FORMAT
%\begin{align}
%   & \int_{[0,1]^d \setminus B(0,2r)} P_X(B(x,r))^{-1} \, Q_X(dx)  \nonumber \\& \quad \leq 2^\nu C(d)^{-1} r^{-d} \int_{[0,1]^d \setminus B(0,2r)} \|x\|^{-\nu} \, dx. \label{awayFromZero1} 
%\end{align} 

Now, we have 
\begin{align}
    \int_{[0,1]^d \setminus B(0,2r)} \|x\|^{-\nu} \, dx & \leq 2^{-d} \int_{B(0, \sqrt{d}) \setminus B(0,2r)} \|x\|^{-\nu} \, dx, \label{awayFromZero2}
\end{align} and switching to spherical coordinates and integrating over the angular variables gives

%UNCOMMENT FOR TWO-COLUMN FORMAT
\begin{align}
   \int_{B(0, \sqrt{d}) \setminus B(0,2r)} \|x\|^{-\nu} \, dx  = \int_{[0,\pi/2]^{d-1}} \int_{2r}^{\sqrt{d}} s^{-\nu} \, C(\varphi) s^{d-1} \, d\varphi \, ds  = C \int_{2r}^{\sqrt{d}} s^{-(\nu - d - 1)} \, ds.  \label{finalQuantity}
\end{align}

%UNCOMMENT FOR TWO-COLUMN FORMAT
%\begin{align}
%   & \int_{B(0, \sqrt{d}) \setminus B(0,2r)} \|x\|^{-\nu} \, dx \nonumber \\& \quad  = \int_{[0,\pi/2]^{d-1}} \int_{2r}^{\sqrt{d}} s^{-\nu} \, C(\varphi) s^{d-1} \, d\varphi \, ds \nonumber
%    \\& \quad = C \int_{2r}^{\sqrt{d}} s^{-(\nu - d - 1)} \, ds.  \label{finalQuantity}
%\end{align}

We distinguish three cases: firstly, if $\nu < d$, then $-(\nu - d - 1) > -1$ and we have \begin{align}
     \int_{2r}^{\sqrt{d}} s^{-(\nu - d - 1)} \, ds = O(1) \label{case1}
\end{align} as $r \to 0$, and therefore in this case by \eqref{awayFromZero1}-\eqref{case1} we have 
\begin{align*}
    \int_{[0,1]^d} P_X(B(x,r))^{-1} \, Q_X(dx) \leq \int_{[0,1]^d \setminus B(0,2r)} P_X(B(x,r))^{-1} \, Q_X(dx) \leq C r^{-d}, 
\end{align*} and so $d = \max(\nu,d)$ is an aggregate exponent in this case. If $\nu > d$, then we have 
\begin{align}
     \int_{2r}^{\sqrt{d}} s^{-(\nu - d - 1)} \, ds = O(r^{-(\nu -d)}) \label{case2}
\end{align} as $r \to 0$, and so by \eqref{aroundZero}, \eqref{awayFromZero1}, \eqref{awayFromZero2} and \eqref{case2} we have 
\begin{align*}
    \int_{[0,1]^d} P_X(B(x,r))^{-1} \, Q_X(dx) \leq C r^{-\nu},
\end{align*} and so $\nu = \max(\nu, d)$ is an aggregate exponent. If $\nu = d$, then we find that 
\begin{align}
     \int_{2r}^{\sqrt{d}} s^{-(\nu - d - 1)} \, ds = O(\log(1/r) %\label{case3}
\end{align} as $r \to 0$; in this case for any $\epsilon > 0$ we have that $d + \epsilon$ is an aggregate exponent, while $\gamma_*(P_X, Q_X) = d = \max(\nu,d)$. 
\section{Proofs of Auxiliary Lemmas.}\label{appendixProofs}

Here we fill in the proofs of the Lemmas that were used to prove the main results. For completeness, we also state the version of the multiplicative version of the Chernoff bound that we employ to prove Lemma \ref{compPhiRare}. 

\subsection{Multiplicative Chernoff Bound.}
Here we recall the Angluin-Valiant form of the relative Chernoff bound, which can be found, for example, in  \cite{histRuleConcentration}. 

\begin{lemma}[Angluin-Valiant]\label{relChernoff}
Let $X_1, \dots, X_n$ be independent Bernoulli random variables with mean $EX_i = p_i$. Let $X = \sum X_i$, and $\mu = EX$. Then for $0 < \beta < 1$ we have \begin{equation*}%\label{AGbound2}
    \pr( X \leq (1 - \beta)\mu) \leq \exp\left\{-\tfrac{1}{2} \beta^2 \mu\right\}.
\end{equation*}  
\end{lemma}

\subsection{Upper Bounds.}

%The following result will be used to prove %lemma \ref{thresholdRiskContrib}. 
\begin{proof}{\textit{(Lemma \ref{firstBoundBias}.)}}
Recall that $\Sone = \{|\tilde A_r(x) \cap \bX| > 0\}$, so we may write 
\begin{align*}
\abs{\tilde{\eta}_r(x) - \eta(x)}\ind\{\Sone\} & = \abs{ \frac{1}{|\tilde{A}(x) \cap \bX|} \sum_{i:X_i \in \tilde{A}(x)} (\eta(X_i) - \eta(x))}\cdot\ind\{\Sone\}
 \\& \leq \frac{1}{|\tilde{A}(x) \cap \bX|} \sum_{i:X_i \in \tilde{A}(x)} \abs{\eta(X_i) - \eta(x)}\cdot\ind\{\Sone\} \leq C_\alpha r^\alpha,
\end{align*} where the final inequality is an application of the H\"older continuity condition \eqref{holderSmooth} on $\eta$. 
\end{proof}

\begin{proof}{\textit{(Lemma \ref{firstBoundVar}.)}}
Using the definitions of $\hat \eta$  \eqref{etaHat} and $\tilde \eta$ \eqref{etaTilde}, we see that we have 
\begin{align*}
    \E_{\bY \mid \bX}[\abs{\hat{\eta}_r(x) - \tilde{\eta}_r(x)}^2  \mid \bX] \ \ind\{\phiTwo(X)\} & = \E \left[\frac{1}{|\tilde A_r(x)\cap \bX|^2} \left(  \sum_{i:X_i \in \tilde{A}(x)} (Y_i - \eta(X_i))\right)^2 \mid \bX \right] \ind\{\phiTwo(X)\} \\ &  = \frac{1}{|\tilde A_r(x) \cap \bX|^2} \E\left[ \sum_{i:X_i \in \tilde{A}(x)} (Y_i - \eta(X_i))^2\mid \bX \right] \ \ind\{\phiTwo(X)\}
   \\& =  \frac{1}{|\tilde A_r(x) \cap \bX|^2} \sum_{i:X_i \in \tilde{A}(x)}\, \eta(X_i)(1 - \eta(X_i)) \ \ind\{\phiTwo(X)\}
    \\ & \leq  \frac{1}{4 |\tilde A_r(x)\cap \bX|} = \tfrac{1}{4} \hat V_r(x),
\end{align*} where the second equality is the Pythagorean theorem, and the inequality is simply $\eta(x)(1 - \eta(x)) \leq 1/4$, since $\eta(x) \in [0,1]$. Since $x \in \cX$ was arbitrary, this completes the proof.

\end{proof}

\begin{figure}
\begin{center}
\resizebox{6cm}{6cm}{%
\begin{tikzpicture}
\fill[red!15!white] (0,1) rectangle (3,4);
\fill[red!50!white] (1,2) rectangle (2,3);
\draw[step=1cm,gray, thin] (-2,-2) grid (6,6);
\draw[blue] (-1,-2) .. controls (-1,2) and (5,3) .. (6,5);
\filldraw[black] (1.9,2.1) circle (2pt) node[anchor=west] {x};
\draw[black] (0.9,1.1) rectangle (2.9,3.1);
\draw[black] (-0.1,0.1) rectangle (3.9, 4.1);
\end{tikzpicture}
}
\end{center}
\caption{When the support of the target distribution lies on a lower-dimensional manifold in $[0,1]^D$, the dyadic partition of the latter can lead to an irregular partition of the former; this problem is mitigated by taking envelopes. Above we see a dyadic grid, along with $\cX_Q$ in blue. One cell from the partition is in red, and its envelope in light red. The black boxes centred on $x$ are $B(x,r)$ and $B(x,2r)$, demonstrate that the envelope $\tilde A (x)$ satisfies the conditions for being a grid element (see \ref{gridDef}).}
\label{fig:gridFig}
\end{figure}
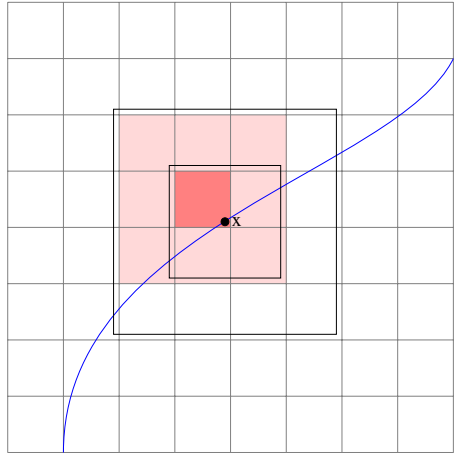

\begin{proof}{\textit{(Lemma \ref{gammaSumLemma}.)}}
Recall that the cells $A \in \Pi_r$ are hypercubes of side length $r$. Observe that by construction each of the $r$-envelopes of sets $A$ with $Q_X(A) > 0$ (that is, the cells $A \in \Pi_r(\cX_Q)$) has a point $x \in A$ with $B(x,r) \cap \cX_Q \subset \tilde{A}_r$, and of course $\tilde{A}_r \subset B(x,2r)$; see Figure \ref{fig:gridFig}. We aim to show that there is a constant $K = K(D)$ such that one can find $K$ grids $\xi_1, \dots, \xi_K$ of $\cX_Q$ so that for each $A \in \Pi_r(\cX_Q)$, the envelope $\tilde A_r$ belongs to one of the grids. In order for it to be possible for the envelopes of two such cells $A,A^\prime$ to belong to the same grid, one must be able to find $x,x^\prime \in \cX_Q$ such that $B(x,r) \cap B(x^\prime, r) = \emptyset$; a sufficient condition ensuring that this can be done is that $d(x,x^\prime) \geq 2r$ for all $x \in A, x^\prime \in A^\prime$, that is, the dyadic cells $A,A^\prime$ are `separated' by at least two other cells. For a given cell $A$, this fails to hold for only the $5^D$ cells that constitute the $2r$ envelope of $A$. Now, consider a graph with a vertex corresponding to each cell $A \in \Pi_r(\cX_Q)$, with an edge between the vertices for $A,A^\prime$ if $A^\prime \in \tilde A_{2r}$. Observe that a \emph{colouring} of this graph - that is, an assignment of colours to each vertex such that no two vertices of the same colour share an edge - corresponds to a group of cells that can be assigned to the same grid. A simple greedy argument shows that one can always colour a graph $G$ using $\max_{v \in G} \mathrm{deg}(v) + 1$ colours, and the above reasoning implies that the maximal degree of our graph is $5^D$, whence we can choose $K = 5^D + 1$. The rest of the argument is immediate: 
  \begin{align*}
     \sum_{A \in \Pi_r(\cX_Q)} \frac{Q_X( \tilde A_r)}{P_X(\tilde A_r)}  \leq 
     \sum_{i =1}^K \sum_{\tilde{A_r} \in \xi_i} \frac{Q_X(\tilde{A_r})}{P_X(\tilde{A_r})}
      \leq   K \, C_\gamma  r^{-\gamma}, 
\end{align*} where the final step uses \eqref{globalTransferExp}.
\end{proof}

\begin{proof}{\textit{Proposition \ref{lowerBound}.}}
Consider the transfer class $\cT(\alpha, \beta, \gamma, d)$ for some $\alpha, \beta, d \geq 0$ and $\gamma \geq d$. Then, by Propositions \ref{sharperBounds} and \ref{relDimEquivalence}, we see that $\cT$ contains the restricted class $\widetilde \cT$, where $(P,Q) \in \tilde \cT$ if assumptions i)-iv) of Definition \ref{classesDef} hold, and $P_X$ has transfer exponent $\gamma - d$ with respect to $Q_X$, and $P_X, Q_X$ are doubling measures. It is easy to check that all measures used in the construction of \cite{KM} are doubling, and so it follows by their Theorem 1 that we have 
$$ \inf_{\hat f} \sup_{(P,Q) \in \widetilde \cT} \E \, \cE(\hat f) \geq C^\prime \min\left\{ n_P^{-\tfrac{\alpha(\beta +1)}{2 \alpha + \alpha \beta + \gamma}}, n_Q^{-\tfrac{\alpha(\beta+1)}{2\alpha + \alpha \beta + d}} \right\}.$$ Since $\widetilde \cT \subset \cT$, the result follows. 
\end{proof}

\subsection{Adaptive Rates.}

\begin{proof}{\textit{(Lemma \ref{lepskiControl}.)}}
For fixed $x,r$, let $Y_{x,r}$ denote the $Y$-values corresponding to those $X$'s that fall in $\tilde{A}_r(x)$. Supposing that the $X$'s are held fixed, let $\phi(Y_{x,r}) = \abs{\hat{\eta}_{r}(x) - \tilde{\eta}(x)}$, and note that in the proof of Lemma \ref{firstBoundVar} it is shown that $\E[\phi(Y_{x,r})] \leq \tfrac{1}{2 \sqrt{|\tilde A_r(x) \cap \bX |}} = \tfrac{1}{2} \hat V_r(x)^{1/2}$, where the expectation is over $Y^n \mid X^n$. Since changing one $Y$ value can change $\phi(Y_{x,r})$ by at most $1/|\tilde A_r(x) \cap \bX |$, for a fixed $t_{x,r} > 0$ an application of McDiarmid's bounded differences inequality gives  
\begin{equation}\label{mcD}
\pr( \phi(Y_{x,r}) > \E[\phi(Y_{x,r})] + t_{x,r})  \leq \exp\{-2t^2_{x,r} |\tilde A_r(x) \cap \bX |\},
\end{equation} for any $x$ such that $| \tilde A_r(x) \cap \bX | > 0$, where the probability on the left hand side is again over $Y^n$ with the $X^n$ held fixed. Note that the above holds trivially on empty cells, since $\phi(Y_{x,r}) \equiv 0$ there. For a fixed level $r$, there are at most $n_P + n_Q$ empty cells, and by construction (see \eqref{levels_def}) there are no more than $\log(n_P + n_Q)$ levels, and so setting $t_{x,r}$ appropriately, \eqref{mcD} reads that with probability at least $1 - \delta/(n_P + n_Q)^2$, we have  
\begin{align*} \phi(Y_{x,r}) &\leq \frac{1}{2 \sqrt{|\tilde A_r(X) \cap \bX |}} + \sqrt{\frac{\log ((n_P + n_Q)/\delta)}{ |\tilde A_r(X) \cap \bX |}}
= \tfrac{1}{2} \sqrt{\hat V_r(x)} \left( 1 + 2\, \sqrt{\log (n_P + n_Q)/\delta) }\right). 
\end{align*} Taking a union bound over each non-empty cell at each depth yields \eqref{PhiEvent} with probability at least $1 - \delta$ over the conditional distribution $Y^n \mid X^n$, and we conclude by taking an expectation over $X^n$. 
\end{proof}

\begin{proof}{\textit{(Lemma \ref{lepskiKeyLemma}.)}}
Fix $x \in \cX_Q$. Recall that we let $\hat{\sigma}_r(x) = C_{n,\delta} \hat V_r(x)^{1/2}$, where $\hat V_r(x) = |\tilde A_r(x) \cap \bX|^{-1}$; let us define 
\begin{equation*}
    r^*(x) \defeq \max\left\{ r \in \hat{R}: \hat \sigma_r(x) \geq C_\alpha r^\alpha \right\}.
\end{equation*} Clearly there is an $N$ for which this set is non-empty provided $n_P + n_Q > N$, and so $r^* \neq - \infty$. Set $\hat{r}(x)$ to be the depth at which Algorithm \ref{lepskiAlg} stops, that is 
\begin{equation*}
    \hat{r}(x) \defeq \min \left\{ r \in \hat{R}: \hat{\eta}_r^+ \leq \hat{\eta}_r^-, \textrm{ or } \hat{\eta}_r^+ \leq 1/2, \textrm{ or } \hat{\eta}^-_r \geq 1/2, \textrm{ or } r = 1 \right\},    
\end{equation*} where $\hat{\eta}^+_r$ and $\hat{\eta}^-_r$ denote the values of $\hat{\eta}^+,\hat{\eta}^-$ at the iteration at which the depth is set to $r$. Though $r^*$ is of course unknown, we will argue that the error at level $\hat{r}$ cannot be too far from the error at $r^*$ (which, in turn, can be linked to the error of the oracle, since the level that optimally trades between bias and variance is close to $r^*$ with high probability; see below). Again, by Lemma \ref{regBoundEstApprox} we have 

\begin{align}\label{PhiAndXi}
    \abs{\hat{\eta}_r(x) - \eta(x) } \leq \frac{1}{2\sqrt{ |\tilde A_r(x) \cap \bX|}}\left( 1 + 2\, \sqrt{\log (n_P + n_Q)/\delta) }\right) + C_\alpha r^\alpha
    = \hat \sigma_r (x) + C_\alpha r^\alpha 
\end{align} 

for each $r \in \hat{R}$. We proceed by arguing by cases. Suppose first that $\hat{r} \leq r^*$. In this case, since $\hat{\sigma}_r(x)$ is non-increasing in $r$, while $C_\alpha r^\alpha$ is increasing, it must be that $\hat{\sigma}_{\hat{r}} \geq C_\alpha \hat{r}^\alpha$, and it follows from \eqref{PhiAndXi} that $\eta(x) \in [\hat \eta_r(x) - 2 \hat \sigma_r(x),\hat \eta_r(x) + 2 \hat \sigma_r(x)]$; in fact this holds for any $r < r^*$, and so we must have $\eta(x) \in [\hat{\eta}_{\hat{r}}^-,\hat{\eta}_{\hat{r}}^+]$, because by Algorithm \ref{lepskiAlg} we have \begin{equation*}[\hat{\eta}_{\hat{r}}^-,\hat{\eta}_{\hat{r}}^+] = \bigcap_{r \in \hat R, \, r \leq \hat r}  [\hat \eta_r(x) - 2 \hat \sigma_r(x),\hat \eta_r(x) + 2 \hat \sigma_r(x)].
\end{equation*} It must be that $\hat{\eta}_{\hat{r}}^- \leq \hat{\eta}_{\hat{r}}^+$, and so the algorithm must have stopped due to either $\hat{\eta}_{\hat{r}}^+ \leq 1/2$ or $\hat{\eta}_{\hat{r}}^- \geq 1/2$. In either case $\hat{\eta}_{\hat{r}}(x)$ is on the same side of $1/2$ as $\eta(x)$, so $\hat{f}(x) = f^*(x)$ and \eqref{lepskiRateBound} holds since the left hand side is equal to zero. \par Suppose on the other hand that $\hat{r} > r^*$. Let 
\begin{equation}\label{sigmaBar}
    \Bar{\sigma}\defeq \min_{r \in \hat{R}} \left\{ \max \left(\hat \sigma_r(x), C_\alpha r^\alpha \right)\right\}.
\end{equation} Again, since $r \mapsto \hat{\sigma}_r$ is non-increasing while $r \mapsto C_\alpha r^\alpha$ is increasing, we see that the minimum in the above expression must be attained at either $r^*$ or $2r^*$. If the minimum occurs at $r^*$, we have $\Bar{\sigma} = \hat{\sigma}_{r^*} \geq \hat{\sigma}_{2r^*}$, and in the latter case we have $\hat{\sigma}_{2r^*} \leq C_\alpha (2r^*)^\alpha = \Bar{\sigma}$; it follows that we have $[\hat{\eta}_{2r^*}^-(x),\hat{\eta}_{2r^*}^+(x)] \subset [\hat \eta_{2r^*} - 2\bar \sigma, \hat \eta_{2r^*} + 2\bar \sigma]$. Now, $\hat{r} > r^*$ means $\hat{r} \geq 2r^*$, and so by construction we have 

$$[\hat{\eta}_{\hat{r}}^-(x),\hat{\eta}_{\hat{r}}^+(x)] \subset [\hat{\eta}_{2r^*}^-(x),\hat{\eta}_{2r^*}^+(x)],$$ and   and $\eta(x) \in [\hat \eta_{2r^*} - 2\bar \sigma, \hat \eta_{2r^*} + 2\bar \sigma]$ holds by \eqref{PhiAndXi}, and so it follows that 
\begin{equation}\label{sigBarErrorBound}
    \abs{\hat{\eta}_{\hat{r}}(x) - \eta(x)} \leq 2 \Bar{\sigma}.
\end{equation} Now, since obviously $\epsilon < r_n$, there exists an $r \in \hat{R}$ such that $r \leq r_n < 2r$, and so \eqref{sigmaBar} and \eqref{sigBarErrorBound} imply $$
\abs{\hat{\eta}_{\hat{r}}(x) - \eta(x)} \leq 4 \left( \hat \sigma_{r_n}(x) + C_\alpha r_n^\alpha \right),$$ and the final claim follows immediately because $$\{ \hat{f}_{\hat{r}}(X) \neq f^*(X)\} \subset \left\{\abs{\eta(X) - \tfrac{1}{2}} \leq \abs{\hat{\eta}_{\hat{r}}(X) - \eta(X)} \right\},$$ and $x \in \cX_Q$ was arbitrary.  
\end{proof}

\begin{proof}{\textit{(Theorem \ref{adaptiveTheorem}.)}}
 We proceed in the same way as for Theorem \ref{oracleRate}. Let $\hat f$ denote the classifier given by Algorithm 1. Recall that we have set $\hat V_r(x) \defeq |\tilde A_r(x) \cap \bX|^{-1}$, and $V_r(x) = \left( n_P P_X(\tilde A_r(x)) + n_Q Q_X(\tilde A_r(x)) \right)^{-1}$, and we let $\Phi \defeq \Phi_{r_n}(X) = \{\hat V_{r_n}(X) < 2 V_{r_n}(X)\}.$ Further, let $$\Omega_{\lambda} \defeq \{V_{r_n}(X) \leq \lambda\}.$$ Applying Lemma \ref{lemmaAdaptiveDecomp} with expectations taken on the event $\Phi$, we see that (recalling that we write $M(x) = \abs{\eta(x) - \tfrac{1}{2}}$) 
 
\begin{align*}
   \E[\cE(\hat f)] &\leq  2 \E \left[M(X) \cdot \ind\left\{M(X) \leq 2 \frac{C_{n,\delta}}{\sqrt{| \tilde A_{r_n}(X) \cap \bX|}} \right\} \cdot  \ind\{\Phi\}\right] \\& \quad + 2\E \left[M(X) \cdot \ind\left\{M(X) \leq 2 C_\alpha r_n^\alpha \right\} \cdot  \ind\{\Phi\}\right]   + \E \left[ \ind\{\Phi^\comp\} \ind\{\Omega_\lambda\} + \ind\{\Omega_\lambda^\comp\} \right]. 
\end{align*} 

By Lemmas \ref{approxErrorProp},\ref{estErrorLemma2} and \ref{compPhiRare}, we can bound the four terms in the above display, yielding

\begin{align*} \E[\cE(\hat f)]  \leq C \left( r_n^{\alpha(\beta + 1)} + \left[C_{n,\delta}^2 \min\left( \frac{r_n^{-\gamma}}{n_P}, \frac{r_n^{-d}}{n_Q}\right)\right]^{\frac{\beta + 1}{\beta + 2}} + \exp\{- \lambda^{-1}/8\} + \frac{1}{\lambda} \min \left( \frac{r_n^{-\gamma}}{n_P}, \frac{r_n^{-d}}{n_Q}\right) \right) + \delta .
\end{align*}

Setting $\delta \defeq \tfrac{1}{n_P + n_Q}$ gives $C_{n,\delta} \leq \sqrt{d \log(n_P + n_Q)}$, and now again we can set $\lambda = r_n^\alpha$ and ignore the third term, which gives a final bound of

\begin{align*} \E\left[ \cE(\hat f)\right] \leq   C \min\left\{ \left(\frac{\log n_P }{n_P}\right)^{\tfrac{\alpha(\beta +1)}{2 \alpha + \alpha \beta + \gamma}}, \left(\frac{\log  n_Q }{n_Q}\right)^{\tfrac{\alpha(\beta+1)}{2\alpha + \alpha \beta + d}} \right\},
\end{align*}
for some $C > 0$ free of $n_P,n_Q$, and this completes the proof.  \par 
%For the case $P,Q \in \cT^\prime$, note that re-arranging Lemma %\ref{lepskiControl} gives a version of Lemma \ref{Qconcentration} with $a_n %\defeq \left(\frac{n_Q}{\log(n_Q)}\right)^{\frac{2\alpha}{2\alpha + d}}$, after %which we can mirror the proof of the $\cT^\prime$ case for Theorem %\ref{oracleRate}, giving a final rate of 
%\begin{align*} & \sup_{P,Q \in \cT^\prime} \E\left[\cE(\hat f)\right] \\ & %\quad \leq C \min\left\{ \left(\frac{\log n_P }{n_P}\right)^{\tfrac{\alpha(\beta %+1)}{2 \alpha + \alpha \beta + \gamma}}, \left(\frac{\log  n_Q %}{n_Q}\right)^{\tfrac{\alpha(\beta+1)}{2\alpha +  d}} \right\}. 
%\end{align*} 
\end{proof}

As noted in Section \ref{secExperiments}, the procedure which we have implemented in the experiments, in which the nodes of the dyadic trees are split successively into two children (rather than $2^d$) comes with the same guarantees. We formalize this here. Let $\Pi_\epsilon$ denote an arbitrary cyclical dyadic partition of $[0,1]^D$ down to $\epsilon$-scale; that is, beginning with the node $[0,1]^D$, each hyper-rectangular cell is split by choosing one of the dimensions along which the node is largest and bisecting along that dimension, and this procedure is repeated until all the cells have been collapsed until the side lengths $r$ satisfy $r < \epsilon$. The regression estimates and classifier are defined as in \eqref{treeClassifier} (with envelopes of cells taken by enlarging according to the width of the smallest side). Set $\epsilon = (n_P + n_Q)^{-1/2}$, and let $\hat f ^\prime$ be the classifier that results from applying Algorithm \ref{lepskiAlg} to this tree. 

\begin{corollary}\label{cyclicalTreeRates}
    Let $\hat f ^\prime$ be as above. Then there is a $C^\prime > 0,$ free of $n_P, n_Q$, such that 
    \begin{align*} \E\left[ \cE(\hat f^\prime)\right] \leq   C^\prime \min\left\{ \left(\frac{\log n_P }{n_P}\right)^{\tfrac{\alpha(\beta +1)}{2 \alpha + \alpha \beta + \gamma}}, \left(\frac{\log  n_Q }{n_Q}\right)^{\tfrac{\alpha(\beta+1)}{2\alpha + \alpha \beta + d}} \right\},
\end{align*}
\end{corollary}

\begin{proof}
    The argument is identical to that used for the regular dyadic tree. Of course, the union bound used to prove Lemma \ref{lepskiControl} will need to be taken over a larger set of cells, but this has no effect on the result other than changing the value of the leading constant in the rate. 
\end{proof}

\subsection{Localizing to the Decision Boundary}\label{appendix_Localization}

\begin{proof}[\textit{Lemma \ref{riskFromGplus}}]
First, note that we have (recall that $\{\hat f_\epsilon \neq f^*\} \subset \{ |\eta(x) - \tfrac{1}{2}| \leq |\hat \eta_{r^+}(x) - \eta(x)|\}$)

%UNCOMMENT FOR ONE-COLUMNS: 
\begin{align*}
    \E \, \cE(\hat f_\epsilon; \cG_\epsilon^+) & = 2 \, \E \left[ M(X) \ind\left\{\hat f_\epsilon \neq f^*\right\} \ind\{X \in \cG_\epsilon^+\} \right] 
     \leq 2 \, \E \left[ M(X) \cdot \ind\left\{M(X) \leq \abs{\hat \eta_{r_+}(X) - \eta(X)}\right\} \cdot \ind\{X \in \cG_\epsilon^+\}\right]
    \\& \leq 2 \, \E \left[ M(X) \cdot\ind\left\{M(X) \leq 2\abs{\hat \eta_{r_+}(X) - \tilde \eta_{r_+}(X)}\right\}\cdot\ind\{X \in \cG_\epsilon^+\}\right] \\& \quad + 2 \, \E \left[ M(X) \cdot\ind\left\{M(X) \leq 2\abs{\tilde \eta_{r_+}(X) - \eta(X)}\right\}\cdot\ind\{X \in \cG_\epsilon^+\}\right]. 
\end{align*} 

%UNCOMMENT FOR TWO-COLUMNS: 
%\begin{align*}
%    & \E \cE(\hat f_\epsilon; \cG_\epsilon^+)  \\& = 2 \, \E \left[ M(X) \ind\{\hat f_\epsilon \neq f^*\} \ind\{X \in \cG_\epsilon^+\} \right] 
%    \\& \leq 2 \, \E M(X) \ind\{M(X) \leq \abs{\hat \eta_{r_+}(X) - \eta(X)}\} \ind\{X \in \cG_\epsilon^+\}
%    \\& \leq 2 \, \E M(X) \ind\{M(X) \leq 2\abs{\hat \eta_{r_+}(X) - \tilde \eta_{r_+}(X)}\}\ind\{X \in \cG_\epsilon^+\} \\& \quad + 2 \, \E M(X) \ind\{M(X) \leq 2\abs{\tilde \eta_{r_+}(X) - \eta(X)}\}\ind\{X \in \cG_\epsilon^+\}. 
%\end{align*} 

Now, let $\phiTwo(X)$ be the event introduced in Notation \ref{phiDef}. On $\phiTwo(X)$, we have (see Lemma \ref{firstBoundBias})  $\abs{\tilde \eta_{r_+}(X) - \eta(X)} \leq C_\alpha r_+^\alpha$, and so $r_+ < (\epsilon/2 C_\alpha)^{1/\alpha}$ implies that 
\begin{align*}
    \ind\left\{M(X) \leq 2\abs{\tilde \eta_{r_+}(X) - \eta(X)}\right\}\ind\{X \in \cG_\epsilon^+\}\ind\{\phiTwo(X)\} = 0.
\end{align*} We then have

\begin{align*}
     \E \ \cE(\hat f_\epsilon; \cG_\epsilon^+)  \leq 2 \, \E  \left[  M(X) \cdot \ind\left\{M(X) \leq 2\abs{\hat \eta_{r_+}(X) - \tilde \eta_{r_+}(X)}\right\}\ind\{X \in \cG_\epsilon^+\} \right] + \pr(\phiTwo(X)^\comp). 
\end{align*} 

By Lemma \ref{compPhiRare}, we have
\begin{align*}
    \pr(\phiTwo(X)^\comp)  \leq C \, \min\left(  \frac{1}{n_P} r_+^{-\gamma^*}, \frac{1}{n_Q} r_+^{-d} \right) 
     \leq \left( \frac{1}{\epsilon} \right) \, C \, \min\left(  \frac{1}{n_P} r_+^{-\gamma^*}, \frac{1}{n_Q} r_+^{-d} \right) 
\end{align*} 
since $\epsilon < 1$. The remaining term is bounded using the same steps as in the proof of Lemma \ref{estErrorLemma2} with $t = \epsilon$, where the first term vanishes because we integrate over $\cG_\epsilon^+$.  
\end{proof}

\begin{proof}{[Theorem \ref{adaptiveAttainsEpsilonBound}]}
Fix $\epsilon > 0$. By Theorem \ref{adaptiveTheorem}, there is a universal $C = C(\cT)$ such that \begin{align*}  \E \ \cE( \hat f; \cG_\epsilon^-) \leq  C \min\left( \left(\frac{\log(n_P)}{n_P}\right)^{\tfrac{\alpha(\beta +1)}{2 \alpha + \alpha \beta + \gamma(\epsilon)}}, \left(\frac{\log(n_Q)}{n_Q}\right)^{\tfrac{\alpha(\beta+1)}{2\alpha + \alpha \beta + d}} \right).
\end{align*} Now, set $r_n^+ = (\epsilon/4C_\alpha)^{1/\alpha}$. Assume for now that  $\epsilon \geq 4 C_\alpha (n_P+n_Q)^{-\alpha}$. Then we have $\tfrac{1}{n_P + n_Q} \leq r_n^+$, and so an application of Lemma \ref{lepskiKeyLemma} at level $r_n^+$ (with $\delta = (n_P + n_Q)^{-1}$ and $C_n = C_{n,\delta}$) yields that, for $x \in \cG_\epsilon^+$, we have with probability at least $1 - (n_P + n_Q)^{-1}$ that 

\begin{align*}  \ind\{\hat f(x) \neq f^*(x)\} \leq \ind\left\{M(x) \leq 2\left( C_{n}| \tilde A_{r_n^+}(x) \cap \bX |^{-1/2} + C_\alpha (r_n^+)^\alpha \right) \right\},
\end{align*}
and then splitting the indicator on the right-hand side gives 
\begin{align*}
    \ind\{\hat f(x) \neq f^*(x)\} \leq \ind\left\{ \abs{\eta(x) - \tfrac{1}{2}} \leq 4 C_{n} | \tilde A_{r_n^+}(x) \cap \bX |^{-1/2}\right\}  + \ind\left\{\abs{\eta(x) - \tfrac{1}{2}} \leq 4 C_\alpha (r_n^+)^\alpha \right\}.
\end{align*} Now, by construction we have $ 4 C_\alpha (r_n^+)^{\alpha} = \epsilon$, and therefore $$\ind\{x \in \cG_\epsilon^+\} \, \ind\{\abs{\eta(x) - \tfrac{1}{2}} \leq 4 C_\alpha (r_n^+)^\alpha \} = 0,$$ and so by Proposition \ref{devroyeLemma2} it remains only to bound 
\begin{align*} \E \left[M(X) \cdot\ind\left\{ M(X) \leq 4C_n | \tilde A_{r_n^+}(X) \cap \bX |^{-1/2}\right\} \right],
\end{align*} which we can achieve by following the exact steps used to prove Theorem \ref{adaptiveTheorem}. It remains only to consider the case $\epsilon < 4 C_\alpha (n_P+n_Q)^{-\alpha}$. Assume $n_P \geq n_Q$. In this case, we have 

\begin{align*}
 \min\left(  \frac{1}{n_P} \epsilon^{-(1 + \gamma^*/\alpha)}, \frac{1}{n_Q} \epsilon^{-(1 + d/\alpha)} \right)  \geq \frac{1}{n_P} \epsilon^{-(1 + d/\alpha)}
 \geq (4 C_\alpha)^{-(1 + d/\alpha)} \ n_P^{-1 + \alpha + d}, 
\end{align*} 

where we use $\gamma^* \geq d$. If $\alpha + d > 1$, this bound exceeds $1$ provided that $n_P \geq (4 C_\alpha)^{\tfrac{\alpha + d}{\alpha(\alpha + d - 1)}}$. If $n_Q \geq n_P$, then using the same argument we get a bound that exceeds $1$ when $n_Q \geq (4 C_\alpha)^{\tfrac{\alpha + d}{\alpha(\alpha + d - 1)}}$, and so the result holds provided $$ \max(n_P,n_Q) \geq (4 C_\alpha)^{\tfrac{\alpha + d}{\alpha(\alpha + d - 1)}}.$$  
\end{proof}

%\begin{remark}[Condition on $\epsilon$]
%Note that Theorem \ref{adaptiveAttainsEpsilonBound} in fact %holds for all $\epsilon \geq 0$ when $d \geq 1$ under the %weak conditions of proposition \ref{noSuperTransfer} (see %Section \ref{secOnDim}; this proposition gives a mild %condition under which we have $\gamma^* \geq d$). To see %this, consider $\epsilon \leq 4 C_\alpha (n_P + %n_Q)^{-\alpha}$, and observe that the second term in %\eqref{highMargin} makes the upper bound vacuous in this %case: we have 

%\end{remark}

%\section{}
%
%We provide here a simple example demonstrating an instance of measures $(P,Q) \in \cT(\alpha, %\beta, \gamma, d)$ (with $Q$ (DM)) such that $\gamma + \alpha \beta < \rho + d$. 
%\par Let $d = 1$, $\cX_Q = \cX_P = [0,1]$, and take $Q_X \sim U([0,1])$. Take $P_X$ to have %density $p \propto (x - 1/2)^{\rho}$ for some $\rho > 1$. Let $\eta(x) = x^2$. Then clearly %$\abs{\eta(x) - \eta(y)} \leq 2|x - y|$, while for $t < 1/2$ we have \begin{align*}
%    Q_X(\abs{\eta(X) - 1/2} \leq t) &= Q_X(\abs{X^2 - 1/2} \leq t) \\& = Q_X(X \in [\sqrt{1/2 - %t}, \sqrt{1/2 + t}])
%    \\& = \sqrt{1/2 + t} - \sqrt{1/2 - t}
%    \\& \leq \sqrt{2t}.
%\end{align*} Finally, note that we have $\gamma = \rho$, and so $(P,Q) \in \cT(\alpha = 1, %\beta = 1/2, \gamma = \rho, d = 1)$, and so $\alpha \beta + \gamma = 1/2 + \rho < \rho + d$. 

\section{Supplemental Experiments}
\label{appendix_experiments}

In this section we present the results of some supplemental experiments. First we carry out a comparison of our method against the dyadic-tree pruning algorithm introduced in \cite{DDT}. We then investigate the performance of our method against the baselines considered in Section \ref{secExperiments}, except we use Algorithm \ref{lepskiAlg} and the baselines to prune trees grown using the CART algorithm \cite{CART}, which are non-dyadic. Our theoretical results do not apply to this setting, but the results nonetheless demonstrate that Algorithm \ref{lepskiAlg} can improve over these baselines in the context of non-dyadic trees.

\subsection{Comparison to Cost-Complexity Pruning}

Decision trees have long been studied in the context of classification, and there exist numerous methods for pruning which employ techniques from structural risk minimization. A noteworthy contribution in this direction is that of \cite{DDT}, who devised an ingenious penalty term for selecting dyadic trees that provably attain minimax optimal learning rates for the expected excess risk under conditions similar to ours in the one-sample case.

Their method consists of penalizing a tree $T$ with leaves $\pi(T)$ according to the following penalty:
\begin{align} \label{DDTpenalty}
    \Phi_n(T) = \sum_{A \in \pi(T)} \sqrt{ 2 p_n(A) \frac{[[A]] \log 2 + \log(2/\delta)}{n}}, 
\end{align} where $p_n(A)$ is the mass of the leaf under the sampling distribution, $\delta \in (0,1)$ is the usual confidence parameter, and $[[A]]$ is chosen such that $ \sum_{A \in \pi(T)} 2 ^{-[[A]]} \leq 1$, which they show can be done via the Kraft inequality for prefix codes \cite{CoverThomas}. The penalty \eqref{DDTpenalty} induces \emph{unbalanced} trees, which can produce better approximations of the decision boundary than balanced trees \cite{DDT}. The local depth selection of Algorithm \ref{lepskiAlg} is similar in that it mimics unbalanced trees by choosing different levels of the tree for different regions of feature space when making predictions, so we provide a comparison against this baseline in order to verify that the advantage of our method is not simply that it yields unbalanced trees, but rather that it does so in a way that correctly aggregates the information added by the source data under covariate shift.  
\par Note that the main result of \cite{DDT} relies on an application of a relative Chernoff bound to derive (c.f. \cite{DDT}, Theorem 2) that with probability at least $1 - \delta$, one has 
\begin{align} \label{DDTtheorem2}
    \abs{R(T) - \hat R_n(T)} \leq \Phi_n(T)
\end{align} for all $T$ in a suitable class of dyadic decision trees, where $R, \hat R_n$ are respectively the risk and empirical risk of the tree $T$. Unfortunately, in the transfer learning setting in which we have access to samples from a source $P$ and a target $Q$, extending the penalty \eqref{DDTpenalty} in the obvious way does not lead to a bound of the form \eqref{DDTtheorem2} with the target risk $R_Q(T)$, but rather would feature the risk under a convex combination of source and target $R_{\alpha P + (1 - \alpha) Q}(T)$, where $\alpha = n_P / (n_P + n_Q)$. It may be possible to derive optimal rates for transfer using a penalty akin to \eqref{DDTpenalty} by employing the risk minimization strategy outlined in \cite{HK_val_targ_data}, although the latter studies a different notion of discrepancy between $P$ and $Q$, and so resulting rates are not directly comparable. We do not pursue this further here. 

We consider two approaches based on the penalty \eqref{DDTpenalty} to further serve as baselines. 

\paragraph{Baseline 1 (SN)} We consider directly implementing the penalty of \cite{DDT}, with the combined sample used to compute the penalty:
\begin{equation*}
    \Phi^{(SN)}_n(T) = \sum_{A \in \pi(T)} \sqrt{ 2 p_n(A) \frac{[[A]] \log 2 + \log(2n)}{n}},
\end{equation*} where $n = n_P + n_Q$ and $p_n(A) = \tfrac{1}{n} \sum_{i=1}^{n_P+n_Q} \ind\{X_i \in A\}$. 

\paragraph{Baseline 2 (SN-Q)} We implement the same penalty, using only the target samples:
\begin{equation*}
    \Phi^{(SNQ)}_{n_Q}(T) = \sum_{A \in \pi(T)} \sqrt{ 2 p_{n_Q}(A) \frac{[[A]] \log 2 + \log(2n_Q)}{n_Q}};
\end{equation*} source samples are used to build the regression estimates in each leaf once the tree has been pruned. 

\par  
In the former case we optimize $R_n(T) + c \, \Phi_n^{(SN)}(T)$ for $T \subset \mathbf{T}$ where $R_n(T)$ is the empirical risk of $T$ under the full sample and $\mathbf{T}$ is an initial cyclical dyadic tree; in the latter, we optimize $R_{n_Q}(T) + c \, \Phi_{n_Q}^{(SNQ)}(T)$, where $R_{n_Q}(T)$ is the empirical risk under the target; the constant $c$ is tuned by cross-validation over the target sample.  

\par
\begin{figure}[t]
    \centering
    \subfloat{{ \includegraphics[height = 6.15cm, width = 8.35cm]{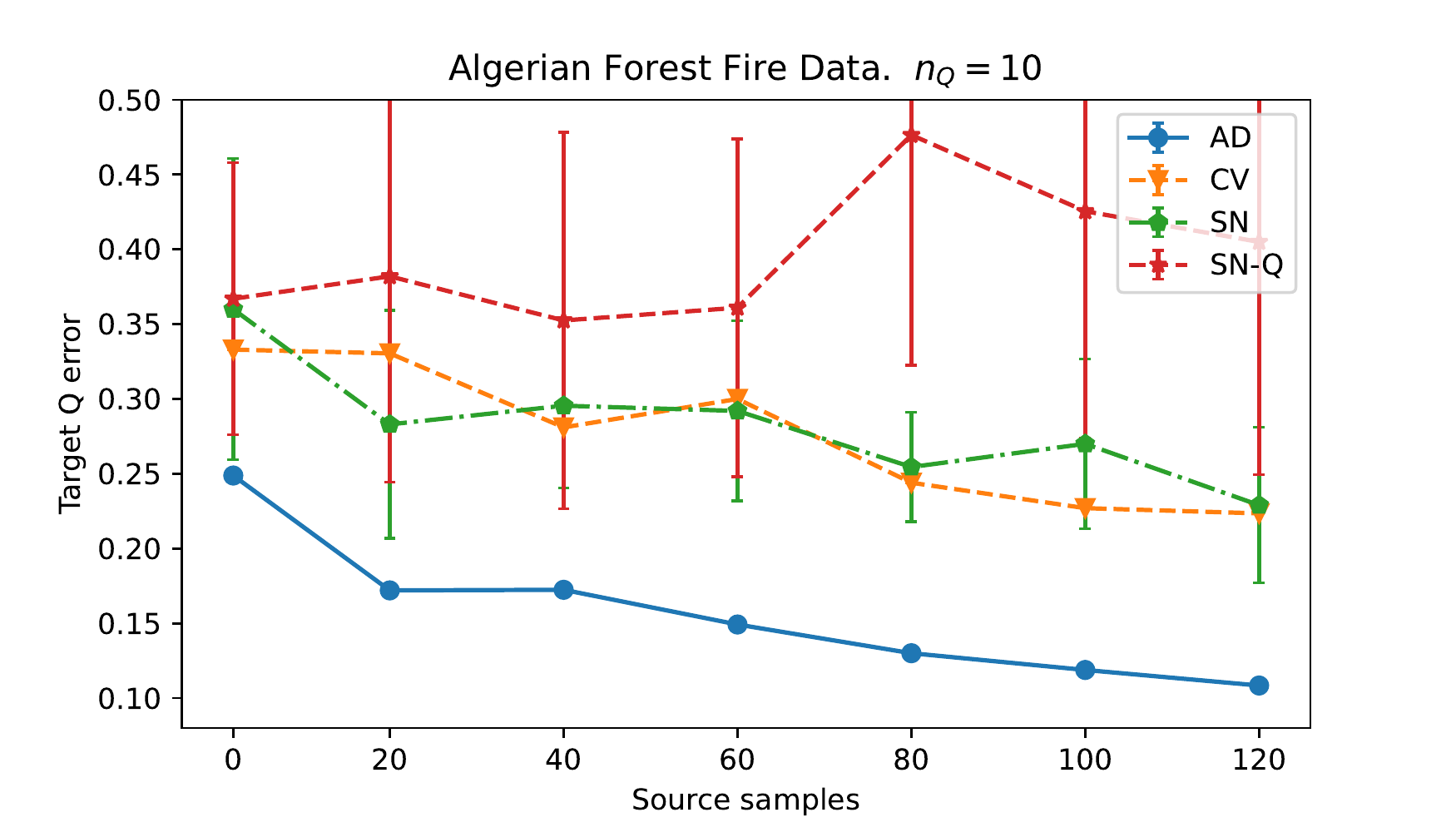} }}
    %\caption{Risk estimates for a regular dyadic tree, with depth selected by 2-fold cross-validation, for the five datasets outlined above. Error bars give the standard deviation over 10 runs. The Bayes risk is $0.18$.}
    %\label{fig:E1_experiment}
%\end{figure}
%\begin{figure}[t]
    \centering
    \subfloat{{\includegraphics[height = 6.05cm, width = 8.3cm]{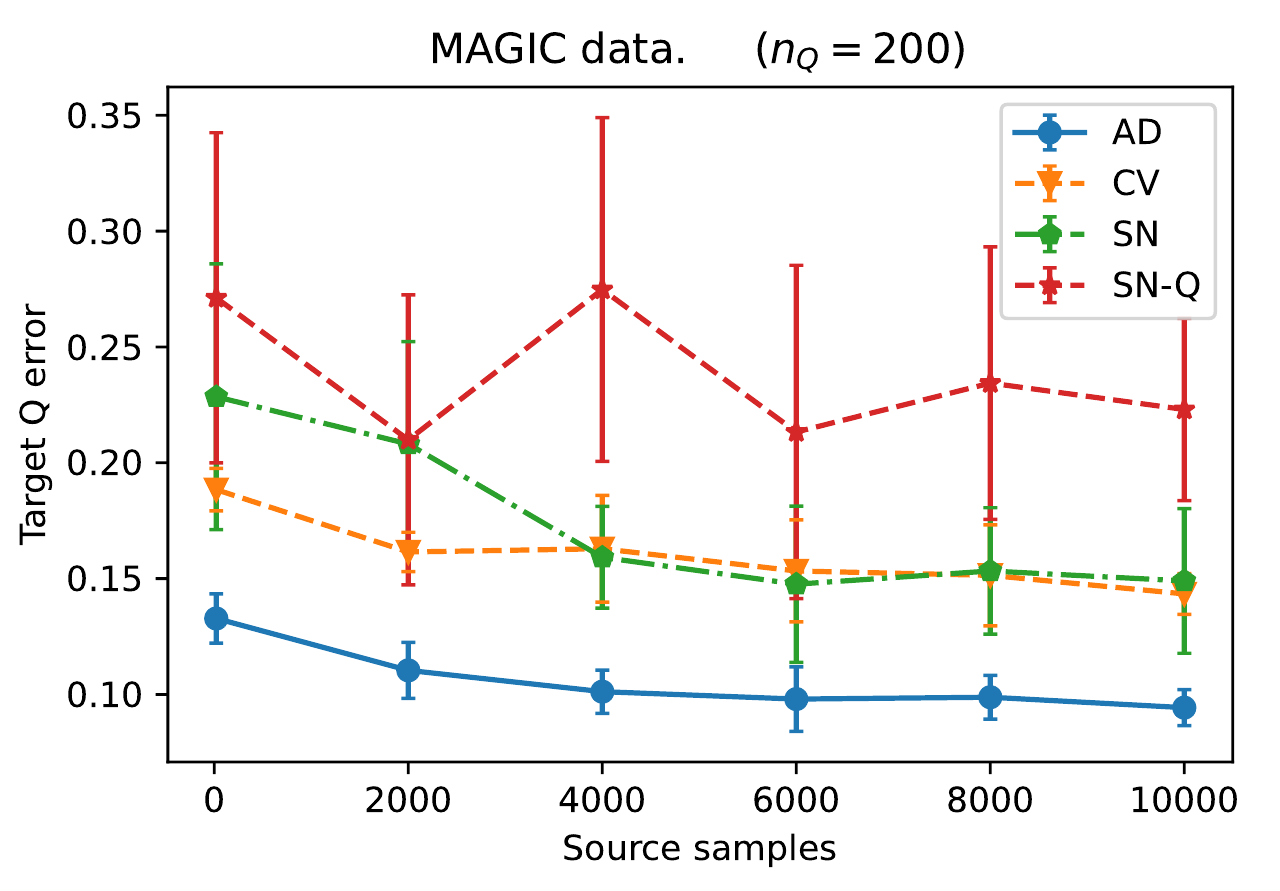} }}
    \caption{Risk estimates for our adaptive method (AD), compared to the baseline cross-validation and the pruning method of \cite{DDT} (SN, SN-Q), applied to the Algerian Forest Fire data (left) and the MAGIC data (right). Error bars show standard errors over 50 and 20 iterations, respectively. }
    \label{fig:SN_baseline}
\end{figure}

 Figure \ref{fig:SN_baseline} shows the result of using the SN penalty to prune a cyclical dyadic tree, against the uniform-depth CV baseline, and against Algorithm \ref{lepskiAlg}. As we see in Figure \ref{fig:SN_baseline}, the cost-complexity pruning approaches (SN and SN-Q) do not achieve the performance of choosing the depth via 2-fold cross-validation over the target sample. We note that, as pointed out in \cite{DDT}, this method is quite sensitive to the choice of dampening constant for the penalty, a problem which is heavily exacerbated in the transfer settings in which there are relatively few target samples (such as those considered here). 

%{\color{red} Of course, as pointed out above, in general neither SN nor SN-Q achieves the rate of Algorithm \ref{lepskiAlg} in the covariate shift setting.} 

%UNCOMMENT FOR ONE-COLUMN FORMAT:

%TWO-COL FORMAT:
%\begin{figure}[t]
%    \centering
%    \includegraphics[height = 12cm]{plot_SNQ_baseline.pdf}
%    \caption{Risk estimates for our adaptive method (AD), compared to the baseline cross-validation and the pruning method of \cite{DDT} (SN, SN-Q), applied to the Algerian Forest Fire data (top) and the MAGIC data (bottom). Error bars show standard errors over 50 and 20 iterations, respectively. }
%    \label{fig:SN_baseline}
%\end{figure}

\subsection{On the Case $n_P = 0$.}

\begin{figure}[t]
    \centering
    \subfloat{{ \includegraphics[height = 6.15cm, width = 8.35cm]{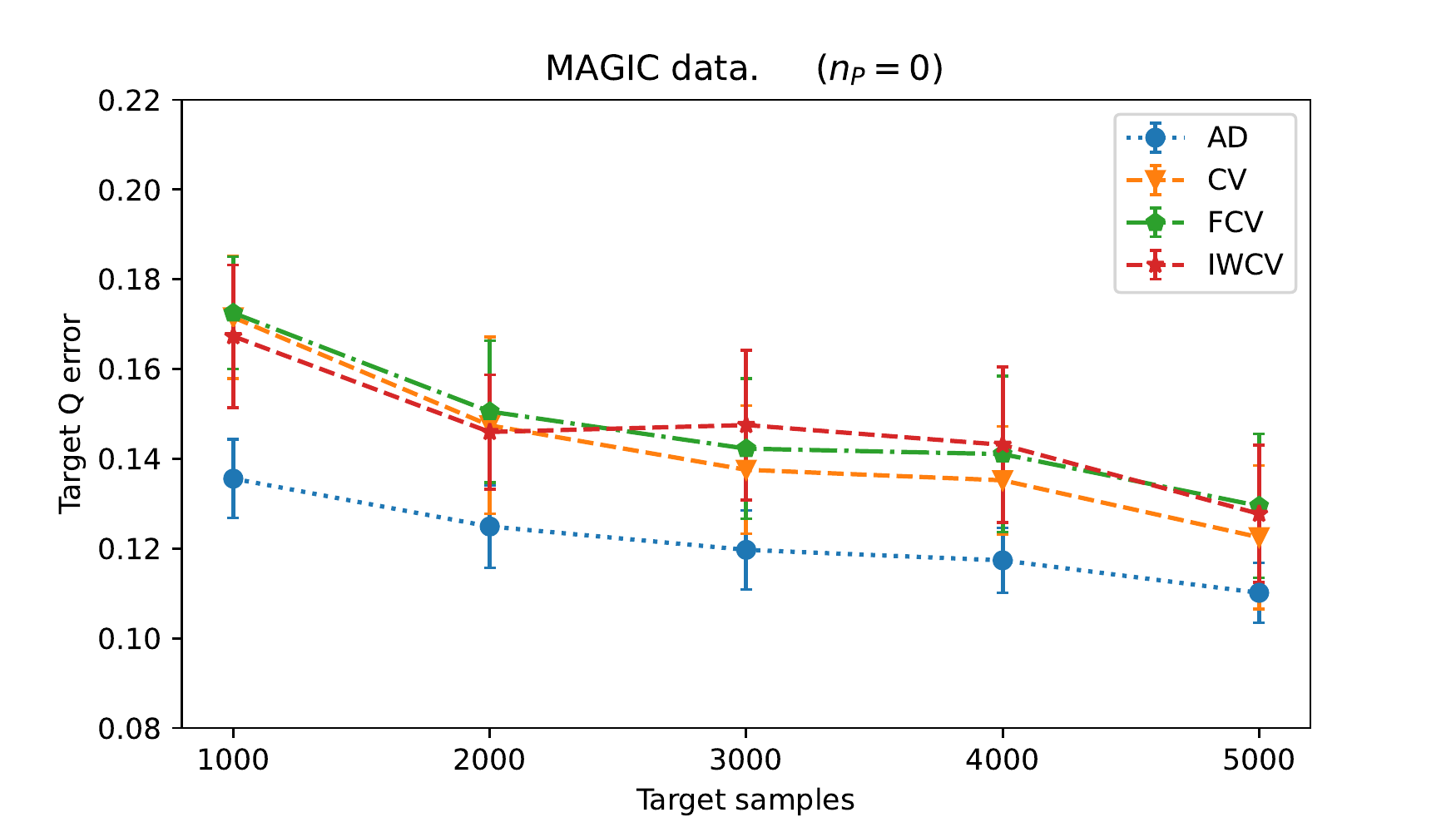} }}
    %\caption{Risk estimates for a regular dyadic tree, with depth selected by 2-fold cross-validation, for the five datasets outlined above. Error bars give the standard deviation over 10 runs. The Bayes risk is $0.18$.}
    %\label{fig:E1_experiment}
%\end{figure}
%\begin{figure}[t]
    \centering
    \subfloat{{ \includegraphics[height = 6.1cm, width = 8.4cm]{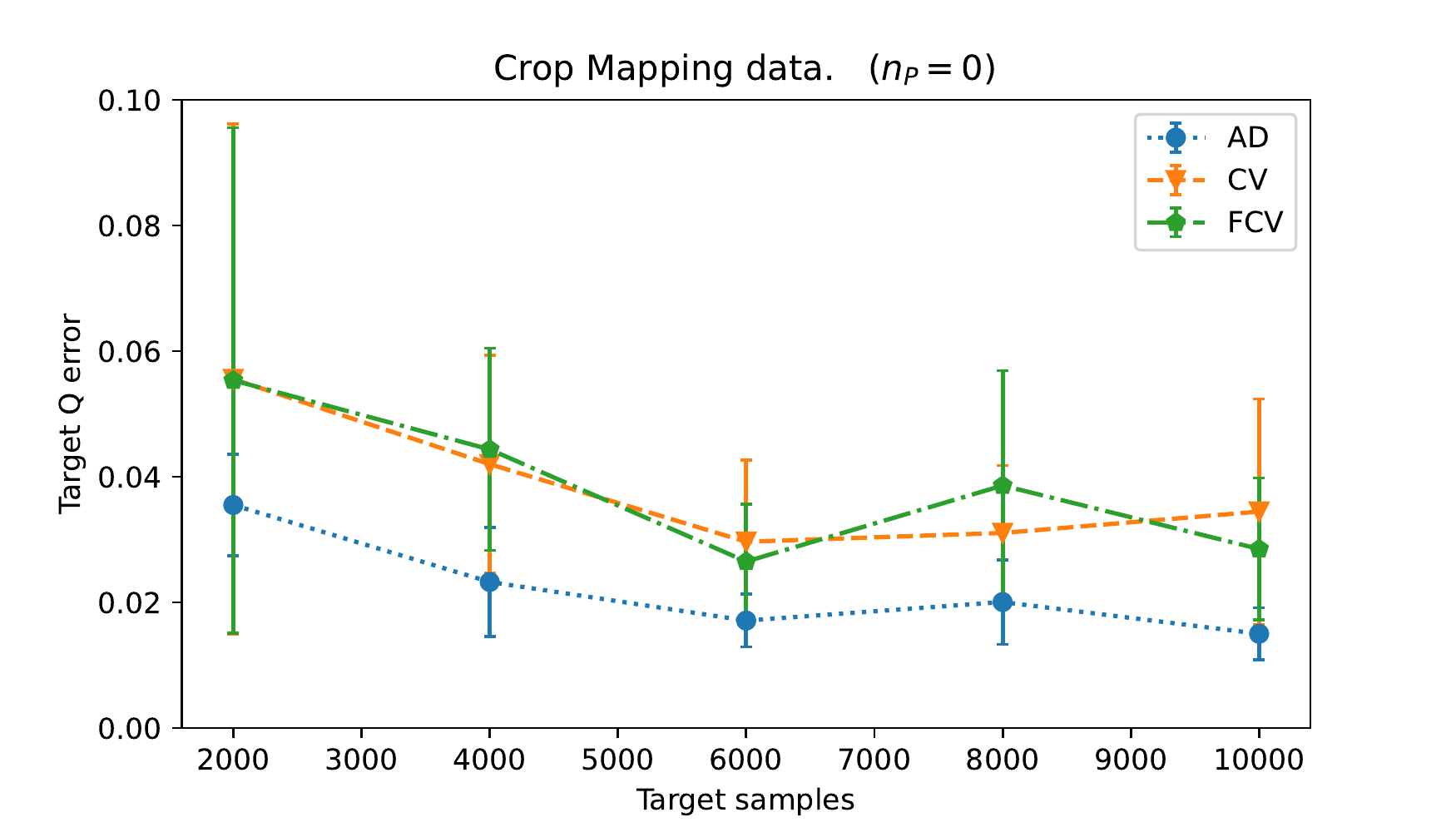} }} 
    \caption{ Target risk estimates for three tree-depth selection methods applied the Crop Mapping (d = 174) and MAGIC (d = 10) datasets when $n_P = 0$. Error bars show the standard errors over 10 iterations.}
    \label{fig:CM_np0}
\end{figure}

Methods such as our Algorithm 1, based on the ICI approach indebted to Lepski's method, are to the best of our knowledge currently the only classifiers that are able to achieve the (log-spoiled) minimax optimal rates that we have presented here. However, there is nothing particular to the two-sample transfer learning problem about ICI in general or Algorithm 1 in particular, and the procedure yields minimax optimal rates in the case $n_P = 0$ as well, leading us to expect that it should be at least competitive with other methods in this case. Figure \ref{fig:CM_np0} demonstrates that this is true for the Crop Mapping and MAGIC data, where we see that AD slightly outperforms the CV-based baselines. We leave a more thorough empirical investigation of this question to further work.

\subsection{Beyond Dyadic Trees} 

Here we repeat the experiments shown in Section \ref{secExperiments}, but we implement the various depth-selection methods on trees grown using the vanilla CART algorithm of \cite{CART} (that is, splits are chosen to minimize the so-called Gini impurity at each node). For the sake of variety, we consider two further datasets: Wine Quality, and Steel Plates Faults, described in Table \ref{tab:datasets}. Note that the Wine Quality dataset has 10 classes indicating various quality levels. We label wines with quality at least $6$ as $1$ and the rest as $0$ to get a binary classification problem. \par
\begin{figure}[t]
    \centering
    \subfloat{{ \includegraphics[height = 6.05cm, width = 8.1cm]{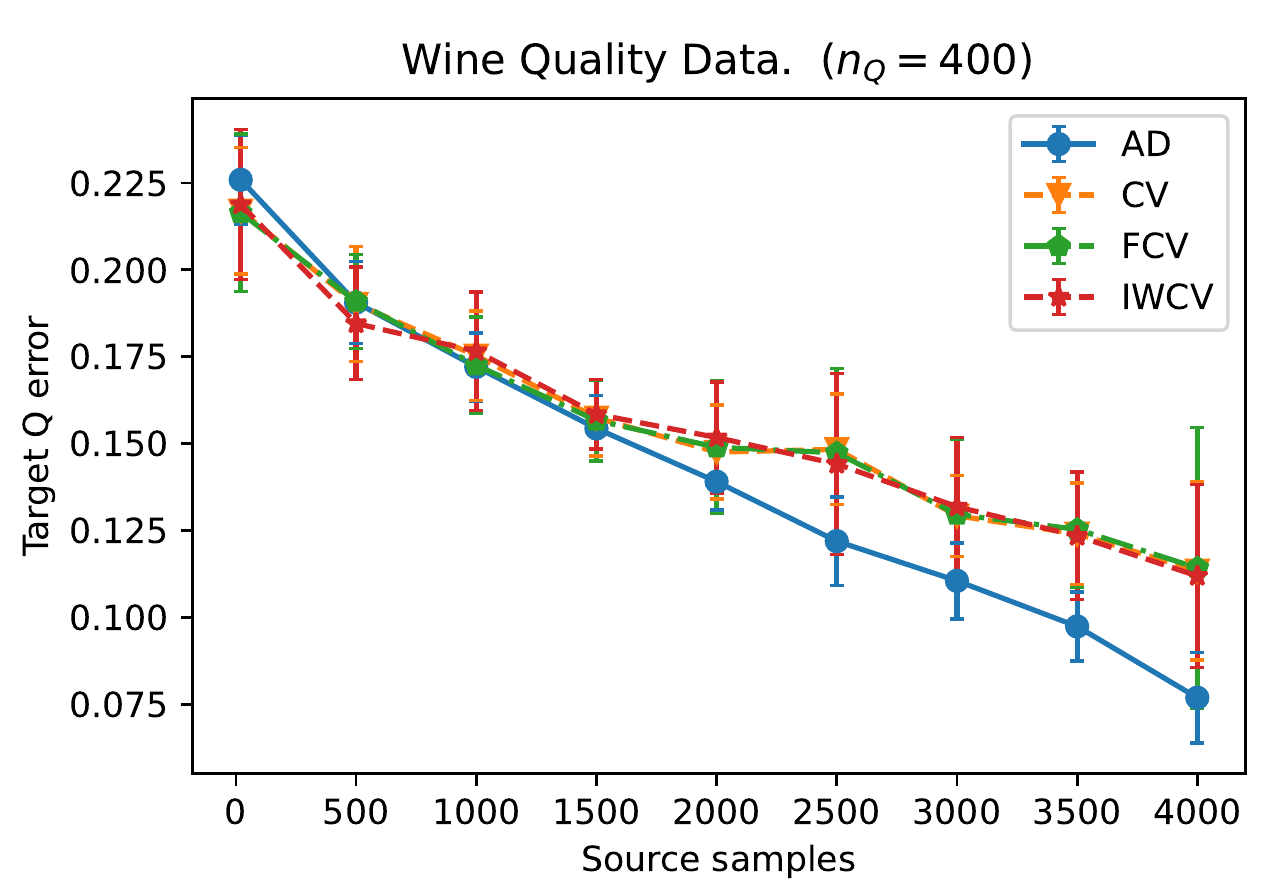} }}
    %\caption{Risk estimates for a regular dyadic tree, with depth selected by 2-fold cross-validation, for the five datasets outlined above. Error bars give the standard deviation over 10 runs. The Bayes risk is $0.18$.}
    %\label{fig:E1_experiment}
%\end{figure}
%\begin{figure}[t]
    \centering
    \subfloat{{\includegraphics[height = 6cm, width = 8.3cm]{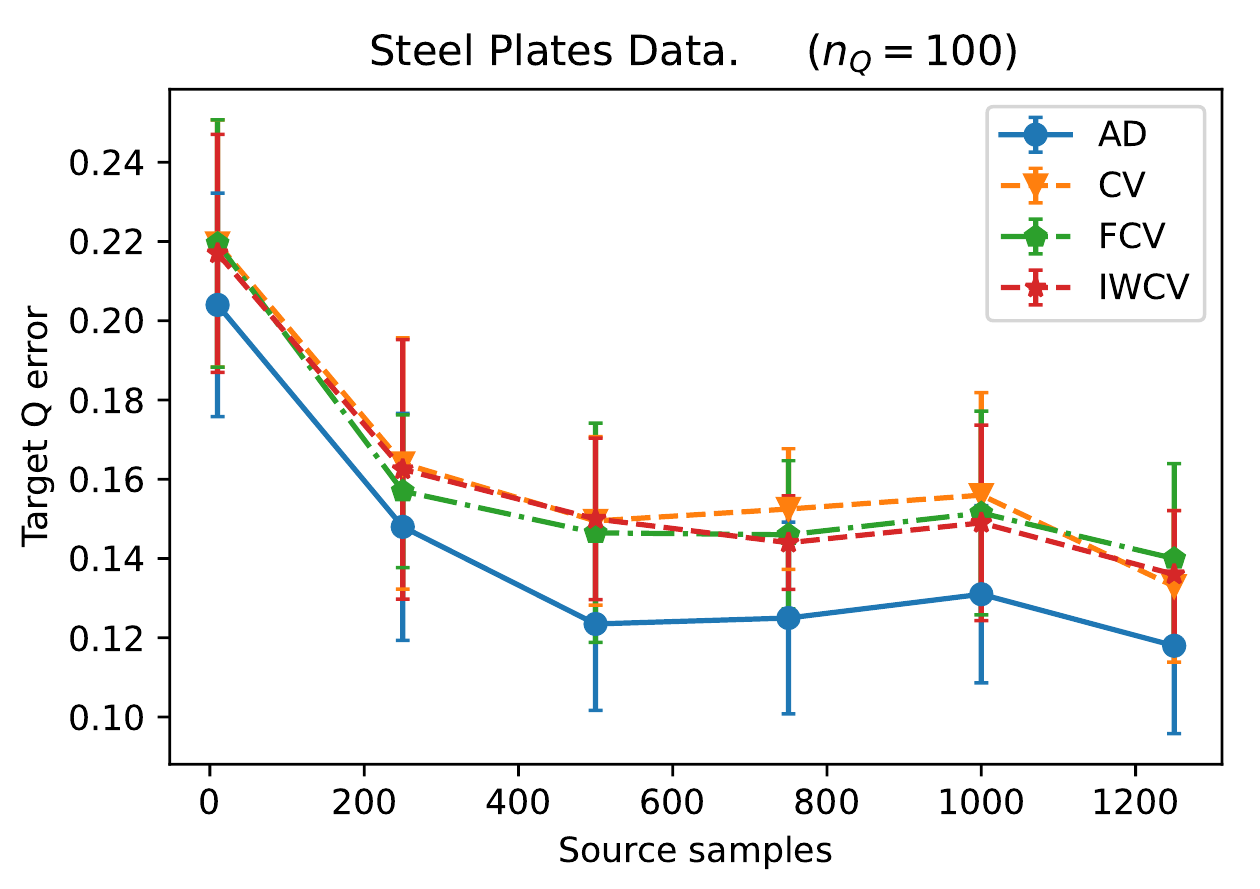} }}
    \caption{ Target risk estimates for four tree-depth selection methods applied to CART trees for the Wine Quality dataset (d = 11; left), and the Steel Plates Faults dataset (d = 27; right). Error bars show the standard errors over 10 iterations.}
    \label{fig:CART_plots}
\end{figure}

 
Although we can no longer claim the rates of Theorem \ref{oracleRate} when Algorithm \ref{lepskiAlg} is applied to these trees, as we see in Figure \ref{fig:CART_plots}, the adaptive  (AD) performs at least as well and sometimes significantly better than the cross-validation based depth-selection methods, suggesting that it could be useful as a pruning procedure even when trees are grown using more elaborate splitting criteria. We leave further exploration of this possibility for future work.

%UNCOMMENT FOR ONE-COLUMN FORMAT:

%two-COL FORMAT:
%\begin{figure}[t]
%    \centering
%    \includegraphics[height = 12cm]{plot_CART_experiments.pdf}
%    \caption{Target risk estimates for four tree-depth selection methods applied to CART trees for the Wine Quality dataset (d = 11) (top), and the Steel Plates Faults dataset (d = 27) (bottom). Error bars show the standard errors over 10 iterations.}
%    \label{fig:CART_plots}
%\end{figure}

%Unfortunately approaches of this nature are sub-optimal in transfer learning problems, because the empirical risk under the combined sample is a biased estimate of the risk under the target measure. As we discuss elsewhere \skr{(I presume)}, this difficulty can in theory be circumvented by leveraging importance-weighting techniques, although these in turn rely on density-ratio estimates, which in our experience can be quite unreliable even in problems of only moderately large dimension.  

\section{On the Multiple Source Problem.} \label{appendix_multiSource}

A simple extension of Lemma \ref{VexpBound} allows us to re-express our results when there are a multiplicity of source distributions. Suppose that we are now in a setting in which there are $k$ source distributions $P_1, \dots, P_k$, such that each $(P_i, Q) \in \cT$ with transfer exponents $\gamma_i$ (with the same constant $C_\gamma$, c.f. \eqref{aggExpDef}); without loss of generality, suppose that the target distribution is $P_1$. Assuming a sample of size $n = n_1 + \dots + n_k$ with $n_i$ samples from $P_i$, the natural extension of the local variance term to multiple samples is 
\begin{equation*}
    V_r(x) = \frac{1}{\sum_{i=1}^k n_i P_i(\tilde A_r(x))}. 
\end{equation*} In this case, we have 

\begin{lemma}\label{multiSourceVarBound}
Consider the multiple source setting with $n_i$ samples from source $i$, $P_i$, where $P_i$ has aggregate exponent $\gamma_i$ w.r.t. the source, with constant $C_\gamma$. Then we have 
$$ \E_{Q_X} [ V_r(X)] \leq C_\gamma \, \frac{1}{n} r^{-\bar \gamma},$$ where $\bar \gamma = \sum_{i=1}^k w_i \gamma_i$, with weights $w_i = n_i/n$. 
\end{lemma}

\begin{proof}
    Let $I$ denote a single draw from the multinomial distribution with probabilities $w = (w_1, \dots, w_k)$. As usual, $X$ is drawn from $Q_X$, $\tilde A_r(x)$ denotes the envelope of the cell containing $X$ at level $r$. We let $P_I(A)$ and $\gamma_I$ be random variables equal to $P_i(A)$ and $\gamma_i$ when $I = i$. We have, for $0 < r < 1$, 
\begin{align*}        
    \E_{Q_X} [ V_r(X)] & = \mathbb{E}_{Q_X} \left[ \left\{ \sum_{i=1}^k n_{i} P_i(\tilde A_r(X))  \right\}^{-1} \right] = \frac{1}{n} \, \mathbb{E}_{Q_X} \left[ \left\{ \sum_{i=1}^k w_i P_i(\tilde A_r(X))  \right\}^{-1} \right] \\&= \frac{1}{n} \, \E_{Q_X} \left\{ \E_I P_I(\tilde A_r(X))\right\}^{-1}  
        \leq  \frac{1}{n} \,  \E_{Q_X} \left\{\E_I P_I(\tilde A_r(X))^{-1}\right\} 
        \\& = \frac{1}{n} \, \E_I \E_{Q_X} P_I(\tilde A_r(X))^{-1} 
        \\& = \frac{1}{n} \, \E_I \left( \sum_{A \in \pi(T_r)} \frac{Q_X(A_r(X))}{P_I(\tilde A_r(X))} \right) 
        \leq \frac{1}{n} \, \E_I \left( \sum_{A \in \pi(T_r)} \frac{Q_X(\tilde A(X))}{P_I(\tilde A(X))} \right)
        \\& \leq \frac{1}{n}  \, \E_I \, C_\gamma r^{-\gamma_I}
        \leq  C_\gamma \, \frac{1}{n} \, r^{- \E \gamma_I}
        =  C_\gamma \, \frac{1}{n} \, r^{- \sum w_i \gamma_i},
    \end{align*} where the inequality on the second line and the second inequality on the fifth line are applications of Jensen's inequality, and the first inequality on the fifth line is an application of Lemma \ref{gammaSumLemma}. 
\end{proof}

Using this Lemma, the proofs of our upper bounds can be repeated with no change (since the aggregate transfer exponent works its way into the rates only via this variance bound) giving results in terms of the full sample size $n$ and the sample-weighted average of the aggregate transfer exponents, $\bar \gamma$. Unlike the one-source case bounds we have presented however, it remains unclear whether the risk bounds for multiple sources $k \geq 2$ expressed in terms of $\bar \gamma$ are tight in any meaningful sense.

\section{A Pathological Example.}\label{notAmbientDyadic}

In this section we provide an example of non-doubling measures $P,Q$ for which the aggregate and integrated transfer exponents are equal - $\gamma^* = \rho^*$ - but that these do not agree with the value that we would obtain if we defined the exponent by taking the sum of mass ratios over the dyadic partition of the ambient space. Suppose that we have measures $P,Q$ supported on $[0,1]^d$, and define
\begin{align*} \Lambda(P,Q,n) = \sum_{C \in \mathcal{D}^n} \frac{Q(C)}{P(C)},
\end{align*}
%\begin{equation}
%    \Dim_D(P,Q) = \lim_{n \to \infty} \frac{ \log \Lambda(P,Q,n)}{n \log 2},
%\end{equation} 
where $\mathcal{D}^n$ is the regular dyadic partition of $[0,1]^d$ of order $n$. In analogy with how we have defined the aggregate exponent between $P$ and $Q$, let $\gamma_D(P,Q) = \gamma_D > 0$ be any constant for which there exists a $C > 0$ such that for all $n = 0,1,2, \dots$ we have 
\begin{align*}
    \Lambda(P,Q,n) \leq C \, 2^{n \gamma_D}, 
\end{align*} and let $\gamma^*_D$ be the minimal $\gamma_D$ such that the above holds. Let $d = 2$, and consider the following construction: consider two countable sequence of boxes $\{B_i; \: i = 1,2, \dots\}$ and $\{B_i^\prime; \: i = 1,2, \dots\}$, where $B_i$ is a square with side length $2^{-i}$ with top-right corner at $x_i = (\sum_{k=0}^{i-1} 2^{-i}, 2^{-i})$, and $B_i^\prime$ is a square of side-length $2^{-2i}$ with bottom-left corner at $x_i$; see Figure~\ref{counterExample} below. We take $Q = \sum_{i=1}^\infty Q_i$ and $P = P_0 + \sum_{i=1}^\infty P_i$, where $Q_i, P_i$ are supported on $B_i \cup B_i^\prime$ for each $i \geq 1$. Let $Q_i$ uniformly assign mass $3 \cdot (1/4)^i$ to $B_i \cup B_i^\prime$, yielding density $q_i$. Let $P_i$ be equal to $Q_i$ on $B_i$. On $B_i^\prime$, let $P_i$ have density $q_i\norm{x - x_i}^\nu$ for some $\nu > 0$. Finally, let $P$ have an atom at $x_0 = (0,1)$ with all remaining mass. Note that, by calculations identical to those from Example 2, we have $\gamma^*(P,Q) = \max(2,\nu)$. %Now, let \mbox{$\bar B_n = \cup_{i=n+1}^\infty (B_i \cup B_i^\prime),$} and note that 
We proceed to estimate $\Lambda(P,Q,n)$ for $n \geq 1$. Observe that by construction, there is a cube $C \in \mathcal{D}^n$ such that $Q(C) = Q(B_n^\prime)$, and $P(C) = P(B_n^\prime)$. Further, there is a constant $c_0$ such that $\cup_{i=1}^n B_i$ is exactly covered by $c_0 2^{2n}$ cubes in $\mathcal{D}^n$, and by construction for each of these cubes we have $Q(C)/P(C) = 1$. Therefore we have 
\begin{align*}
    \Lambda(P,Q,n) = \sum_{C \in \mathcal{D}^n} \frac{Q(C)}{P(C)} \geq c_o 2^{2n} + \frac{Q(B_n^\prime)}{P(B_n^\prime)} 
    = c_0 2^{-2n} + \frac{q_n 2^{4n}}{q_n c(\nu) 2^{2n(2 + \nu)}} = c_0 2^{2n} + c(\nu)^{-1} \, 2^{n\,  2 \nu},
\end{align*} where $c(\nu) > 0$ is some constant depending only on $\nu$. It follows immediately that $\gamma^*_D(P,Q) \geq \max(2,2\nu)$, and so for $\nu > 1$ we have $\gamma^*_D(P,Q) > \gamma^*(P,Q)$. In particular, this shows that if we have used $\gamma_D$ to derive our results, the induced rates would not be as sharp as those using the aggregate exponent $\gamma$. Observe also that since $P$ is not a doubling measure, this example also serves to demonstrate that doubling is not necessary for the equivalence in Proposition \ref{relDimEquivalence} to hold, since one can straightforwardly verify that $\gamma^* = \rho^*$ in this case.

%\begin{wrapfigure}{r}{7cm}\label{counterExample}
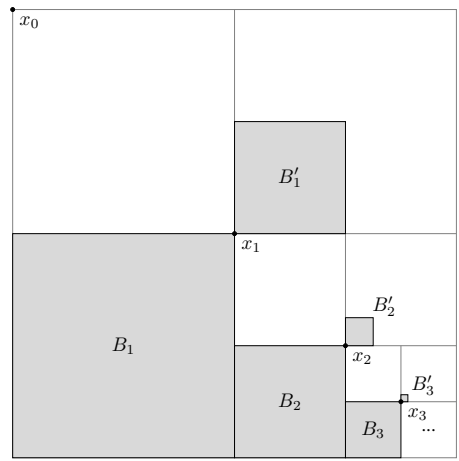
\begin{figure}[ht]
    \centering
    \resizebox{6cm}{6cm}{%
    \begin{tikzpicture}
    \draw[gray, thin] (0,0) rectangle (8,8);
    \filldraw[fill=black!15!white, draw=black] (0,0) rectangle (4,4);
    \filldraw[fill=black!15!white, draw=black] (4,0) rectangle (6,2);
    \filldraw[fill=black!15!white, draw=black] (6,0) rectangle (7,1);
    \draw[gray, thin] (4,4) -- (4,8);
    \draw[gray, thin] (4,4) -- (8,4);
    \draw[gray, thin] (6,2) -- (6,4);
    \draw[gray, thin] (6,2) -- (8,2);
    \draw[gray, thin] (7,1) -- (7,2);
    \draw[gray, thin] (7,1) -- (8,1);
    \filldraw[fill=black!15!white, draw=black] (4,4) rectangle (6,6);
    \filldraw[fill=black!15!white, draw=black] (6,2) rectangle (6.5, 2.5);
    \filldraw[fill=black!15!white, draw=black] (7,1) rectangle (7.125, 1.125);
    
    \node[] at (2,2) {$B_1$};
    \node[] at (5,5) {$B_1^\prime$};
    \node[] at (5,1) {$B_2$};
    \node[] at (6.7,2.7) {$B_2^\prime$};
    \node[] at (6.5,0.5) {$B_3$};
    \node[] at (7.4,1.3) {$B_3^\prime$};
    \node[] at (7.5,0.5) {...};
    
    \filldraw[black] (4,4) circle (1pt) node[anchor=north west] {$x_1$};
    
    \filldraw[black] (0,8) circle (1pt) node[anchor=north west] {$x_0$};
    
    \filldraw[black] (6,2) circle (1pt) node[anchor=north west] {$x_2$};
    
    \filldraw[black] (7,1) circle (1pt) node[anchor=north west] {$x_3$};
    
    \end{tikzpicture}
    }
    \caption{Illustration of the support of the measures $P,Q$ described above on the unit square $[0,1]^2$.}
    \label{counterExample}

\end{figure}

\end{document}